%% file: main.tex
\documentclass{article}

\PassOptionsToPackage{numbers, compress}{natbib}

    \usepackage[final]{neurips_2024}

\usepackage[utf8]{inputenc} %
\usepackage[T1]{fontenc}    %
\usepackage{hyperref}       %
\usepackage{url}            %
\usepackage{booktabs}       %
\usepackage{amsfonts}       %
\usepackage{nicefrac}       %
\usepackage{microtype}      %
\usepackage{xcolor}         %
\usepackage{subcaption}
\usepackage{graphicx}

\usepackage{algorithm}
\usepackage{algorithmic}
\usepackage{natbib}
\usepackage{multirow}
\usepackage{xcolor,colortbl}
\definecolor{Gray}{gray}{0.85}
\definecolor{LightCyan}{rgb}{0.88,1,1}
\newcolumntype{g}{>{\columncolor{Gray}}r}
\newcolumntype{b}{>{\columncolor{LightCyan}}r}
\usepackage{wrapfig}
\usepackage{comment}

\usepackage{bm}
\usepackage{amsmath}
\usepackage{amsthm}
\usepackage[capitalize,noabbrev]{cleveref}
\theoremstyle{plain}
\newtheorem{theorem}{Theorem}[section]

\newtheorem{corollary}[theorem]{Corollary}
\theoremstyle{definition}

\theoremstyle{remark}

\newcommand{\new}[1]{{{\textcolor{black}{#1}}}}

\newcommand{\alg}{\textsc{S2L}}
\newcommand{\fullname}{\textsc{\underline{S}mallTo\underline{L}arge}}

\title{{\fullname} ({\alg)}: Scalable Data Selection for Fine-tuning Large Language Models by Summarizing Training Loss Trajectories of Small Models}

\author{%
Yu Yang$^1$ \quad Siddhartha Mishra$^1$ \quad Jeffrey Chiang$^2$ \quad Baharan Mirzasoleiman$^1$\\
$^1$Department of Computer Science, $^2$Department of Computational Medicine\\
University of California, Los Angeles (UCLA)\\
}

\begin{document}

\maketitle

\begin{abstract}
\input{sections/0_abs}
\end{abstract}

\input{sections/1_intro}
\input{sections/2_related}

\input{sections/3_method}

\input{sections/4_experiments}

\input{sections/5_conclusion}

\section*{Acknowledgments}
This research was partially supported by the National Science Foundation CAREER Award 2146492, National Science Foundation 2421782 and Simons Foundation, Cisco Systems, Optum AI, and a UCLA Hellman Fellowship.

\bibliography{ref}
\bibliographystyle{plainnat}

\appendix

\newpage
\input{sections/6_appendix}

\end{document}

%% file: sections/0_abs.tex
Despite the effectiveness of data selection for pretraining and instruction fine-tuning large language models (LLMs), %
improving data efficiency in supervised fine-tuning (SFT) for specialized domains poses significant challenges due to the complexity of fine-tuning data. 
To bridge this gap, we introduce an effective and scalable data selection method for SFT, 
{\fullname} ({\alg)}, 
which trains a small model, clusters loss trajectories of the examples, and samples from these clusters 
to guide data selection for larger models. 
We prove that during fine-tuning, samples within the same loss trajectory cluster exhibit similar gradients. Then, we show that S2L subsets have a bounded gradient error w.r.t. the full data, hence guarantee convergence to the neighborhood of the optimal solution.
We demonstrate 
through extensive experiments
that {\alg} significantly improves data efficiency in SFT for mathematical problem-solving, reducing the training data requirement to just 11\% of the original MathInstruct dataset \cite{yue2023mammoth} to match full dataset performance while outperforming state-of-the-art data selection algorithms by an average of $4.7\%$ across 6 in- and out-domain evaluation datasets.
Remarkably, selecting only 50K data for SFT, {\alg} achieves a 32.7\% accuracy on the challenging MATH \cite{hendrycks2021measuring} benchmark, improving Phi-2 \cite{textbooks2} by 16.6\%. 
In clinical text summarization on the MIMIC-III dataset \cite{johnson2016mimic}, {\alg} again outperforms training on the full dataset using only 50\% of the data. 
Notably, {\alg} can perform scalable data selection using a reference model $100\times$ smaller than the target model, proportionally reducing the computational cost. \footnote{Code is available at \href{https://github.com/BigML-CS-UCLA/S2L}{https://github.com/BigML-CS-UCLA/S2L}.}

%% file: sections/1_intro.tex
\section{Introduction}
In recent years, large language models (LLMs) have revolutionized artificial intelligence by demonstrating an unprecedented ability to understand and generate human language \cite{brown2020gpt}. Among all the contributing factors, the quality and selection of data is becoming increasingly recognized for its importance in training LLMs effectively.
Recent research indicates that LLMs benefit more from training for additional epochs on carefully curated data rather than on larger, uncurated ones during pretraining \cite{tirumala2023d} and instruction fine-tuning \cite{zhou2023lima}, making data selection one of the most promising means to unlock the next level of LLMs' language capability.  
\begin{figure*}[ht!]
\begin{center}
     \begin{subfigure}[b]{0.3\textwidth}
         \centering
         \includegraphics[width=0.8\textwidth]{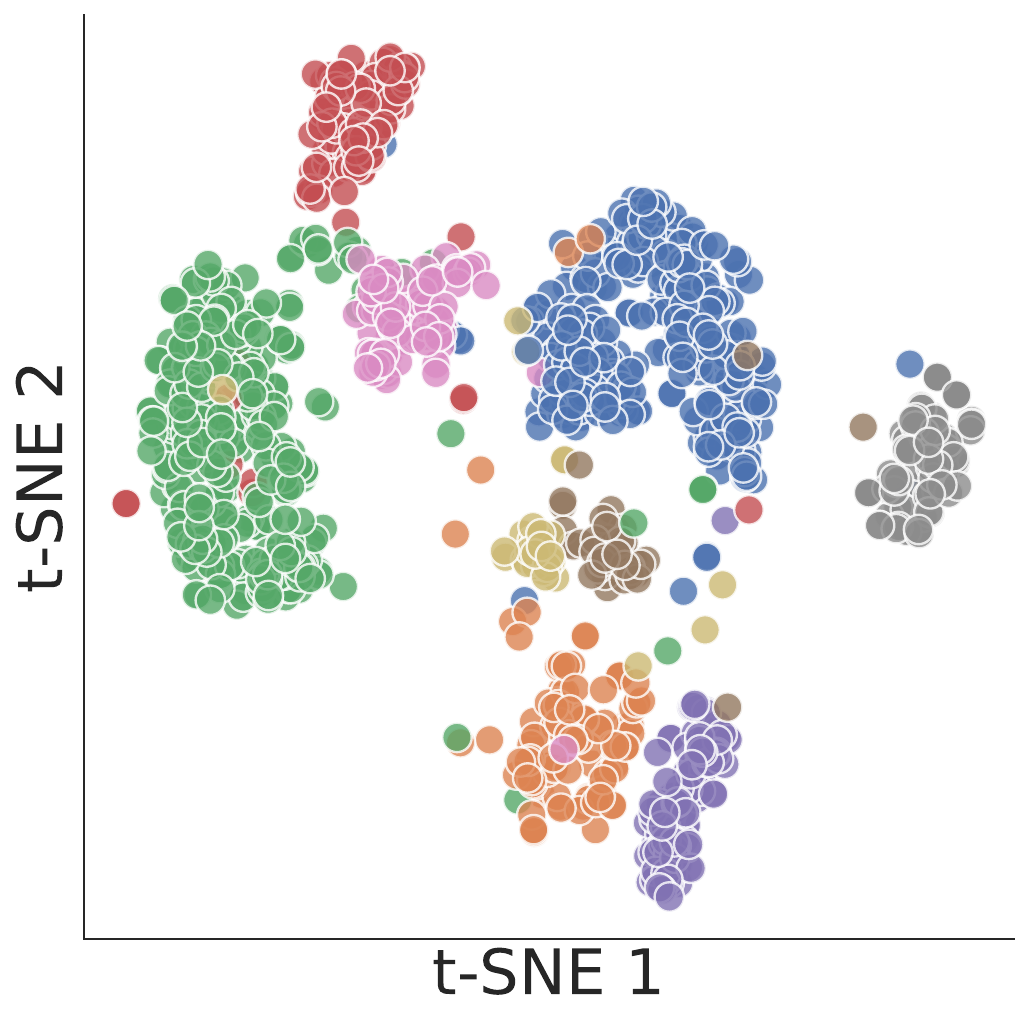}
         \caption{Hidden states of the Pile on pretrained Pythia-410M}
         \label{fig:pretrain-pile}
     \end{subfigure}
     \hfill
     \begin{subfigure}[b]{0.3\textwidth}
         \centering
         \includegraphics[width=0.8\textwidth]{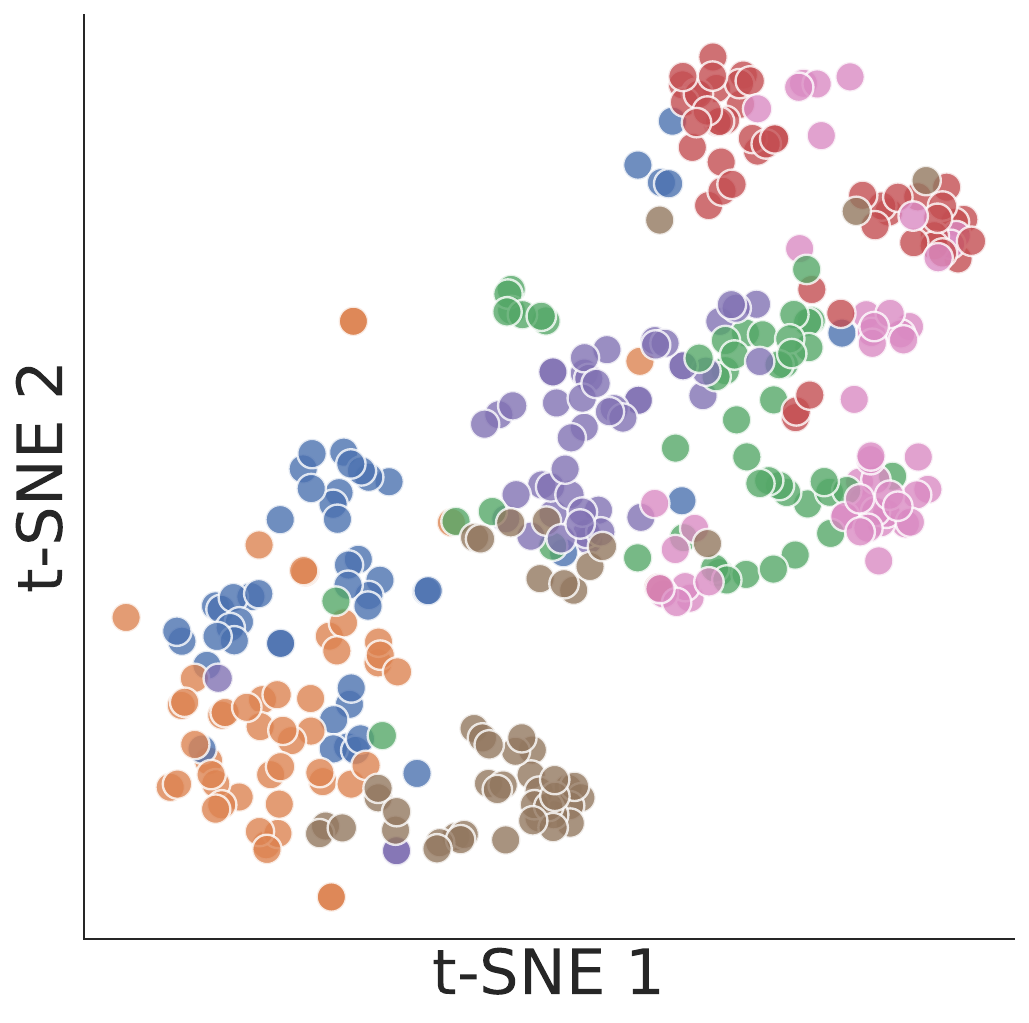}
         \caption{Hidden states of MathInstruct on pretrained Pythia-410M}
         \label{fig:pretrain-math}
     \end{subfigure}
     \hfill
     \begin{subfigure}[b]{0.3\textwidth}
         \centering
         \includegraphics[width=0.8\textwidth]{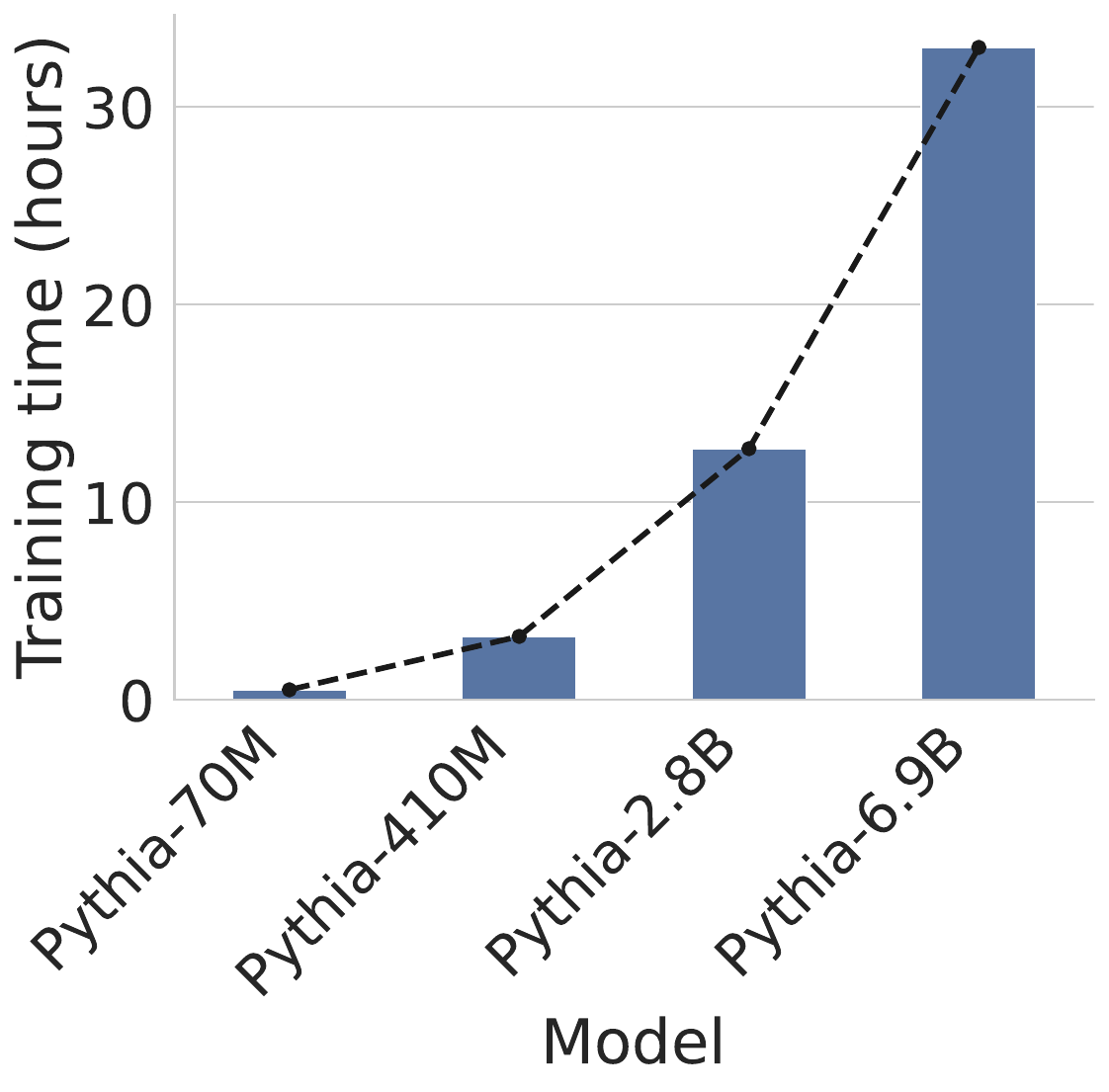}
         \caption{Increase in training time as the size of the model scales up}
         \label{fig:training-time}
     \end{subfigure}
\caption{Existing data selection methods depend heavily on the feature representations from a reference model, which makes their effectiveness vulnerable to the quality of training on the target domain \cite{marion2023less}. For supervised fine-tuning (SFT), while pretrained models can effectively separate topics (shown in different colors) in natural language (\cref{fig:pretrain-pile}), they struggle with fine-tuning data that deviates from the pretraining distribution (\cref{fig:pretrain-math}). Additionally, the cost of training a reference model escalates with model size (\cref{fig:training-time}), making existing data selection methods for large models prohibitively expensive.}
\label{fig:embed}
\end{center}
\vskip -0.2in
\end{figure*}
However, while generalist models obtained through pre-training or instruction fine-tuning excel in \textit{general language tasks}, 
they may not deliver optimal outcomes in \textit{specialized domain}, such as mathematics \cite{azerbayev2023llemma,luo2023wizardmath,yu2023metamath,liu2023tinygsm,yue2023mammoth}, code \cite{roziere2023code,luo2023wizardcoder}, medicine \cite{singhal2023medpalm,singhal2023medpalm2,cheng2023adapting}, or finance \cite{wu2023bloomberggpt,cheng2023adapting}. These domains are not only critical for real-world applications but also hold substantial economic and societal impacts. 

To maximize performance in specialized domains, models fine-tuned on domain data offer superior capabilities over generalist models \cite{pmlr-v202-jang23a}. 
Yet, maximizing the data efficiency in supervised fine-tuning (SFT) for specialized domains remains a challenging and under-explored problem. %
Firstly, heuristic approaches that are effective in the instruction fine-tuning stage, like manual curation \cite{zhou2023lima} or using advanced models such as GPT-4 for dataset evaluation \cite{chen2023alpagasus}, are less reliable due to the need for specialized knowledge and become costly with large volumes of uncurated fine-tuning data. 
Beyond these heuristic methods, other approaches rely on generating representations for each training example using a reference model, often utilizing metrics like perplexity \cite{marion2023less}, confidence \cite{swayamdipta-etal-2020-dataset,varshney-mishra-and-chitta-baral-2022-model}, or hidden states \cite{abbas2023semdedup,tirumala2023d,yang2023decoding,bhatt2024experimental} as features. 
However, these techniques also fall short in SFT for specialized domains for two reasons: (1) the significant shift between pretraining and SFT data can render these %
metrics less informative (\cref{fig:pretrain-math}), and (2) the computation and memory demands associated with generating representations for each training example become prohibitive, as these specialized domains often require larger models, some with up to 540 billion parameters \cite{chowdhery2023palm,singhal2023medpalm}, leading to substantial scalability challenges (\cref{fig:training-time}).

To tackle the challenges of data efficiency in SFT for specialized domains, we present {\fullname} ({\alg)}, an effective and scalable data selection algorithm. {\alg} operates by first gathering training loss trajectories for each training example using a small model. These trajectories are then clustered, and similar number of examples are selected from these clusters uniformly at random. \new{This process is grounded in our theoretical findings that examples within the same cluster exhibit similar gradients during training, thereby affecting the model similarly. Consequently, %
subsets sampled from these clusters have a bounded gradient error w.r.t. the full data, allowing for training a comparable model with only a subset of data. 
Furthermore, we provide a convergence rate analysis for training on these subsets, establishing a robust theoretical foundation for {\alg}'s effectiveness and efficiency.}

To validate {\alg}'s effectiveness, we applied it to the challenging tasks of SFT for (1) mathematical problem-solving and (2) clinical text summarization. 
Our experiments on MathInstruct \cite{yue2023mammoth} 
shows that {\alg} can significantly reduce the required training data size to just 11\% of the original dataset size while still matching the performance levels of the full dataset, outperforming current state-of-the-art one-shot and online data selection algorithms by an average of 4.7\% across 6 in- and out-domain evaluation datasets. 
Remarkably, on the MATH benchmark \cite{hendrycks2021measuring}, {\alg} attained a 32.7\% accuracy with just 50K data points, improving the best open-sourced model under 3 billion parameters, Phi-2, by 16.6\%. For clinical text summarization tasks on the MIMIC-III \cite{johnson2016mimic} dataset, {\alg} outperforms training on the full dataset, using only half of the data.
Unlike existing methods that require training and getting features from large models, {\alg} achieves superior data efficiency using a model with as few as 70 million parameters, which is $100\times$ smaller than the largest target model we train with 7 billion parameters. 

%% file: sections/2_related.tex
\section{Related Work}

\textbf{Foundations of Data Selection.} Data selection has been well studied for small models and classification tasks. There are one-shot algorithms that select data based on rankings of the proposed training statistics, for example, the L2-norms of error and gradient vectors (EL2N and GraNd) \cite{paul2021deep}, confidence and its variability across epochs \cite{swayamdipta-etal-2020-dataset}, and the number of times each example is learned but then forgot at the subsequent training step \cite{toneva2018an}. Besides these heuristic indicators, there are embedding-based pruning algorithms \cite{sorscher2022beyond} and online selection algorithms with theoretical performance guarantees for efficiency \cite{pmlr-v119-mirzasoleiman20a,killamsetty2021grad,killamsetty2021glister,pmlr-v162-pooladzandi22a,pmlr-v202-yang23g} and robustness \cite{pmlr-v162-yang22j,yang2023identifying,deng2023robust}. \citeauthor{Coleman2020Selection} proposed to use the intermediate feature representation of a small proxy model to select data for image classification. Most recently, data selection has shown great potential in more substantial training speedup when implemented on near-storage hardware \cite{nessa}, and data selection beyond supervised learning of image data, e.g., for self-supervised learning \cite{joshi2023data} and multimodal learning \cite{abbas2023semdedup,mahmoud2023sieve}, also emerged.

\textbf{Data Efficient Training of Large Language Models.}
For the pre-training of LLMs, \citeauthor{marion2023less} studied data quality indicators including Perplexity,  Error L2-Norm (EL2N) \cite{paul2021deep}, and memorization ranking \cite{biderman2023emergent},
and found training on examples with middle Perplexity rankings outperforms training on examples selected based on the other two metrics, and sometimes even outperforms training on the entire dataset. \citeauthor{tirumala2023d} uses pre-trained model embeddings to select data for LLM pre-training. The proposed algorithm, D4, first applies an embedding-based data de-duplication algorithm \cite{abbas2023semdedup} and then discards data points that are the closest to the K-Means cluster centroids in the embedding space \cite{sorscher2022beyond} to ensure diversity. 
On fine-tuning LLMs, existing work on data efficiency primarily focused on manually curating high-quality instructions \cite{zhou2023lima}, or using strong closed-source models (e.g., GPT-4 \cite{Achiam2023GPT4TR} or ChatGPT) to rate the quality of each training example \cite{eldan2023tinystories,li2023textbooks,chen2023alpagasus}. 
\citeauthor{bhatt2024experimental} implemented an experimental design framework to evaluate the existing data selection methods for instruction fine-tuning of LLMs and found selecting facility locations based on hidden representations (i.e., embeddings) is the most effective.
As the only data selection algorithm for specialized domains, SCIP \cite{yang2023decoding} focuses on pruning low-quality code data for training code LLMs. Since it relies on breaking the code syntax to understand the characteristics of low-quality code in the embedding (i.e, hidden states) space, adapting SCIP to domains other than Python code data is non-trivial.

\textbf{Adapting Large Language Models for Specialized Domains.}
The rapid development of large language models (LLMs) gives rise to new state-of-the-art models in specialized domains. 
For mathematical reasoning, Galactica \cite{taylor2022galactica}, MINERVA \cite{lewkowycz2022solving} and Llemma \cite{azerbayev2023llemma} continue to train an LLM on large-scale math-related web data to improve a model’s general scientific reasoning; WizardMath \cite{luo2023wizardmath} and TinyGSM \cite{liu2023tinygsm} fine-tune LLMs using supervised data. Similarly for medical LLMs, \citeauthor{cheng2023adapting} continued training pre-trained LLMs on medical text, and \cite{singhal2023medpalm,singhal2023medpalm2} fine-tuned PaLM with instruction prompt tuning on medical domain data.

%% file: sections/3_method.tex
\section{Problem Formulation}
\textbf{LLM Fine-tuning Objective.}
Consider a transformer-based language model, parameterized by $\bm{\theta}$, and denoted as $p_{\bm{\theta}}$. 
This model, when provided with a sequence of prompt tokens $\mathbf{x}=(x_1,\ldots,x_M)$, generates a sequence of response tokens $\mathbf{y}=(y_1,\ldots,y_L)$. 
The conditional probability of generating $\mathbf{y}$ given $\mathbf{x}$ is then formulated as
\begin{align}
    p_{\bm{\theta}}(\mathbf{y}|\mathbf{x})=\prod_{l=1}^{L} p_{\bm{\theta}}(y_l|\mathbf{y}_{1:l-1},\mathbf{x}). 
\end{align}
Note that $\mathbf{y}_{1:0}$ is an empty sequence. To adapt the pre-trained LLM for a specialized domain of distribution $\mathcal{D}$, supervised fine-tuning (SFT) is usually employed with a domain-specific training dataset $D_{\text{train}}=\{(\mathbf{x},\mathbf{y})_i\}_{i=1}^n \sim \mathcal{D}$ containing pairs of prompt $\mathbf{x}$ and annotated response $\mathbf{y}$. 
The fine-tuning objective is thus to minimize the following negative log likelihood loss, expressed as: 
\begin{align}\label{eq:loss}
    \min_{\bm{\theta}} \mathcal{L}(\bm{\theta}, D_{\text{train}} ) = - \frac{1}{n}\sum_{(\mathbf{x},\mathbf{y})_i\in D_{\text{train}}}\big[\log p_{\bm{\theta}}(\mathbf{y}_i|\mathbf{x}_i)\big]. 
\end{align}

\textbf{Data Selection Objective.}
In a general setting for data selection, we consider
a target language model $p_{\bm{\theta}}$ with parameters \( \bm{\theta} \). Given a fixed data budget \( B \), which constrains the number of data points that can be used for training, our objective is to select a subset \( S \subseteq D_{\text{train}} \) 
to train the target model, such that it obtains a superior generalization performance.
In practice, the subset \( S \) is selected based on a reference model \( r_{\bm{\phi}} \) parameterized by \( \bm{\phi} \), which generates representations, confidence scores, or other metrics for each data point \( (\mathbf{x},\mathbf{y})_i \in D_\text{train} \), denoted by \( r_{\bm{\phi}}((\mathbf{x},\mathbf{y})_i) \), which will be utilized by a data selection algorithm to produce \( S \).

In existing data selection algorithms, $\bm{\phi}$ is commonly either weights of the pre-trained target model or a target model that has been fully trained on the dataset \( D_\text{train} \). 
However, as evidenced by \cref{fig:embed}, representations generated by the pretrained model may not always be good enough for data selection in specialized domains, and fine-tuning the target model significantly increases the computational cost of data selection.

\section{Methodology}\label{sec:method}

Training a large target model to obtain feature representations for each example in \( D_{\text{train}} \) can be computationally intensive. However, a recent finding demonstrates that the training dynamics of most examples are consistent across differently sized models of the same family, and this phenomena even generalizes across different model families  \citep{xia-etal-2023-training}. Our proposed method, \textbf{{\fullname} ({\alg)}}, %
leverages \textit{loss trajectories} of training examples collected during fine-tuning a \textit{small} reference model on the full or a subset of training data.

\paragraph{Loss Trajectory.} Let $\bm{\phi}^{(t)}$ be the parameters of a small LM during training on $D_\text{train}$ at times $t_q, q\in \{1,...,T\}$.
{${\alg}$ records the loss trajectory for each data point $i$ at times $t_q$} during training the reference model 
{$[\mathcal{L}_i^{\text{proxy}}(\bm{\phi}^{(t_1)}), \ldots, \mathcal{L}_i^{\text{proxy}}(\bm{\phi}^{(t_T)})]$}  where
\begin{align}
\begin{split}
    \mathcal{L}_i^{\text{proxy}}(\bm{\phi}^{(t)}) = \mathcal{L}^{\text{proxy}}(\bm{\phi}^{(t)}, (\mathbf{x}_i, \mathbf{y}_i)) 
    = -\log p_{\bm{\phi}^{(t)}}(\mathbf{y}_i | \mathbf{x}_i), 
\end{split}
\end{align}
and $T$ is the length of the loss trajectory. 
Note that $\bm{\phi}^{(t)}$ is trained for a fixed number of iterations from $\bm{\phi}^{(t-1)}$.

Assume the parameter vector \( \bm{\theta}^{(t)} \) represents the parameters of the target model at the time $t$. Define \( \mathbf{L}_i^{\text{proxy}} = [\mathcal{L}_i^{\text{proxy}}(\bm{\phi}^{(t_1)}), \ldots, \mathcal{L}_i^{\text{proxy}}(\bm{\phi}^{(t_T)})] \) and \( \mathbf{L}_i^{\text{target}} = [\mathcal{L}_i^{\text{target}}(\bm{\theta}^{(t_1)}), \ldots, \mathcal{L}_i^{\text{target}}(\bm{\theta}^{(t_T)})] \) as the training loss trajectory of the example 
$i$ on the small proxy model and the large target model, respectively. Let \( \bm{H}_i \in \mathbb{R}^{d \times d} \) be the Hessian matrix for each example \(i\) and assume that the loss function for each example during fine-tuning can be modeled by a second-order Taylor approximation with bounded curvature (\( c \leq \| \bm{H}_i \| \leq C \)), a reasonable assumption in fine-tuning settings. The following lemma shows that examples with similar loss trajectories on the proxy model have similar gradients throughout the training of the target model.

\begin{theorem}\label{lemma}
{
If examples \( i \) and \( j \) have similar loss trajectories on the proxy model, i.e., \( \| \mathbf{L}_i^{\text{proxy}} - \mathbf{L}_j^{\text{proxy}} \| \leq \epsilon \), and 
their loss trajectories on the proxy and target model is similar, %
i.e., \( \| \mathbf{L}_p^{\text{proxy}} - \mathbf{L}_p^{\text{target}} \| \leq \delta \) for $p\in\{i,j\}$, then \( i \) and \( j \) have similar gradients throughout training the target model: 
\begin{align}
\| \nabla \mathcal{L}_i^{\text{target}}(\bm{\theta}) - \nabla \mathcal{L}_j^{\text{target}}(\bm{\theta}) \| \leq \frac{2\epsilon' + 2CD^2}{d}=\Delta.
\end{align}
where \( \epsilon' = \epsilon + 2\delta \) and \( \|\bm{\theta}\| \leq D \) for all \( t \).
}
\end{theorem}

The proof of \cref{lemma} can be found in \cref{proof:lemma}. 
\cref{lemma} shows that examples 
with similar loss trajectories
have similar gradients during the training, thereby influencing the model in a similar manner. 
\begin{figure*}[ht!]
\begin{minipage}[b]{0.49\textwidth}
\begin{center}
     \begin{subfigure}[b]{0.45\columnwidth}
         \centering
         \includegraphics[width=\textwidth]{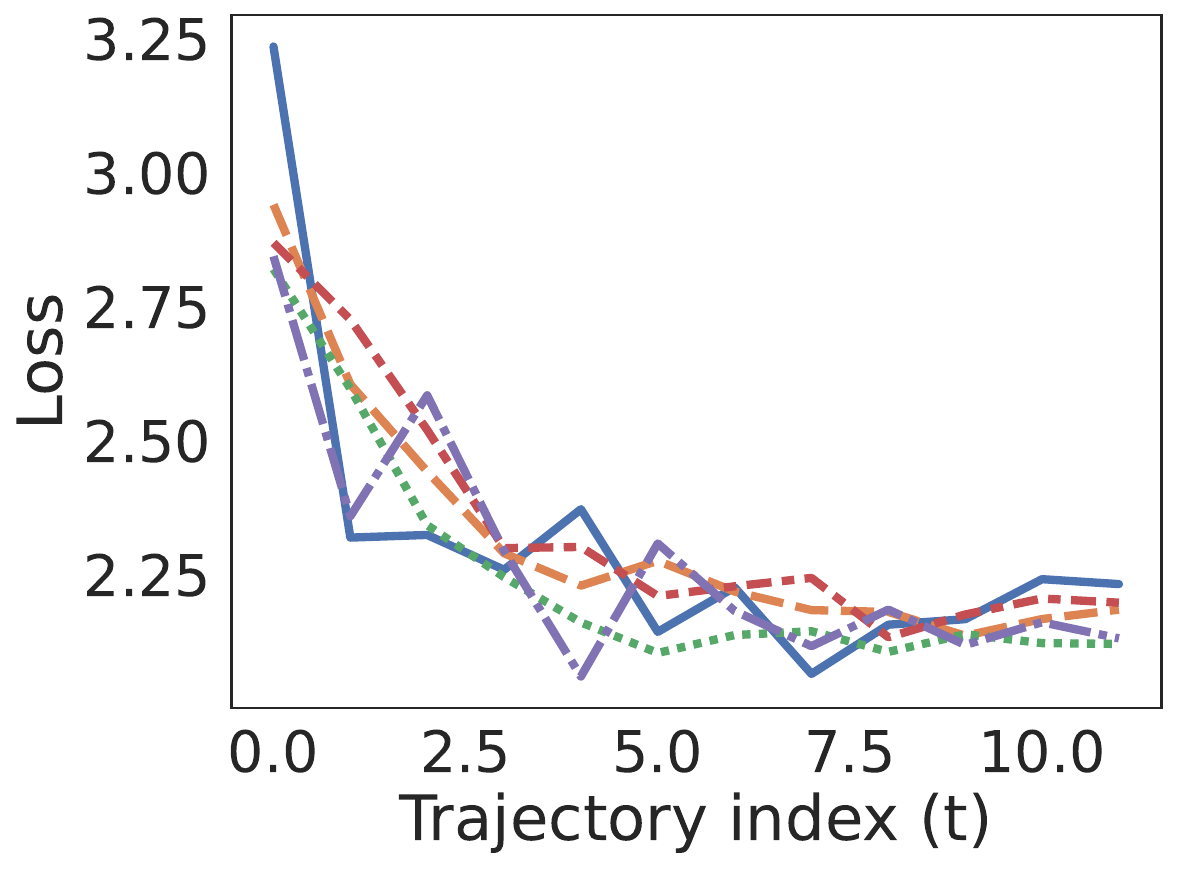}
         \caption{In the same cluster.}
         \label{fig:same-cluster}
     \end{subfigure}
     \begin{subfigure}[b]{0.42\columnwidth}
         \centering
         \includegraphics[width=\textwidth]{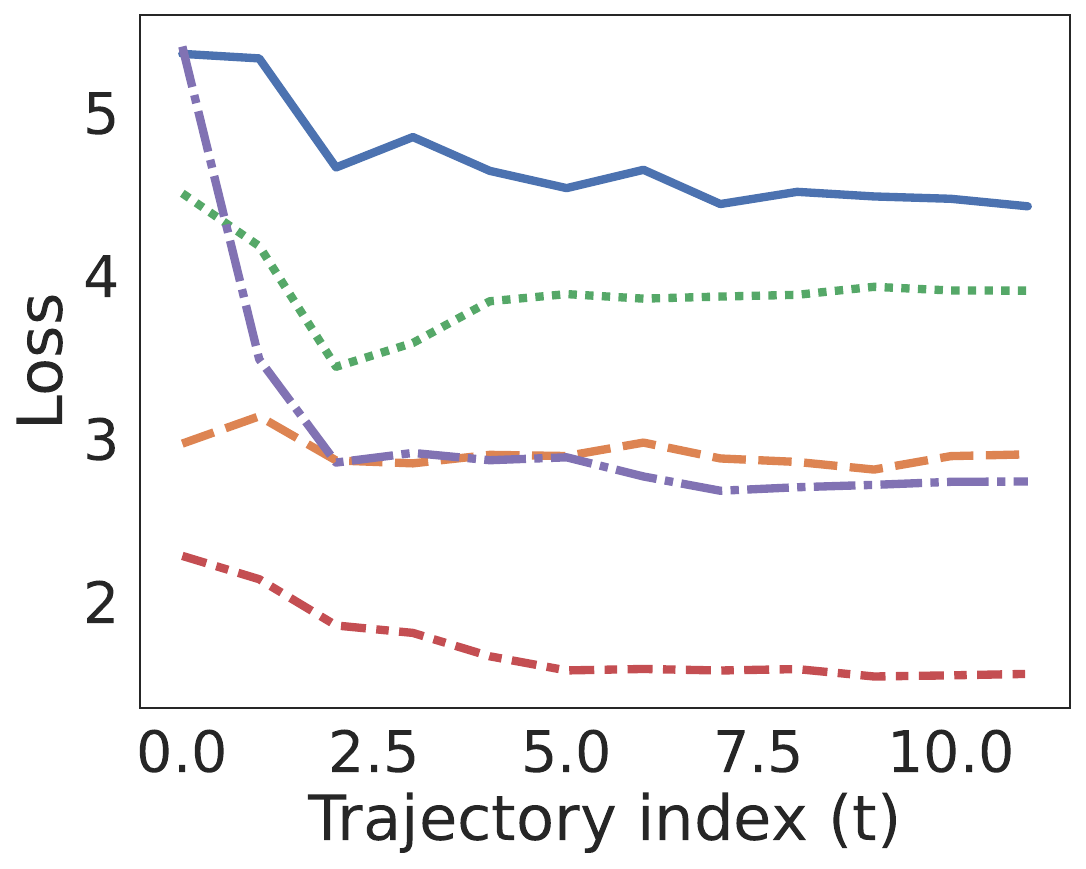}
         \caption{In different clusters.}
         \label{fig:diff-cluster}
     \end{subfigure}
\caption{Examples in the same clusters have very similar loss trajectories (\cref{fig:same-cluster}) while the loss trajectories of examples in different clusters are very different (\cref{fig:diff-cluster}). }
\label{fig:cluster-loss}
\end{center}
\end{minipage}
\hfill
\begin{minipage}[b]{0.49\textwidth}
\begin{center}
     \begin{subfigure}[b]{0.30\columnwidth}
         \centering
         \includegraphics[width=\textwidth]{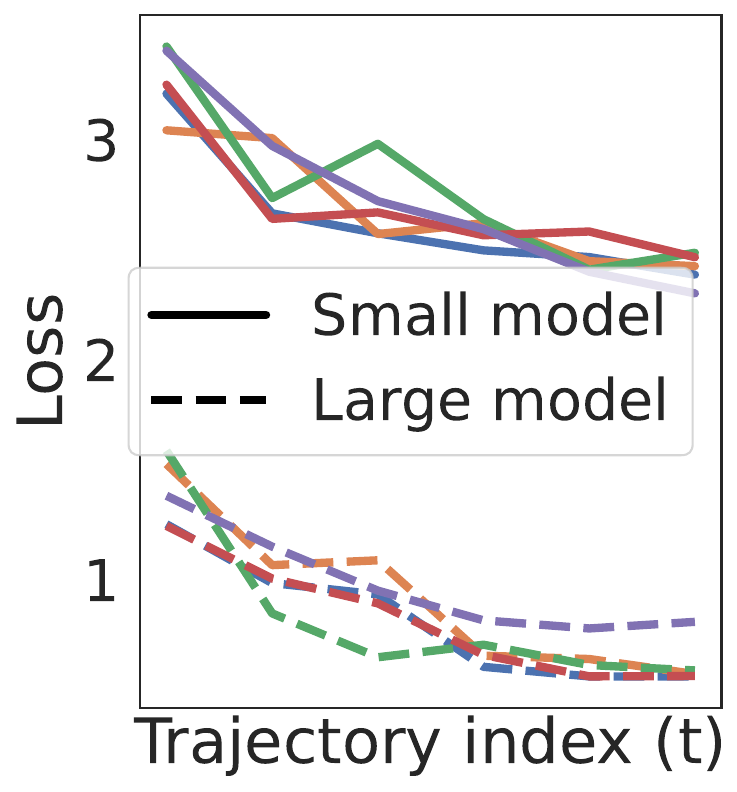}
         \vskip -0.05in
         \caption{}
         \label{fig:down}
     \end{subfigure}
     \begin{subfigure}[b]{0.34\columnwidth}
         \centering
         \includegraphics[width=\textwidth]{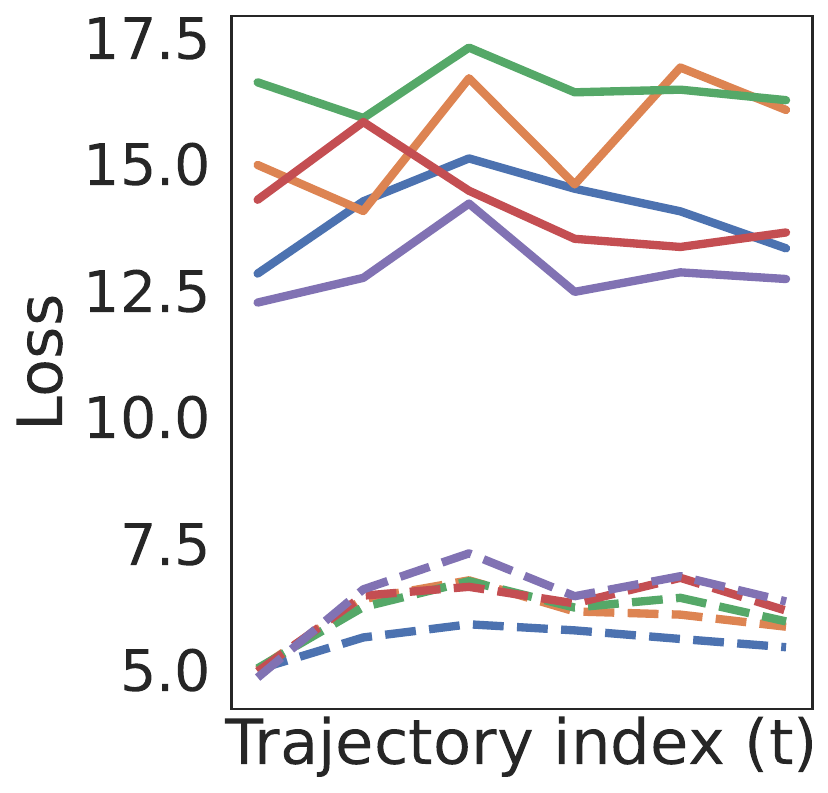}
         \vskip -0.05in
         \caption{}
         \label{fig:up}
     \end{subfigure}
     \begin{subfigure}[b]{0.30\columnwidth}
         \centering
         \includegraphics[width=\textwidth]{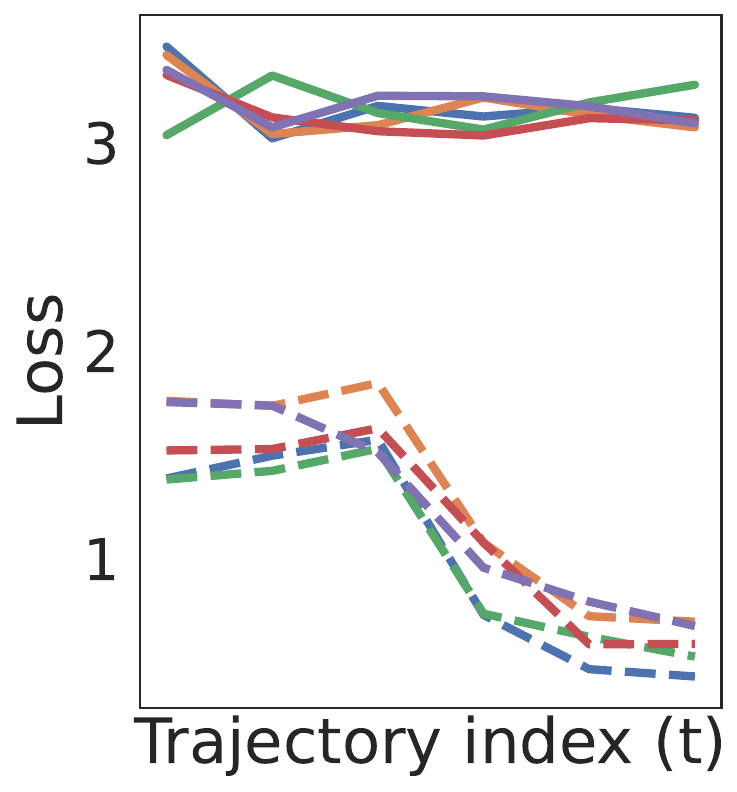}
         \vskip -0.05in
         \caption{}
         \label{fig:double}
     \end{subfigure}
\vskip -0.05in
\caption{Examples in the same clusters of training trajectories on a small model (Pythia-70M) also have similar training trajectories on a large model (Pythia-2.8B), even if the trends may not be the same on both models. 
}
\label{fig:loss-trend}
\end{center}
\end{minipage}
\end{figure*}
\paragraph{Data selection from Loss Trajectory Clusters.} Once the loss trajectories are recorded on the proxy model, we apply a clustering algorithm to group examples based on the similarity of their loss trajectories. This results in a set of clusters \( \{C_1, C_2, \ldots, C_K\} \), where each cluster \( C_i \) contains examples with similar loss and gradient trajectory throughout the training: 
\begin{align}\label{eq:clusters}
\begin{split}
C_i = \{ (\mathbf{x},\mathbf{y})_j\in D_{\text{train}} | i = \arg\min_{j\in[K]} d(\mathbf{L}_{j}, \mathbf{L}_{\bar{C_j}}&
) \},
\end{split}
\end{align}
where $\mathbf{L}_{\bar{C_i}}$ is the centroid of the loss trajectories in cluster $C_i$, and $d(\cdot, \cdot)$ is a distance metric, such as Euclidean distance, used for clustering.
For datasets that contain different sources of data, we cluster each source separately.

\begin{wrapfigure}{L}{0.56\textwidth}
    \begin{minipage}{0.56\textwidth}
    \vspace{-1em}
\begin{algorithm}[H]
\caption{Data Selection Based on Training Trajectories (\alg)}
\label{alg:select}
\begin{algorithmic}[1]
\REQUIRE 
Training dataset \( D_{\text{train}} \) with corresponding training trajectories, a fixed data budget \( B \), number of clusters $K$.
\ENSURE Subset \( S \subseteq D_{\text{train}}, |S|\leq B \). %
\STATE Initialize \( S \) as an empty set.
\STATE Train a small proxy model and cluster examples in (each data source of) \( D_{\text{train}} \) based on their loss trajectories and sort them by size to get \( \mathcal{C} = \{C_1, C_2, \ldots, C_K\} \).\looseness=-1
\FOR{each cluster \( C_k \) in \( \mathcal{C} \)}
    \STATE Calculate \( R_k \), the number of examples to randomly sample from \( C_k \), i.e., $R_k = (B - |S|) / (K-k+1)$.
    \IF{\( |C_k| \leq R_k \)}
        \STATE $S \leftarrow \{S \bigcup C_k\}$.
    \ELSE
        \STATE $S \leftarrow \{S \bigcup S_k$\}, where $S_k \subset C_k$ is selected uniformly at random from $C_k$ and $|S_k|=R_k$
    \ENDIF
\ENDFOR
\STATE Return \( S \)
\end{algorithmic}
\end{algorithm} 
\vspace{-2em}
\end{minipage}
\end{wrapfigure}

As shown in \cref{fig:cluster-loss}, clustering algorithms can effectively find groups of examples with similar training dynamics. In \cref{fig:loss-trend}, we empirically show that we can identify groups of examples with similar training dynamics on a larger model by clustering the training trajectories of \( D_{\text{train}} \) on a smaller proxy model. 
With the clusters formed, the data selection strategy %
selects equal number of examples at random from all clusters, as detailed in \cref{alg:select}. 
In doing so, it effectively prioritizes selecting examples from smaller clusters. This is particularly important for datasets containing multiple imbalanced sources. In this setting, training and test distributions often differ, and balanced selection from clusters ensures superior test performance on all groups in the test data.

The following theorem shows that, under the assumptions of \cref{lemma}, training with Incremental Gradient (IG) methods on the subset selected by {\alg} converges to a close neighborhood of the optimal solution found by training the target model on the full dataset. IG methods such as Stochastic Gradient Descent (SGD) update parameters iteratively based on the gradient of the loss of individual examples, multiplied by stepsize $\alpha$. Formally,
\begin{align}
    \bm{\theta}^{t+1} = \bm{\theta}^{t} - \alpha^{} \nabla \mathcal{L}_{i}^{\text{target}}(\bm{\theta}^{t}).
\end{align}

\begin{corollary}\label{theorem}
Under the assumptions of \cref{lemma}, applying IG with stepsize %
$\alpha$ to subsets found by S2L, converges to the neighborhood of the optimal solution, as follows: 
\begin{equation}
    \|\bm{\theta}^{t+1}-\bm{\theta}^*\|^2\leq (1-\alpha c)^{t+1} \|\bm{\theta}^{t}-\bm{\theta}^*\|^2 + 2\xi R/c^2+\alpha B^2 (r_{\min}/k)^2 \bm{g}_{\max}^2
\end{equation}
where $c\leq \|\bm{H}\|$, $B=k\cdot K$ is the total size of the subset, $\bm{g}_{\max}$ is the largest gradient norm of individual examples during training, $r_{\min}=\min_j |C_j|, r_{\max}=\max_j|C_j|$, 
$R = \min \{d_0, B \bm{g}_{\max} + \xi / c \}$ and $d_0 = \|\bm{\theta}^0 - \bm{\theta}^*\|$ is the initial distance to the optimal solution $\bm{\theta}^*$, and \(\xi\) is given by:
\begin{align}
    \xi = K [r_{\min}\Delta + (r_{\max}-r_{\min})\bm{g}_{\max}].
\end{align}

\end{corollary}

The proof can be found in \cref{proof:theorem}.

%% file: sections/4_experiments.tex
\section{Experiments}\label{sec:experiments}
In this section, we present the comprehensive experiments conducted to evaluate the efficacy of the proposed data selection method, {\fullname} ({\alg)}, across two challenging domains (mathematical reasoning and clinical text summarization).

\subsection{Baselines}\label{sec:math-baseline}
We systematically compare {\alg} against a comprehensive set of open-sourced data selection methods. These methods are categorized based on the type of representation they use and selected as the most representative or best-performing methods as identified in prior work. These include: (1) \textbf{Random Sampling}; selecting examples with the (2) \textbf{Least Confidence} \cite{bhatt2024experimental} or (3)\textbf{Middle Perplexity} \cite{marion2023less}; (4) \textbf{High Learnability}, determined by the loss decrease before and after full fine-tuning \cite{zhou2023lobass}; and (5) \textbf{Facility Locations} selection based on hidden states \cite{bhatt2024experimental}. Additionally, we incorporate one online selection techniques: (6) 
\textbf{Confidence Curriculum} proposed by \citeauthor{varshney-mishra-and-chitta-baral-2022-model}, which 
selects examples with decreasing confidence during the training.
Given that the optimal reference model may vary for each one-shot selection method, we ensure a fair comparison by adopting the approach used in \cite{marion2023less}, which runs each method with both the fully fine-tuned target model and an early fine-tuning checkpoint as the reference model. We report the best results from these setups.

\subsection{Specialized Domain 1: Mathematical Reasoning}\label{sec:math}

\textbf{Training Settings.}\label{sec:math-train}
We focus on fine-tuning using the \textbf{MathInstruct} dataset \cite{yue2023mammoth} with 262,040 training examples for its comprehensive coverage of diverse mathematical fields and its capability in training models to achieve state-of-the-art performance on the standard evaluation benchmarks. We employ the open-source model suites Pythia \cite{biderman2023pythia}, Phi-2 \cite{textbooks2}, Llama-2 \cite{touvron2023llama} as our base models to validate our {\alg} algorithm and directly compare its performance against the state-of-the-art.

\textbf{Evaluation Datasets.} We follow the framework established in \cite{yue2023mammoth} for a comprehensive assessment using several well-regarded datasets, including in-domain and out-of-domain datasets.
For the in-domain datasets, we consider \textbf{GSM8K} \cite{cobbe2021gsm8k}, \textbf{MATH} \cite{hendrycks2021measuring}, and \textbf{NumGLUE} \cite{mishra-etal-2022-numglue}. For the out-of-domain datasets, we consider \textbf{SVAMP} \cite{patel-etal-2021-nlp}, \textbf{Mathematics} \cite{davies2021advancing}, \textbf{SimulEq} \cite{koncel-kedziorski-etal-2016-mawps}. These datasets collectively span a diverse range of mathematical subjects, such as algebra, probability, number theory, calculus, and geometry. Additionally, some questions in these datasets require the application of commonsense, reading comprehension, and multi-step reasoning. All questions are open-formed.

\textbf{Evaluation Metric.} We use the standard evaluation metric for open-formed questions, \textbf{exact match}, which measures the model's accuracy by comparing its generated answers against the correct solutions. For an answer to be considered correct, it must match the reference solution precisely.  

\new{More details about the settings and baseline implementations can be found in \cref{sec:app-details}.}

\begin{figure}
\vskip -0.05in
\begin{center}
     \begin{subfigure}[b]{0.32\columnwidth}
         \centering
         \includegraphics[width=\textwidth]{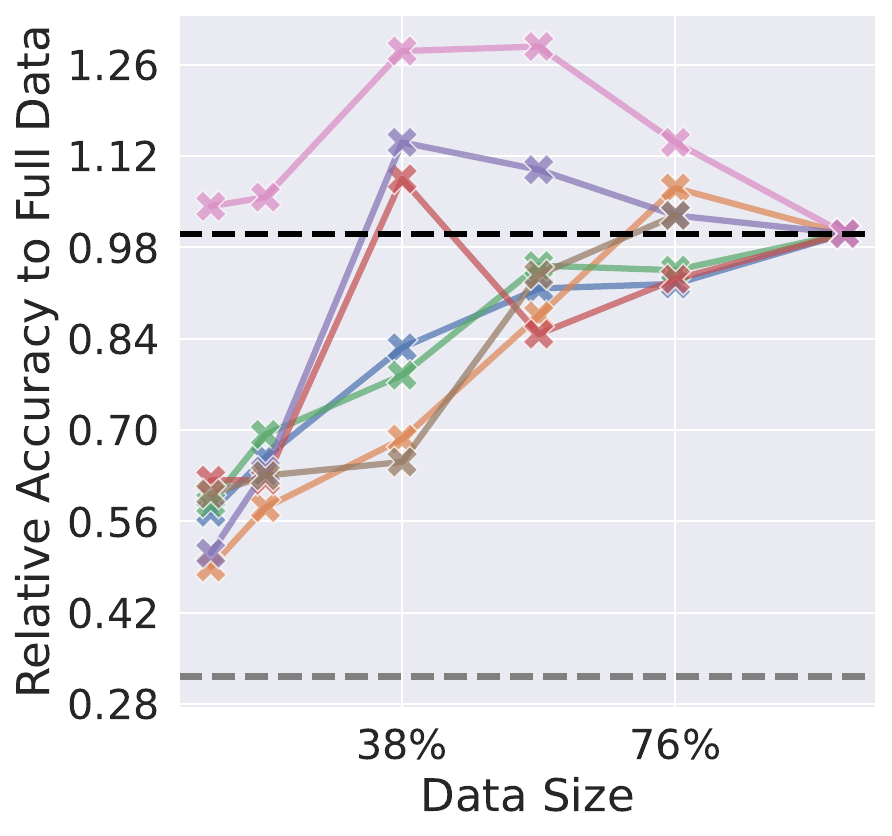}
         \caption{In-domain Avg}
     \end{subfigure}
     \begin{subfigure}[b]{0.54\columnwidth}
         \centering
         \includegraphics[width=\textwidth]{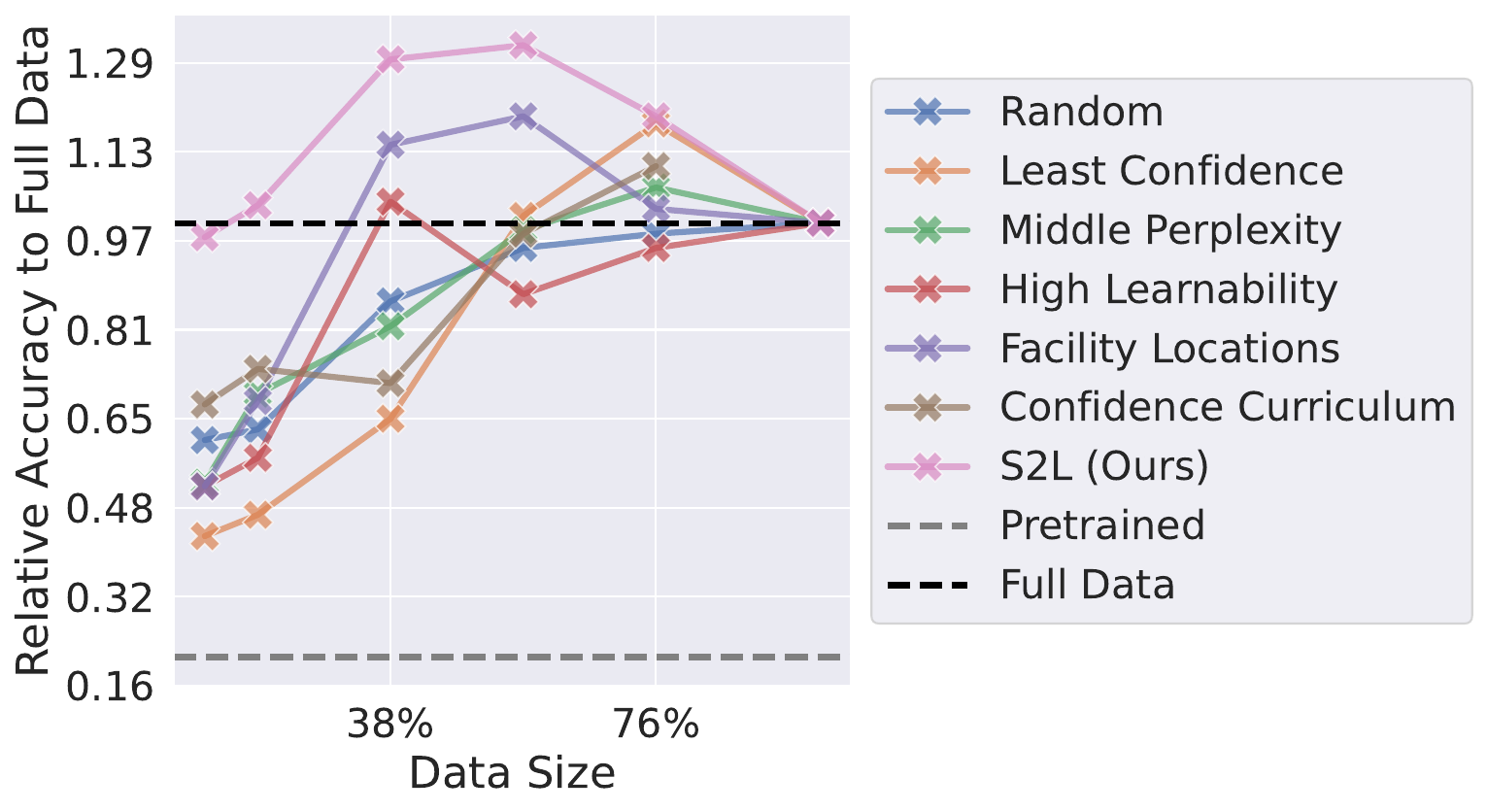}
         \caption{Avg}
     \end{subfigure}
\caption{\textbf{Data Scaling:} Accuracies ($\uparrow$) on in-domain and out-of-domain datasets using Pythia-410M. Data size refers to the total number of unique training examples used. All experiments in this table share the same total training steps and learning rate schedule (see \cref{sec:math-train}). {See breakdowns in \cref{fig:410m-full}.}
}
\label{fig:410m}
\end{center}

\begin{minipage}{\textwidth}
\captionof{table}{\textbf{Less Data, Same Compute:} Zero-shot accuracies (\%, $\uparrow$) \new{when {\alg} and the baselines select 50K data to train with the same number of iterations as the full-data training}. 
Results surpassing full training are highlighted in bold. \cref{fig:410m} follows the same setting but uses the Pythia-410M model.}
\label{tab:math-scale}
\vspace{-1em}
\begin{center}
\begin{small}
\begin{sc}
\resizebox{\textwidth}{!}{%
\begin{tabular}{lc|ccc|b|ccc|b}
\toprule
Target & Fine-tuning & \multicolumn{4}{c|}{In-domain} & \multicolumn{3}{c|}{Out-domain} & \\ 
Model & Data & GSM8K & MATH & NumGLUE & \textbf{Avg} & SVAMP & Mathematics & SimulEq & \textbf{Avg} \\
\midrule
\multirow{3}{*}{Phi-2 (2.7B)} & (Pretrained) & 53.4 & 16.1 & 34.9 & $34.8$ & 67.9 & 31.1 & 27.4 & 38.5\\
    & Random & 67.9 & 30.1 & 60.7 & 52.9 & 77.1	& 51.2 & 37.5	& 54.1 \\
    & High Learnability & 59.4 & 25.2 & 62.1 & 48.9 & 76.6 & 41.8 & 27.2 & 48.7 \\ 
    & Middle Perplexity & 66.4 & 29.5 & 54.1 & 50.0 & 74.8 & 50.4 & 39.8 & 52.5 \\
    & Least confidence & 61.7 & 24.7 & 67.0 & 51.1 & 76.5	& 43.3 & 52.5 & 54.3 \\
    & Facility Locations & 66.2 & 31.3 & 62.4 & 53.3 & 74.4 & 58.4 & 34.6 & 54.5 \\
    & \textbf{\alg(Ours)} &  $\mathbf{69.1}$ & $\mathbf{32.6}$ & $\mathbf{65.7}$ & $\mathbf{55.8}$ & $\mathbf{79.6}$ & 56.4 & 40.1 & 57.3 \\
    & Full-262K & $68.3$ & $32.6$ & $64.3$ & $55.1$ & 78.4 & 58.4 & 44.2 & $\mathbf{57.7}$ \\
\bottomrule
\end{tabular}
}%
\end{sc}
\end{small}
\end{center}
\end{minipage}
\end{figure}

\begin{wrapfigure}{R}{0.3\textwidth}
\begin{center}
\centerline{\includegraphics[width=0.3\columnwidth]{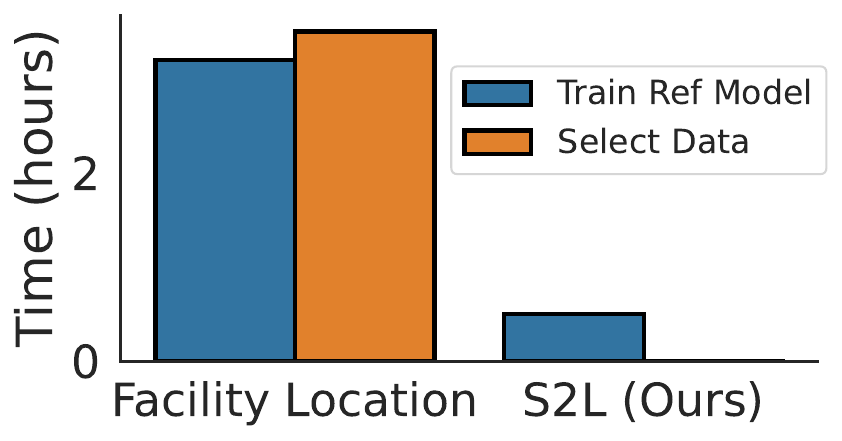}}
\caption{Wall-clock time required to train the reference model and select 100K data from MathInstruct for training Pythia-410M. }
\label{fig:cost}
\end{center}
\vskip -2em
\end{wrapfigure}

\subsubsection{\new{Setting 1: Less Data for Better Models}} \label{sec:math-results}
In the first setting, we standardize the number of training steps to correspond to 3 epochs on the full dataset, aligning with \cite{yue2023mammoth}. This allows us to maintain a consistent training schedule across different methods and data budgets, ensuring fair and accurate comparisons of the quality of data.

\textbf{\textsc{Scaling the Data:} SOTA algorithms fail with small data budgets while {\alg} stands out across data scales.} In \cref{fig:410m}, we compare {\alg} against the baselines from \cref{sec:math-baseline} on Pythia-410M across varying data sizes. The training trajectories used by {\alg} are based on Pythia-70M, a model approximately 6x smaller than Pythia-410M. With the same number of training steps as the full training, {\alg} exceeds the full dataset's performance using only 30K examples, only 11\% of the full dataset. 
It leads the runner-up baselines by an average of 4.7\%, 4.6\% and 2.4\% with data budget 30K, 50K and 100K across all six evaluation datasets. While state-of-the-art data selection algorithms like Facility Locations \cite{bhatt2024experimental} and High Learnability \cite{zhou2023lobass} have decent performance with a large enough data budget (i.e., 100K), all SOTA algorithms except {\alg} cannot even outperform the random sampling baseline when the allowed data size is small (i.e., 30K). Unlike the existing algorithms, {\alg} consistently outperforms all baselines and even full training across all data sizes. Note that compared to the runner-up algorithm in this setting, Facility Locations, the cost of {\alg} is much lower in both training the reference model and data selection stages (\cref{fig:cost}), and therefore more scalable to both larger target models or larger data sizes.

\textbf{\textsc{Scaling the Model:} Data selected by {\alg} can transfer to larger models in different model suites.} We also 
test whether this subset, chosen using Pythia-70M, can effectively train larger models beyond 410M and models outside the Pythia suite. 
As shown in \cref{tab:math-scale}, our experiments with Phi-2
reveal that fine-tuning on only 50K {\alg}-selected data again outperforms full dataset training on the most challenging MATH \cite{hendrycks2021measuring} benchmark
improving the pretrained Phi-2 by $16.6\%$ and is more data efficient than training on the full MathInstruct dataset to get the same performance.

\begin{wrapfigure}{R}{0.4\textwidth}
\vspace{-1em}
\begin{center}
\centerline{\includegraphics[width=0.4\textwidth]{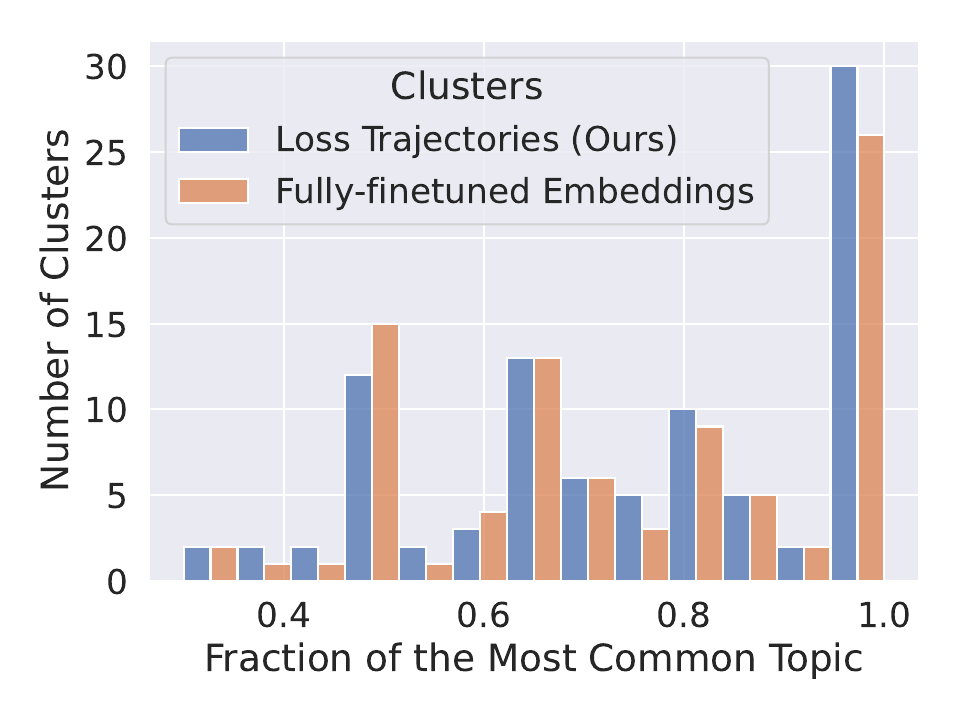}}
\vskip -0.05in
\caption{Distribution of the coverage of top-1 topic in each cluster. Taller bars on the right end of the plot indicate clusters with a higher concentration of a single topic and therefore suggest a grouping of similar examples.}
\label{fig:purity}
\end{center}
\vspace{-2em}
\end{wrapfigure}

\subsubsection{\new{Setting 2: Less Data for Faster Training}} The second setting we consider is when fixing the number of times each example can be seen over the entire course of training, directly translating smaller datasets into reduced training and storage costs. This is particularly beneficial for large models that would require extensive training times without data selection. By experimenting with models of larger sizes than the previous setting, we observe in \cref{tab:math-epoch} that {\alg} can achieve comparable performance to full-data training when using only 50\% data and thereby reducing both the data storage space and the training time by half. 

\begin{table*}[t]
\caption{\textbf{Less Data, Same Epochs:} Zero-shot accuracies (\%, $\uparrow$) \new{when {\alg} trains 50\% data for the same number of epochs as the full-data training}. 
\new{{\alg} can achieve comparable performance to full-data training while reducing both the data storage space and the training time by half.}\looseness=-1
}
\label{tab:math-epoch}
\vspace{-1em}
\begin{center}
\begin{small}
\begin{sc}
\resizebox{\textwidth}{!}{%
\begin{tabular}{lc|ccc|b|ccc|b}
\toprule
Target & Fine-tuning & \multicolumn{4}{c|}{In-domain} & \multicolumn{3}{c|}{Out-domain} & \\ 
Model & Data & GSM8K & MATH & NumGLUE & \textbf{Avg} & SVAMP & Mathematics & SimulEq & \textbf{Avg} \\
\midrule
\multirow{3}{*}{Phi-3-mini (3.8B)} & (Pretrained) & 74.5 & 26.5 & 52.1 & 51.1 & 83.7 & 44.3 & 34.8 & 52.7\\
            & \textbf{\alg-50\%(Ours)} & 76.3 & 42.5 & 76.4 & 65.1 & 83.8 & 62.1 & 51.6 & 65.4\\
            & Full & 76.4 & 42.9 & 75.3 & 64.9 & 84.6 & 60.2 & 51.9 & 65.2\\
\midrule
\multirow{3}{*}{Llama-2-7B} & (Pretrained) & 3.1 & 4.2 & 16.5 & 7.9 & 14.1 & 8.3 & 2.3 & 8.1\\
& \textbf{\alg-50\%(Ours)} & 53.3 & 28.9 & 65.0 & 49.1 & 65.1 & 45.2 & 31.9 & 48.2\\
& Full-262K \cite{yue2023mammoth} & 52.2 & 30.4 & 60.5 & 47.7	& 65.3 & 43.9 & 50.2 & 50.4 \\
\bottomrule
\end{tabular}
}%
\end{sc}
\end{small}
\end{center}
\vspace{-1em}
\end{table*}

\subsubsection{\new{Why is {\alg} So Effective?}} 
\textbf{Examples in Clusters Encode the Same Knowledge/Skill.} In \cref{sec:examples-full}, we compare actual training examples in MathInstruct that get clustered together due to their similar training trajectories on the small Pythia-70M model. We observe that examples in the same cluster are of the same type and related to the same knowledge/skill, e.g., open-formed algebra questions (\cref{fig:ex-down-full}), examples requiring extracting useful information from long text and writing programs (\cref{fig:ex-up-full}), and multiple choice questions that require multi-step reasoning (\cref{fig:ex-double-full}), etc. Therefore, by sampling from different clusters, we make sure the selected examples cover the knowledge required for all topics and skills required for all types of questions.

\textbf{Loss Trajectories can Capture the Similarity Between Data Points As Much As Embeddings of a Fully Fine-tuned Model.} We conducted a quantitative analysis to assess how effectively {\alg} identifies similar examples using loss trajectories from a small model. Assuming math problems under the same topic require similar knowledge and share question formats, we used unknown topic labels during {\alg}'s data selection to check if each cluster predominantly contains a single topic. By calculating the fraction of the most common topic in each cluster and plotting this in \cref{fig:purity} (with K=100, excluding clusters of size one), we compared the loss trajectory clusters from {\alg} (in blue) against those from the embeddings of a fully fine-tuned Phi-2 model (in orange)—considered the ground truth for similarity. Results show that most clusters formed by {\alg} using the Pythia-70M model are based on a single topic and capture topic similarities more effectively than those from the Phi-2 model's embeddings. This analysis not only confirms the homogeneity within {\alg} clusters but also highlights the computational efficiency of using loss trajectories on small models to identify representative examples. \looseness=-1

\subsection{Specialized Domain 2: Clinical Text Summarization}\label{sec:med}
{\alg} can improve data efficiency not only for fine-tuning data not only in mathematics but also in other specialized domains. This subsection explores its application to clinical text summarization within radiology reports. This task involves processing the detailed analysis and results listed in the findings section of a radiology report and distilling them into a concise impression section. Such summaries are crucial for providing clinicians with quick and actionable insights from radiological studies. \looseness=-1

\textbf{Dataset \& Setup.} We use the \textbf{MIMIC-III} dataset \cite{johnson2016mimic}, a comprehensive collection of radiology reports and findings authored by attending physicians in routine clinical care. 
We use the same preprocessing procedures as \cite{delbrouck-etal-2023-overview,bionlp-2023-biomedical} to extract the findings and impression sections and remove invalid reports. Given that access to MIMIC-III requires specific credentials, we provide a synthetic example of a radiology report generated by GPT-4 \cite{Achiam2023GPT4TR} for illustrative purposes in \cref{tab:report-example}. 
We employ the Pythia-1B model and keep the training setting consistent with the mathematical reasoning task.  %

\begin{wrapfigure}{L}{.49\columnwidth}
\vspace{-1em}
\begin{minipage}{.49\columnwidth}
\begin{center}
     \begin{subfigure}[b]{0.32\textwidth}
         \centering
         \includegraphics[width=\textwidth]{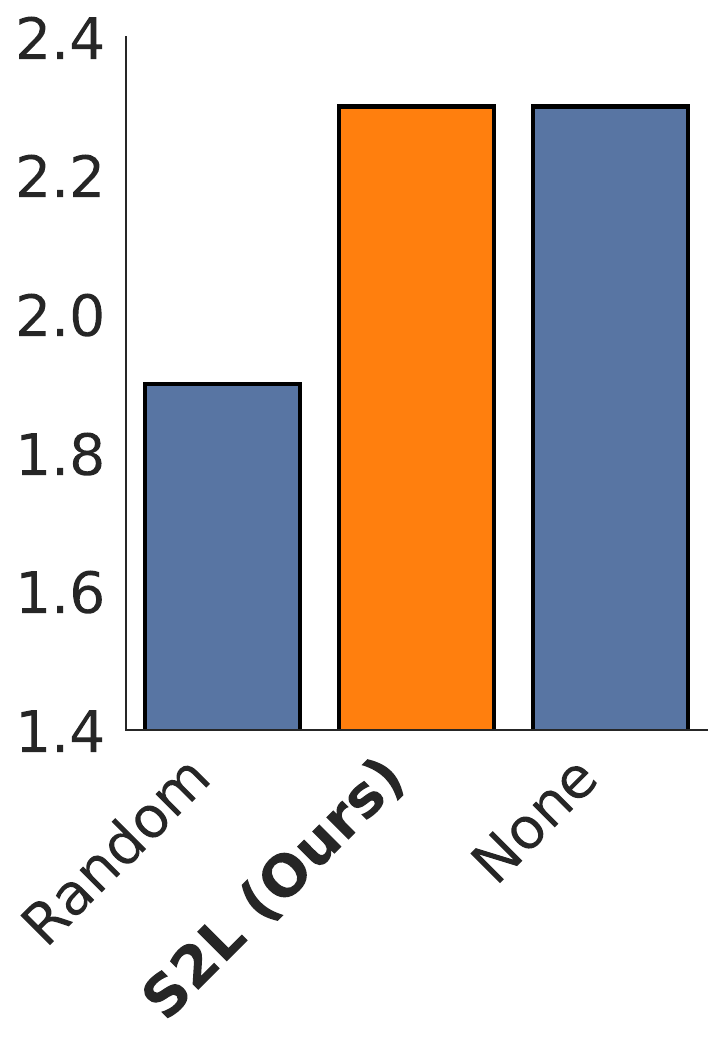}
         \caption{BLEU}
         \label{fig:down}
     \end{subfigure}
     \begin{subfigure}[b]{0.32\textwidth}
         \centering
         \includegraphics[width=\textwidth]{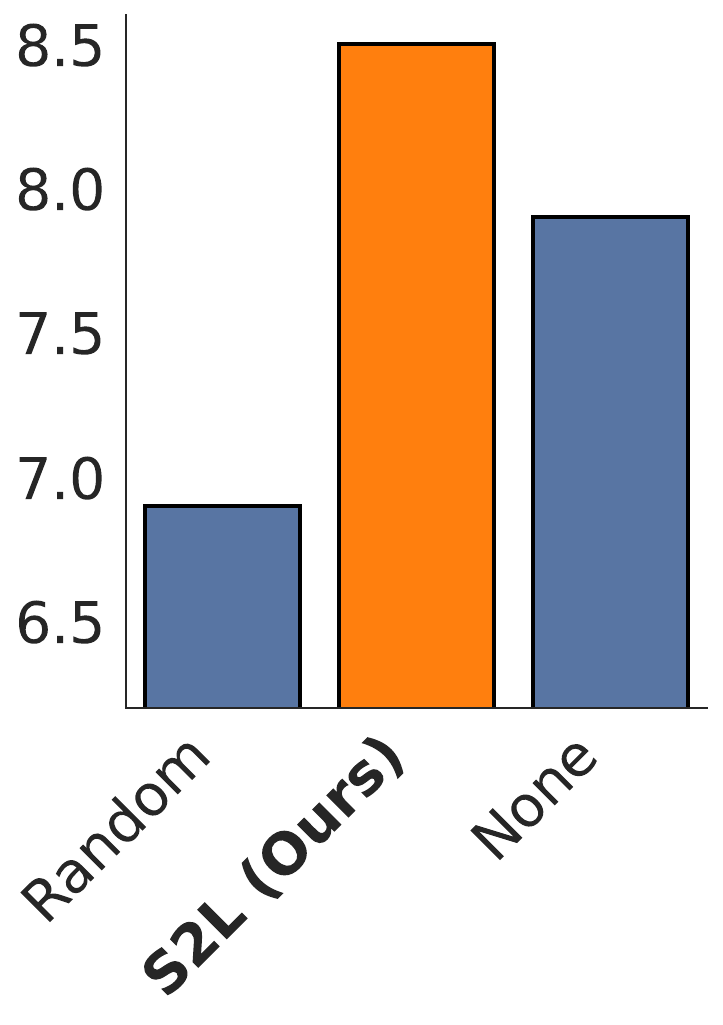}
         \caption{ROUGE-L}
         \label{fig:up}
     \end{subfigure}
     \begin{subfigure}[b]{0.33\textwidth}
         \centering
         \includegraphics[width=\textwidth]{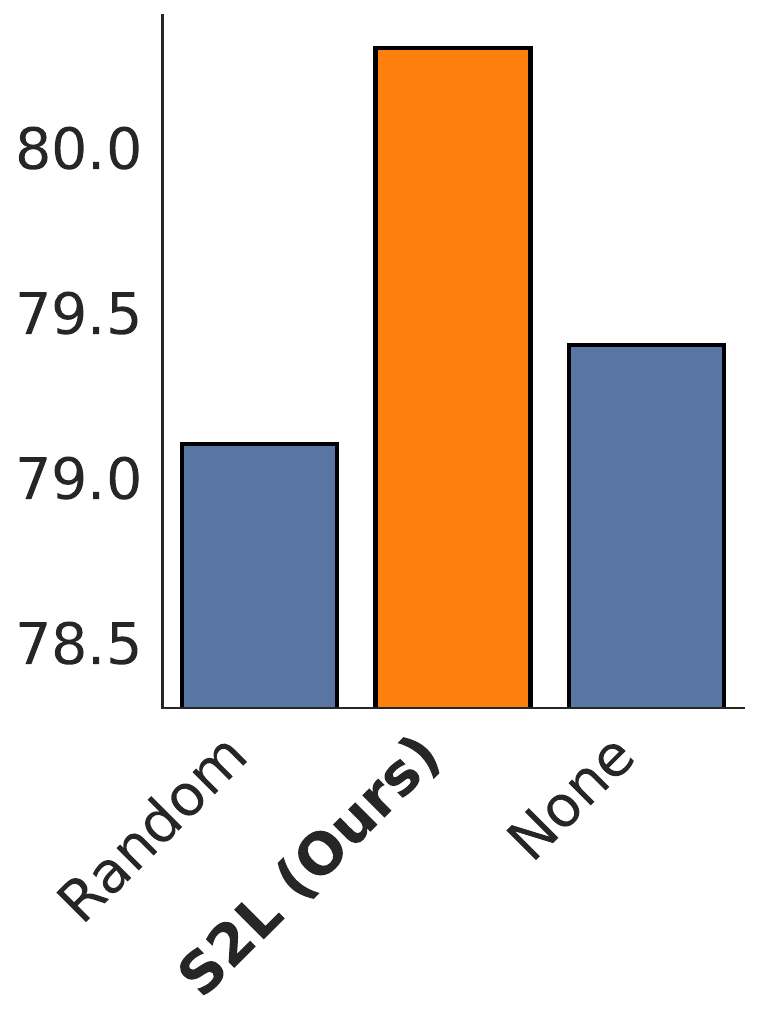}
         \caption{BERTScore}
         \label{fig:double}
     \end{subfigure}
\caption{Performance ($\uparrow$) of models trained on (1) \textbf{random}ly selected 30K examples, (2) \textbf{{\alg}} selected 30K examples, and (3) full 61K examples (\textbf{none}) evaluated with 3 different metrics. The minimum value on the y-axis is the performance of the model before fine-tuning. {\alg} improves the data efficiency for the clinical text summarization task by outperforming training on the full dataset with only less than half of the data. }
\label{fig:mimic}
\end{center}
\end{minipage}
\end{wrapfigure}

\textbf{Evaluation.} Our evaluation of generated clinical summaries on the MIMIC-III dataset's test split employs three key metrics as recommended in \cite{van2023clinical,tu2023towards}: (1) \textbf{BLEU} \cite{papineni2002bleu}, which measures word sequence overlap between the generated and reference texts; (2) \textbf{ROUGE-L} \cite{lin-2004-rouge}, assessing the longest common word sequence; and (3) \textbf{BERTScore} \cite{Zhang2020BERTScore}, evaluating semantic similarity using BERT's contextual embeddings.
These metrics
together offer a comprehensive evaluation, ensuring our summaries are not only precise in language but also meaningful and coherent in the context of clinical information. We compare {\alg} to random selection, a surprisingly strong baseline as evidenced in \cref{sec:math}, to check the validity of the data selection problem on this dataset and then compare it to training on the full dataset to assess its effectiveness.

\textbf{Results.} 
We compare using 30K examples selected by random vs. selected through {\alg}. Even with only half of the data, the model trained with {\alg} selected data achieves similar BLEU and significantly higher ROUGE-L and BERTSCore compared to the model trained on the entire 61.5K data. 
Meanwhile, training on randomly selected 30K examples performs worse than training on the full dataset on all 3 metrics. These results together demonstrate {\alg}'s effectiveness.

\subsection{Ablation Studies}\label{sec:ablation}
We conduct ablation studies on MathInstruct and Pythia-410M to further understand the best practices for using {\alg}. \looseness=-1

\begin{wrapfigure}{R}{.48\columnwidth}
\centering
\begin{minipage}{.23\columnwidth}
  \centering
  \includegraphics[width=\linewidth]{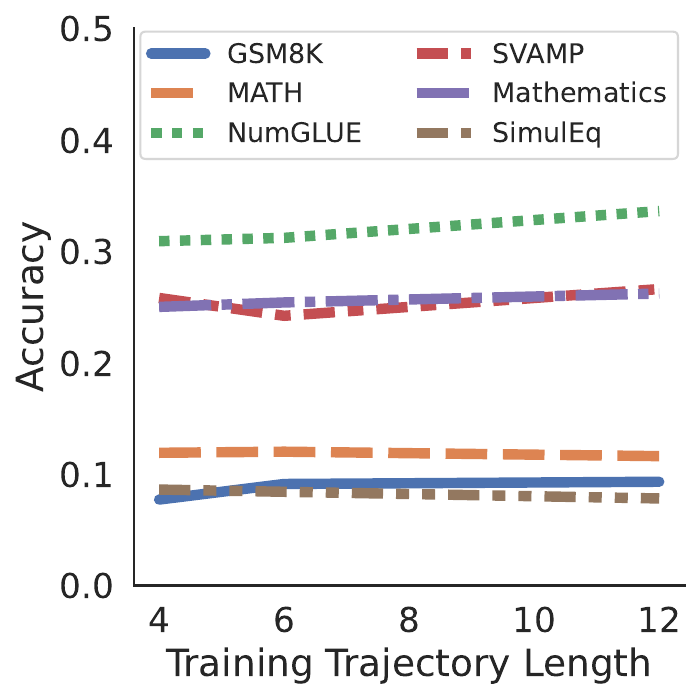}
  \captionof{figure}{{\alg} is robust to the length of training trajectories.}
  \label{fig:traject-length}
\end{minipage}%
\hfill
\begin{minipage}{.23\columnwidth}
  \centering
  \includegraphics[width=\linewidth]{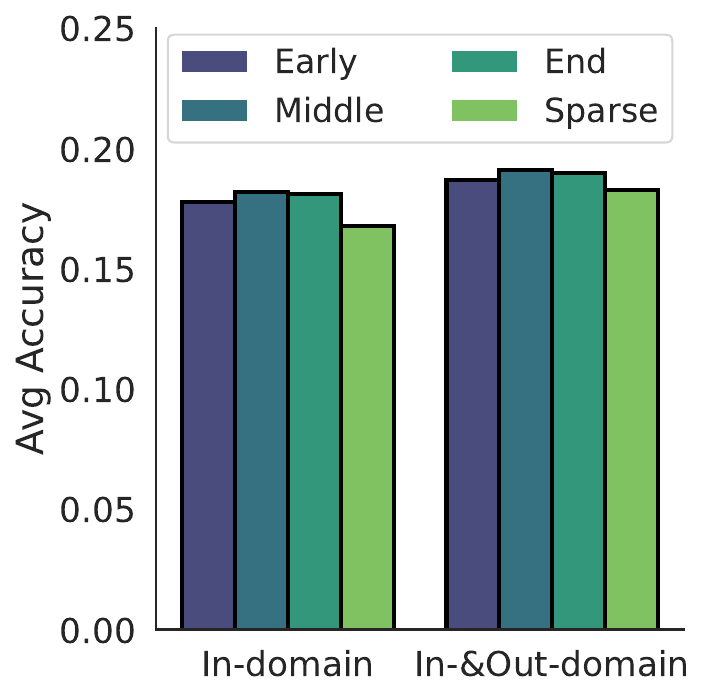}
  \captionof{figure}{{\alg} prefers dense trajectories over sparse ones.}
  \label{fig:traject-stage}
\end{minipage}
\end{wrapfigure}

\textbf{{\alg} is robust w.r.t. the length of the trajectories but can benefit more from longer trajectories.}
\cref{fig:traject-length} compares models trained with data selected by {\alg} based on training trajectories of different lengths.
The shorter trajectories are derived from a uniform sample of the longer trajectories. From the small slopes of the lines, we can conclude that {\alg} is relatively robust to the length of the training trajectories. Nevertheless, we can also observe a slight increase in the performance on some of the datasets when longer trajectories are used, so having longer trajectories is still preferred. 

\textbf{{\alg} can utilize training trajectories collected at any stage of training but preferably denser ones.} With the length of the trajectories fixed to 4, we can observe in \cref{fig:traject-stage} that denser trajectories recorded at any training stage (early, middle, or late) are more helpful for {\alg} than trajectories recorded sparsely throughout the training.

\begin{figure}[htbp]
    \centering
    \begin{minipage}[b]{0.49\textwidth}
        \centering
        \includegraphics[width=\textwidth]{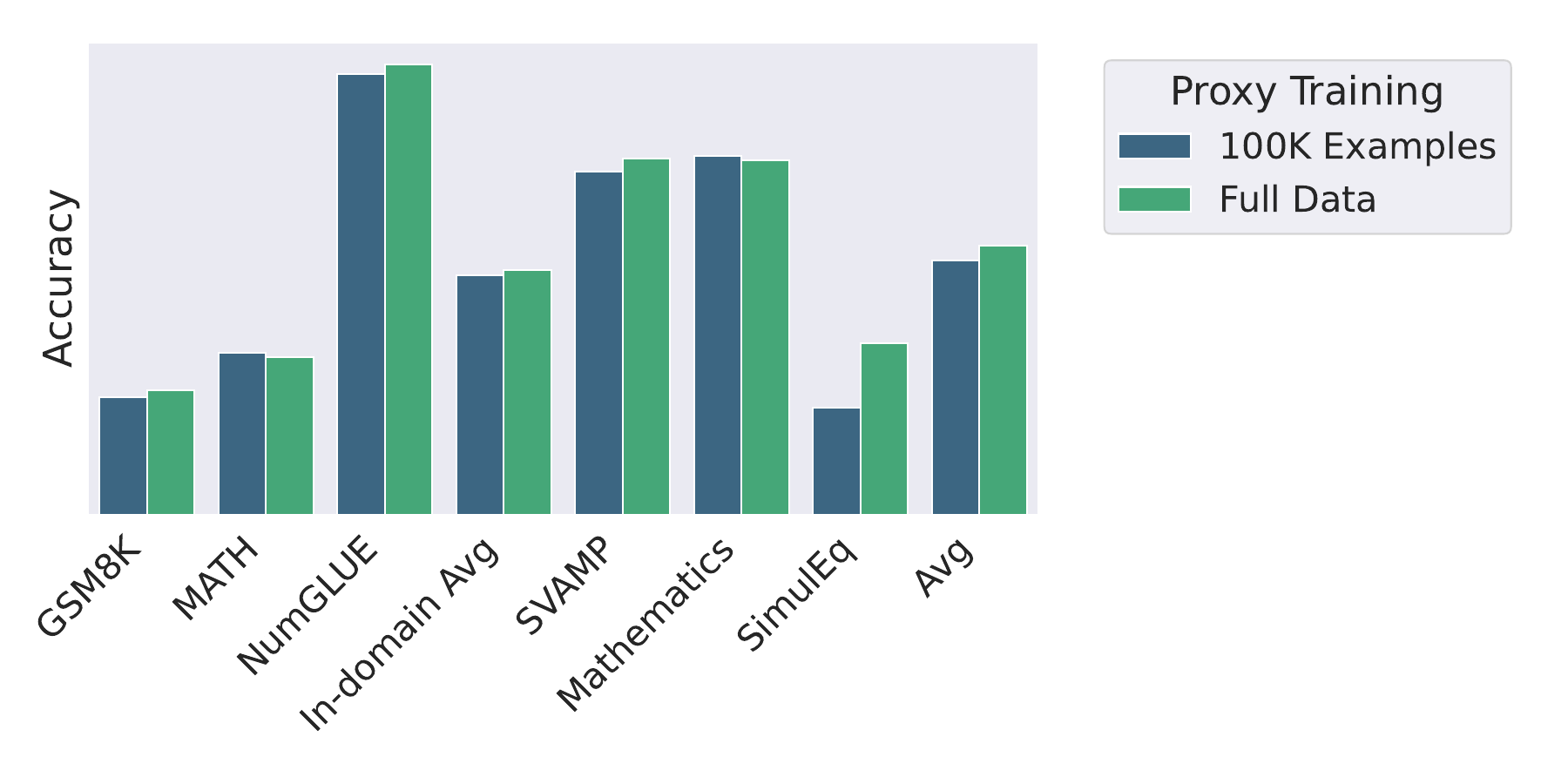}
        \caption{Per-dataset and average accuracy comparing proxy training on 100K examples and full data. {\alg} remains effective.}
        \label{fig:proxy-data-size}
    \end{minipage}
    \hfill
    \begin{minipage}[b]{0.49\textwidth}
        \centering
        \includegraphics[width=\textwidth]{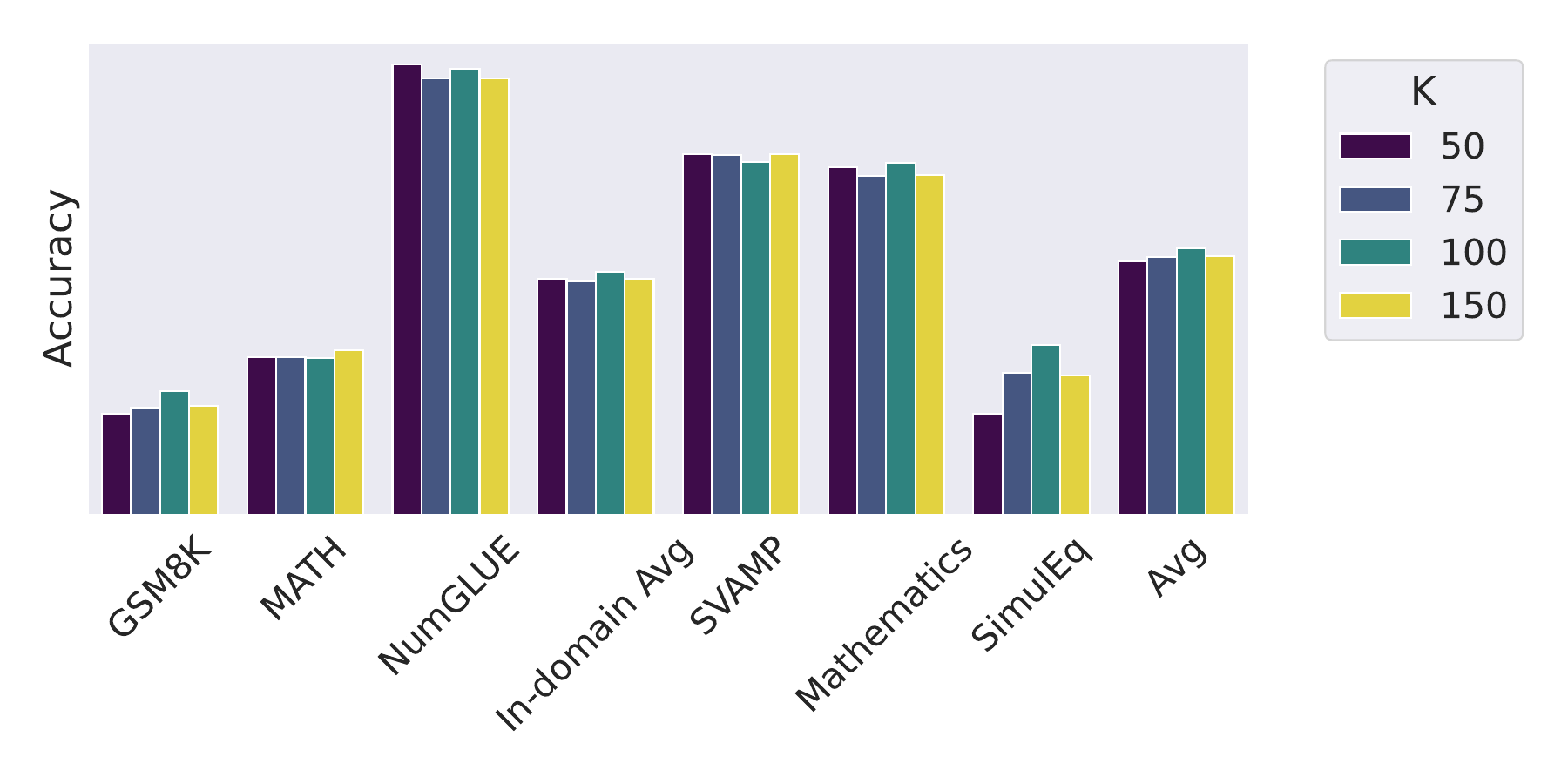}
        \caption{Per-dataset and average accuracy across different values of the clustering parameter $K$. {\alg} is relatively robust to the choice of $K$.}
        \label{fig:k}
    \end{minipage}
\end{figure}

\textbf{{\alg} does not require the full training data to train the proxy and can scale efficiently to larger datasets.} To further demonstrate the scalability of the proposed {\alg} method, we conducted experiments by training the proxy on a smaller sample of the data (100K examples) for the same number of epochs (3 epochs) and saving the loss for all examples. The results, shown in \cref{fig:proxy-data-size}, confirm that {\alg} remains effective when the proxy model is trained on a smaller subset of training data and therefore is scalable to larger datasets without a proportional increase in computational costs.

\textbf{{\alg} is robust across different clustering parameter values for K.} We conducted detailed experiments varying the clustering parameter K, as shown in \cref{fig:k}. The results demonstrate that {\alg} maintains high performance across different values of K, highlighting the robustness of our method to different clustering parameter choices. We chose K=100 for our experiments as it provided the best average accuracy across the evaluation datasets for the math reasoning task.

\begin{figure}[htbp]
    \centering
    \centering
    \begin{subfigure}[b]{0.49\textwidth}
        \centering
        \includegraphics[width=\textwidth]{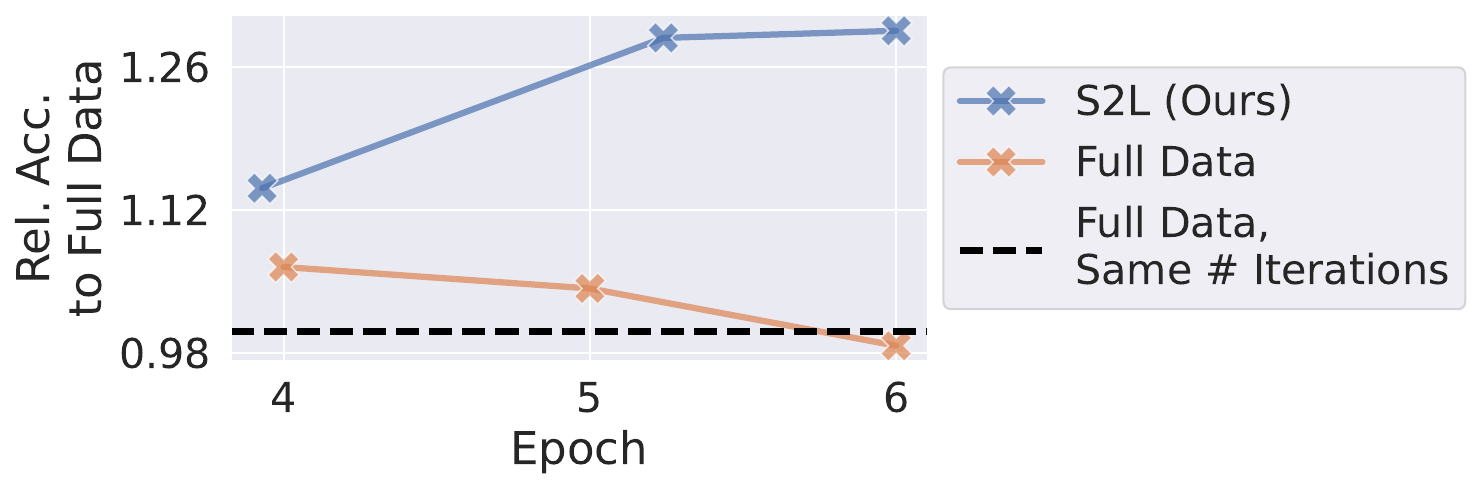}
        \caption{In-domain Average Accuracy}
        \label{fig:epoch-in-domain-avg}
    \end{subfigure}
    \hfill
    \begin{subfigure}[b]{0.49\textwidth}
        \centering
        \includegraphics[width=\textwidth]{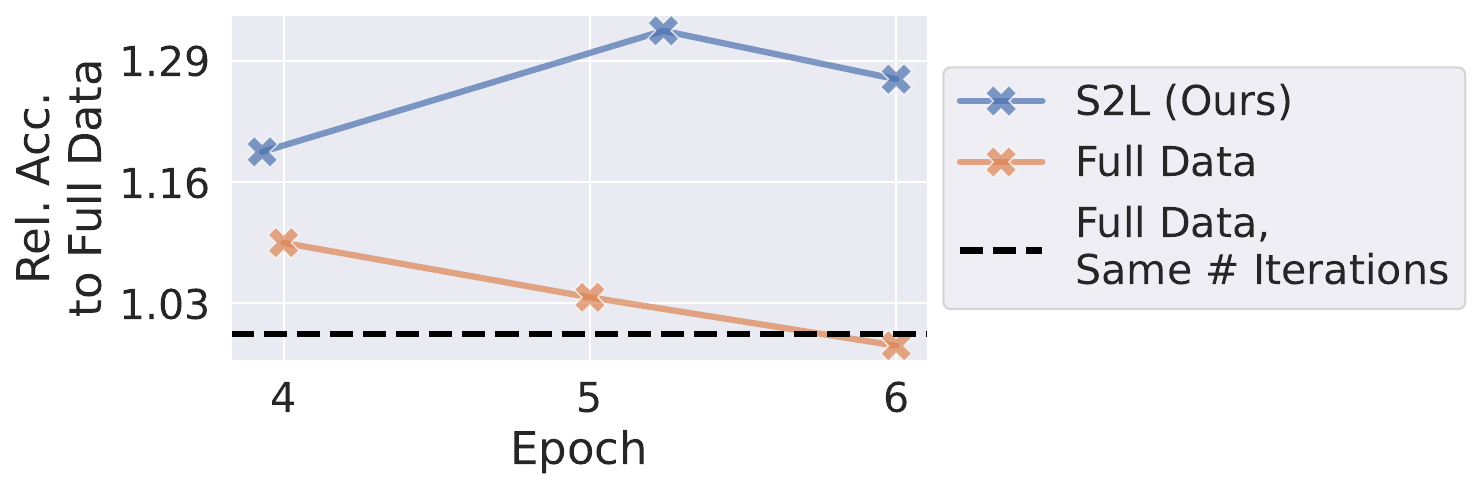}
        \caption{Overall Average Accuracy}
        \label{fig:epoch-avg}
    \end{subfigure}
    \caption{Relative accuracy to full data across different epochs, comparing S2L-selected data and full data. S2L achieves superior performance with fewer data and fewer training iterations.}
    \label{fig:more-epochs}
\end{figure}

\textbf{{\alg} remains effective and efficient compared to using full data when trained for the same number of epochs.} \cref{fig:more-epochs} illustrates the relative accuracy to full data across different epochs, comparing {\alg}-selected data and full data with the same number of epochs. Both in-domain and overall average accuracy are shown. {\alg} demonstrates superior performance with fewer data and fewer training iterations.

\begin{wrapfigure}{R}{.48\columnwidth}
    \centering
    \vspace{-2em}
    \includegraphics[width=0.5\textwidth]{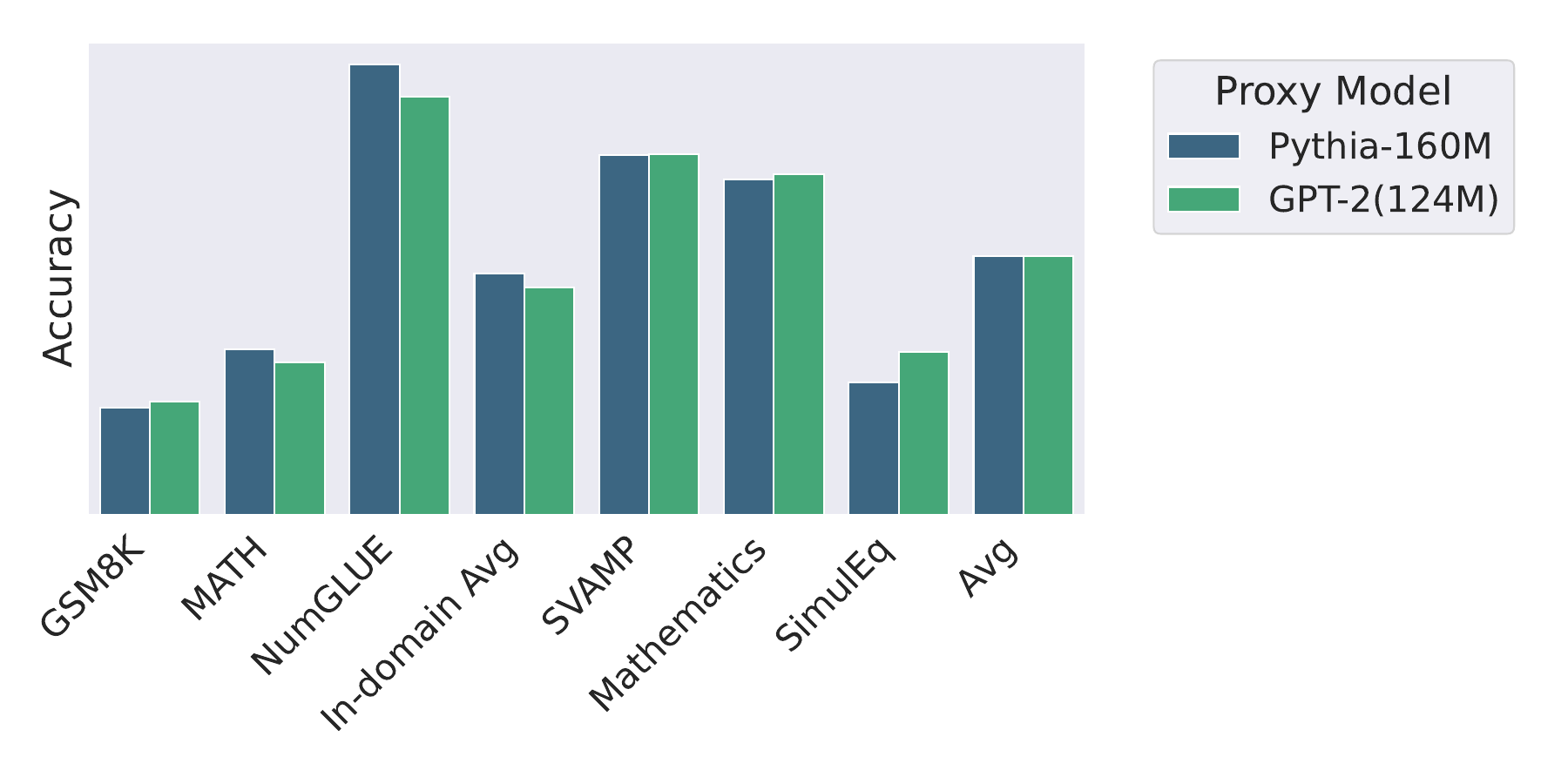}
    \caption{Per-dataset and average accuracy comparison between using different proxy models (Pythia-160M and GPT-2 (124M)) for data selection. Using both proxy models show comparable performance, demonstrating the effectiveness of different small models as reference models for {\alg}.}
    \label{fig:other-proxy}
    \vspace{-2em}
\end{wrapfigure}

\textbf{{\alg} supports a range of small models as effective proxies.} To understand whether different small models could serve as effective proxies, we used GPT-2 (124M) and Pythia-160M as proxy models for data selection to train Pythia-410M. The results, illustrated in \cref{fig:other-proxy}, show that both proxy models perform comparably in guiding the data selection, demonstrating the versatility and effectiveness of using different small models for S2L.

%% file: sections/5_conclusion.tex
\section{Conclusion and Limitations}\label{sec:limitations}
In this work, we introduced {\fullname} ({\alg}), a scalable data selection method to improve the data efficiency of supervised fine-tuning (SFT) for large language models (LLMs) in specialized domains. By clustering data points based on their training dynamics on smaller models and balanced sampling from all clusters, {\alg} significantly reduces the required training data size without compromising performance compared to using the entire training dataset. Our comprehensive experiments across the mathematical problem-solving and clinical text summarization domains demonstrate the effectiveness of {\alg}. 

Our study does come with its limitations.  {\alg} has been only tested within two domains, mathematics and medicine, and on models up to 7 billion parameters, constrained by our computational resources. Additionally, our experiments employed a fixed training schedule across all methods without further optimization or hyperparameter tuning for each method, including {\alg}. This unified approach, while it ensures a fair comparison, may not fully capture the potential performance improvement that could be achieved with more tailored training strategies. We encourage further research to extend the application of {\alg} across a broader spectrum of domains and investigate the impact of hyperparameter tuning on its effectiveness.

%% file: sections/6_appendix.tex
\section{Proofs}\label{sec:app-proof}
\subsection{Proof of \cref{lemma}}\label{proof:lemma}

\begin{proof}\label{proof-loss-functions}
From the assumption that the loss trajectories of examples on the proxy and target models are close:
\begin{align}
    \| \mathbf{L}_i^{\text{proxy}} - \mathbf{L}_i^{\text{target}} \| \leq \delta, \quad \forall i.
\end{align}
Since \( i \) and \( j \) are in the same cluster \( C_k \) based on the proxy model, we have:
\begin{align}
    \| \mathbf{L}_i^{\text{proxy}} - \mathbf{L}_j^{\text{proxy}} \| \leq \epsilon.
\end{align}
Using the triangle inequality:
\begin{align}
    \| \mathbf{L}_i^{\text{target}} - \mathbf{L}_j^{\text{target}} \| \leq \| \mathbf{L}_i^{\text{target}} - \mathbf{L}_i^{\text{proxy}} \| + \| \mathbf{L}_i^{\text{proxy}} - \mathbf{L}_j^{\text{proxy}} \| + \| \mathbf{L}_j^{\text{proxy}} - \mathbf{L}_j^{\text{target}} \| \leq 2\delta + \epsilon = \epsilon'.
\end{align}
Therefore, at any iteration \( t \):
\begin{align}\label{bounded-loss-diff}
    | \mathcal{L}_i^{\text{target}}(\bm{\theta}^{(t)}) - \mathcal{L}_j^{\text{target}}(\bm{\theta}^{(t)}) | \leq \epsilon', \quad \forall t.
\end{align}
Assuming that the loss functions can be approximated by:
\begin{align}
    \mathcal{L}_i^{\text{target}}(\bm{\theta}) = \frac{1}{2} d\bm{\theta}^\top \bm{H}_i d\bm{\theta} + \bm{g}_i^\top d\bm{\theta} + c_i,
\end{align}
where $c_i$ is the loss of example $i$ at the beginning of fine-tuning, and $d\bm{\theta}$ is the distance between the parameters of the pretrained model and those during fine-tuning.
Similarly for \( \mathcal{L}_j^{\text{target}}(\bm{\theta}) \).
The loss difference between \( i \) and \( j \) is:
\begin{align}
    \mathcal{L}_i^{\text{target}}(\bm{\theta}) - \mathcal{L}_j^{\text{target}}(d\bm{\theta}) = \frac{1}{2} d\bm{\theta}^\top (\bm{H}_i - \bm{H}_j) d\bm{\theta} + (\bm{g}_i - \bm{g}_j)^\top d\bm{\theta} + (c_i - c_j).
\end{align}
Given that \( | \mathcal{L}_i^{\text{target}}(\bm{\theta}) - \mathcal{L}_j^{\text{target}}(\bm{\theta}) | \leq \epsilon' \), we can write:
\begin{align}
\left| \frac{1}{2} d\bm{\theta}^\top (\bm{H}_i - \bm{H}_j) d\bm{\theta} + (\bm{g}_i - \bm{g}_j)^\top d\bm{\theta} + (c_i - c_j) \right| \leq \epsilon'.
\end{align}
Let us choose two different values, \( \bm{\theta}^{(1)} \) and \( \bm{\theta}^{(2)} \), to generate two inequalities. For \( d\bm{\theta}^{(1)} \), we have:
\begin{align}
\left| \frac{1}{2} (d\bm{\theta}^{(1)})^\top (\bm{H}_i - \bm{H}_j) d\bm{\theta}^{(1)} + (\bm{g}_i - \bm{g}_j)^\top d\bm{\theta}^{(1)} + (c_i - c_j) \right| \leq \epsilon',
\end{align}
and for \( d\bm{\theta}^{(2)} \), we have:
\begin{align}
\left| \frac{1}{2} (d\bm{\theta}^{(2)})^\top (\bm{H}_i - \bm{H}_j) d\bm{\theta}^{(2)} + (\bm{g}_i - \bm{g}_j)^\top d\bm{\theta}^{(2)} + (c_i - c_j) \right| \leq \epsilon'.
\end{align}
Subtracting these two inequalities, we get:
\begin{align}
\left| \frac{1}{2} \left( (d\bm{\theta}^{(1)})^\top (\bm{H}_i - \bm{H}_j) \bm{\theta}^{(1)} - (d\bm{\theta}^{(2)})^\top (\bm{H}_i - \bm{H}_j) d\bm{\theta}^{(2)} \right) + (\bm{g}_i - \bm{g}_j)^\top (d\bm{\theta}^{(1)} - d\bm{\theta}^{(2)}) \right| \leq 2\epsilon'.
\end{align}
{
\begin{align}
\begin{split}
    \left| (d\bm{\theta}^{(1)})^\top (\bm{H}_i - \bm{H}_j) d\bm{\theta}^{(1)} - (d\bm{\theta}^{(2)})^\top (\bm{H}_i - \bm{H}_j) d\bm{\theta}^{(2)} \right| 
    &\leq \|\bm{H}_i - \bm{H}_j\| \left( \|d\bm{\theta}^{(1)}\|^2 + \|d\bm{\theta}^{(2)}\|^2 \right)\\
    &\leq (\|\bm{H}_i\| + \|\bm{H}_j\|)\left( \|d\bm{\theta}^{(1)}\|^2 + \|d\bm{\theta}^{(2)}\|^2 \right)\\
    &\leq 4CD^2
\end{split}
\end{align}
This gives us:
\begin{align}
\left| (\bm{g}_i - \bm{g}_j)^\top (d\bm{\theta}^{(1)} - d\bm{\theta}^{(2)}) \right| \leq 2\epsilon' + 2CD^2.
\end{align}
Assuming \( \|d\bm{\theta}^{(1)} - d\bm{\theta}^{(2)}\| \geq d \), we get:
\begin{align}\label{eq:grad-diff-bound}
\| \bm{g}_i - \bm{g}_j \| \leq \frac{2\epsilon' + 2CD^2}{d}=\Delta.
\end{align}
}

\end{proof}

\subsection{Proof of \cref{theorem}}\label{proof:theorem}

Without loss of generality, assume we select $k$ example from each cluster and we have $k \leq \min_{j\in[K]} |C_j|$. Then the error of the subset in capturing the full gradient will be
\begin{equation}
    \xi \leq \sum_j (|C_j| - k) (\bar{\bm{g}}_{j}+\Delta),
\end{equation}
where $\bar{\bm{g}}_j$ is the norm of the average gradient of the selected examples from $C_j$.
In practice, we can %
weight elements of the subset by $r_{\min}/k$, which has a similar effect to scaling the step size when training on the subset. 
Let %
${\bm{g}}_{\max} = \max_j \|{\bm{g}}_j\|$ be the maximum gradient norm during training, $r_{\max}=\max_j |C_j|, r_{\min}=\min_j |C_j|$.
Then, we get
\begin{align}\label{eq:bound}
    \xi' &\leq \sum_j (r_{\min}-k)\Delta+(|C_j| - r_{\min}) (\bar{\bm{g}_j}+\Delta) \\
    &\leq K [r_{\min}\Delta + (r_{\max}-r_{\min})\bm{g}_{\max}]
\end{align}
The first term in RHS of Eq \eqref{eq:bound} is the error of the subset selected from $C_j$ to capture its full gradient and the second term is due to selecting the same number of examples, $k$, from the larger clusters.

Using the above error and following the proof of Theorem 1 in \cite{pmlr-v119-mirzasoleiman20a}, for a constant step size $\alpha\leq 1/c$ we get:
\begin{equation}
    \|\bm{\theta}^{t+1}-\bm{\theta}^*\|^2\leq (1-\alpha c)^{t+1} \|\bm{\theta}^{t}-\bm{\theta}^*\|^2 + 2\xi' R/c^2+\alpha B^2 (r_{\min}/k)^2 \bm{g}_{\max}^2,
\end{equation}
where $c\leq \|\bm{H}\|$, and %
$B=k\cdot K$ is the total size of the subset, 
$R = \min \{d_0, B \bm{g}_{\max} + \xi' / c \}$ and $d_0 = \|\bm{\theta}^0 - \bm{\theta}^*\|$ is the initial distance to the optimal solution $\bm{\theta}^*$. 

If $k \geq |C_j|$ for any cluster $C_j$, one can simply add 
$(r_{\min}/k-1)\cdot \hat{\bm{g}}_{j}$ to $\xi'$ for the corresponding clusters, where $\hat{\bm{g}_{j}}$ is the norm of the total gradient of cluster $C_j$ 
and we replace $r_{\min}$ in Eq \eqref{eq:bound} with the size of smallest cluster that has larger than $k$ examples.

\section{Experiment Details}\label{sec:app-details}
\subsection{Models}\label{sec:app-models}
\paragraph{Pythia.} The Pythia models \citep{biderman2023pythia} are a suite of large language models (LLMs) developed by EleutherAI licensed under the Apache License 2.0. These models range in size from 70 million to 12 billion parameters and are designed to enable controlled scientific research on transparently trained LLMs across various scales.

\paragraph{Phi.} The Phi models \citep{textbooks2} developed by Microsoft are under the MIT License. Phi-1.5, a transformer-based model with 1.3 billion parameters, and its successor, Phi-2, with 2.7 billion parameters, have been trained on a diverse set of data sources, including synthetic texts and curated websites. The Phi models underscore the potential of small yet powerful language models in understanding and generating human language, empowering a range of NLP tasks. Phi-2, in particular, has raised the bar for reasoning and language understanding among foundation models, matching or even exceeding the performance of models 25 times its size on complex benchmarks.

\paragraph{LLaMA 2.} The LLaMA 2 models \cite{touvron2023llama}, released by Meta AI and licensed under the LLaMA 2 Community License Agreement, are designed for improved natural language understanding and generation. LLaMA 2-7B, the smallest in this series with 7 billion parameters, has demonstrated competitive performance across various NLP benchmarks despite its moderate size.

\subsection{Datasets}\label{sec:app-datasets}
\paragraph{MathInstruct.} The MathInstruct dataset \citep{yue2023mammoth} is compiled from 13 diverse math rationale datasets, using both chain-of-thought (CoT) and program-of-thought (PoT) rationales. It ensures comprehensive coverage across various mathematical fields in the 262K training examples, making it a popular resource for fine-tuning large language models (LLMs) for general math problem-solving. MathInstruct is licensed under the MIT license. 

\paragraph{MIMIC-III.} The MIMIC-III (Medical Information Mart for Intensive Care III) dataset \citep{johnson2016mimic} is a comprehensive collection of de-identified health data associated with over 40,000 patients who stayed in critical care units of the Beth Israel Deaconess Medical Center in Boston, Massachusetts. This large dataset includes information such as demographics, vital signs, laboratory tests, medications, and more, making it an invaluable resource for a wide range of research in healthcare, including clinical decision support systems, medical procedure efficacy studies, and patient care optimization strategies.

The MIMIC-III dataset is made freely available to the research community under the Health Insurance Portability and Accountability Act (HIPAA) compliance, ensuring patient confidentiality and data protection. Access to the dataset is granted under a data use agreement (DUA) to individuals affiliated with an institution that approves the use of the data for research purposes. Researchers seeking to utilize the MIMIC-III dataset must complete a required training course on human research protections, which ensures that all researchers are aware of the responsibilities involved in handling sensitive patient data.

\begin{table}
\caption{A synthetic radiology report 
(MRI of the brain), 
generated by the GPT-4 model \citep{Achiam2023GPT4TR} to demonstrate the typical data format and content used in the clinical text summarization task. It is not suitable for clinical or diagnostic use.}
\label{tab:report-example}
\begin{center}
\begin{small}
\resizebox{\columnwidth}{!}{%
\begin{tabular}{p{0.15\linewidth}  p{0.8\linewidth}}
\toprule
Findings & The brain parenchyma demonstrates normal morphology with no evidence of mass effect or midline shift.
No acute infarcts are seen on diffusion-weighted images.
There are no signs of intracranial hemorrhage.
Mild generalized cerebral atrophy is noted.
The ventricles and sulci appear within normal limits for the patient's age.
The pituitary gland and sella turcica are unremarkable.
There are no abnormal signal intensities within the brain parenchyma.
The orbits, paranasal sinuses, and mastoid air cells are clear.\\
\midrule
Impression & Normal MRI of the brain.
Mild cerebral atrophy, likely age-related.
No acute intracranial pathology.\\
\bottomrule
\end{tabular}
}%
\end{small}
\end{center}
\end{table}

\subsection{Implementation Details}
\paragraph{{\alg}} The training trajectories for both MathInstruct and MIMIC-III are gathered from training a Pythia-70M model, the smallest model in the Pythia model suite, recorded every $500$ training iterations. We utilize the Faiss library \citep{douze2024faiss} to perform efficient K-means clustering of loss trajectories with Euclidean distance with $K=100$ and $20$ iterations. The hyperparameter $K$ is tuned in the range of $\{50, 100, 200\}$ based on the average accuracy of the model trained on $30K$ selected data. We found $K=100$ worked the best for both datasets we studied in this paper. Ablations studies on the length and the best time in the training to record the trajectories can be found in \cref{sec:ablation}. 

\paragraph{Comparing Reference Models for the Baselines}\label{sec:ablation-quality}
For one-shot selection methods (excluding \alg), we use representations from either step 1000 or the end of fine-tuning Pythia-410M on MathInstruct and reported the better result in \cref{fig:410m} and \cref{tab:math-scale}. In \cref{tab:reference-model}, we include the complete comparison between using early-fine-tuning vs. end-of-fine-tuning model checkpoints as the inference model. For Facility Locations, we further compared using the first hidden states as the feature representation as suggested in \citep{bhatt2024experimental} to using the last hidden states \citep{wu2023self} for the tasks we studied.The ranges for confidence, perplexity, and learnability are 
 chosen according to the best-performing intervals reported in prior research (\cref{sec:math-baseline}). 

Due to memory and computational constraints, for Facility Locations,
we calculate pairwise similarity and perform greedy selection on a per-data-source basis. We found this per-source selection approach also yields benefits for {\alg} as different data sources within MathInstruct exhibit distinct common patterns in their training trajectories. Therefore, we implement {\alg} also on a per-source basis for MathInstruct, and recommend applying {\alg} per source when dealing with datasets composed of multiple data sources. 

\paragraph{Hyperparameters} Following the setup used in \citep{yue2023mammoth}, we adopt a training regimen with a learning rate of 2e-5, a batch size of 128, a maximum length of 512, and a cosine scheduler with a 3\% warm-up period. 

\paragraph{Experiments Compute Resources}\label{sec:compute} We fine-tune all the models with the Huggingface transformers library \citep{wolf-etal-2020-transformers} with Fully Sharded Data Parallel (FSDP) \citep{zhao2023pytorch} on 4 48G NVIDIA RTX A6000. 

\begin{table*}[ht!]
\caption{Complete results used for selecting the best reference model for each one-shot data selection baseline. The choice of early-fine-tuning (step 1000) and end-of-fine-tuning checkpoint follows \citep{marion2023less}. The best results selected for \cref{fig:410m} are highlighted in cyan.} 
\label{tab:reference-model}
\vskip -0.2in
\begin{center}
\begin{small}
\begin{sc}
\resizebox{\textwidth}{!}{%
\begin{tabular}{llr|rrr|r|rrr|r}
\toprule
 & Ref & Data  & \multicolumn{4}{c|}{In-domain} & \multicolumn{3}{c|}{Out-domain} & \\ 
Selection & Model & Size & GSM8K & MATH & NumGLUE & \textbf{Avg} & SVAMP & Mathematics & SimulEq & \textbf{Avg} \\ 
\midrule
\multirow{6}{*}{\shortstack[l]{Least\\Confidence}} 
& \multirow{3}{*}{Early} & 30K & 2.3 & 1.7 & 15.5 & 6.5 & 13.6 & 1.2 & 0.5 & 5.8 \\
& & 50K & 1.7 & 2.6 & 20.5 & 8.3 & 16.0 & 4.0 & 1.8 & 7.8  \\
& & 100K & 3.9 & 2.7 & 22.5 & 9.7 & 19.2 & 8.0 & 3.3 & 9.9  \\
\cmidrule{2-11}
& \multirow{3}{*}{End} & 30K & 2.7 & 1.3 & 18.0 & \cellcolor{LightCyan} 7.0 & 13.7 & 3.3 & 1.4 & \cellcolor{LightCyan} 6.7\\
& & 50K & 2.1 & 1.7 & 21.0 & \cellcolor{LightCyan} 8.3 & 14.5 & 3.5 & 1.0 & \cellcolor{LightCyan} 7.3 \\
& & 100K & 2.5 & 3.3 & 23.5 & \cellcolor{LightCyan} 9.8 & 20.8 & 6.3 & 3.7 & \cellcolor{LightCyan} 10.0 \\
\midrule
\multirow{6}{*}{\shortstack[l]{Middle \\ Perplexity}} & \multirow{3}{*}{Early} & 30K & 3.3 & 3.8 & 17.5 & 8.2 & 11.8 & 1.2 & 1.2 & 6.5 \\
& & 50K & 2.9 & 4.1 & 19.6 & 8.9 & 15.6 & 7.6 & 2.9 & 8.8 \\
& & 100K & 4.8 & 7.1 & 20.4 & 10.8 & 19.6 & 16.1 & 3.9 & 12.0\\
\cmidrule{2-11}
& \multirow{3}{*}{End} & 30K & 5.3 & 3.7 & 16.2 & \cellcolor{LightCyan} 8.4 & 14.2 & 8.7 & 1.2 & \cellcolor{LightCyan} 8.2 \\
& & 50K & 3.2 & 5.9 & 20.5 & \cellcolor{LightCyan} 9.9 & 18.1 & 11.3 & $\mathbf{5.1}$ & \cellcolor{LightCyan} 10.7 \\
& & 100K & 5.4 & 7.2 & 20.9 & \cellcolor{LightCyan} 11.2 & $\mathbf{23.8}$ & 15.3 & 3.3 & \cellcolor{LightCyan} 12.6 \\
\midrule
\multirow{6}{*}{\shortstack[l]{High\\Learnability}}& 
\multirow{3}{*}{Early} & 30K & 6.1 & 1.6 & 19.1 & \cellcolor{LightCyan} 8.9 & 10.7 & 9.9 & 1.4 & \cellcolor{LightCyan} 8.1\\
& & 50K & 6.1 & 2.1 & 18.6 & \cellcolor{LightCyan} 8.9 & 14.5 & 14.0 & 2.1 & \cellcolor{LightCyan} 8.9 \\
& & 100K & $\mathbf{7.4}$ & 9.2 & $\mathbf{29.8}$ & \cellcolor{LightCyan} $\mathbf{15.5}$ & 20.7 & 19.4 & $\mathbf{10.3}$ & \cellcolor{LightCyan} $\mathbf{16.1}$ \\
\cmidrule{2-11}
& \multirow{3}{*}{End} & 30K & 3.0 & 1.4 & 14.7 & 6.4 & 2.1 & 6.8 & 1.8 & 5.0 \\
& & 50K & 1.3 & 2.1 & 16.0 & 6.5 & 4.7 & 6.9 & 3.1 & 5.7\\
& & 100K & 4.3 & 7.2 & 23.0 & 11.5 & 16.7 & 16.1 & 4.3 & 11.9\\
\midrule
\multirow{4}{*}{\shortstack[l]{Facility\\Location}} & Early (First) & 50K & 3.9 & 7.6 & 12.4 & 8.0 & 11.1 & 14.6 & 1.9 & 8.6\\
\cmidrule{2-11}
& Early (Last) & 50K & 5.7 & 9.1 & 12.4 & \cellcolor{LightCyan} 9.1 & 15.4 & 18.6 & 1.6 & \cellcolor{LightCyan} 10.5 \\
\cmidrule{2-11}
& End (First) & 50K & 3.8 & 7.7 & 14.8 & 8.7 & 19.2 & 11.4 & 2.3 & 9.9\\
\cmidrule{2-11}
 & End (Last) & 50K & 5.2 & $\mathbf{9.7}$ & 11.8 &  8.9 & 12.4 & 18.2 & 1.0 &  9.7\\
\bottomrule
\end{tabular}
}%
\end{sc}
\end{small}
\end{center}
\vskip -0.1in
\end{table*}

\subsection{Evaluation}\label{sec:app-eval}
\subsubsection{MathInstruct}

\paragraph{Datasets.} We utilize 6 diverse datasets with open-formed questions for evaluating the mathematical reasoning capabilities of models trained with both the full MathInstruct dataset and selected subsets. These datasets, detailed in \cref{tab:math-eval-data}, span a range of mathematical disciplines from early algebra to calculus and linear algebra, covering various types of questions such as multi-step reasoning, arithmetic word problems, and problems from mathematics competitions. This variety ensures a comprehensive assessment across both in-domain and out-domain tasks.

\begin{table*}[t!]
\caption{Types of questions in the evaluation datasets for the mathematical reasoning task.}
\label{tab:math-eval-data}
\vskip 0.1in
\begin{center}
\begin{small}
\resizebox{0.8\textwidth}{!}{%
\begin{tabular}{p{0.1\linewidth} p{0.1\linewidth} p{0.24\linewidth} p{0.4\linewidth}}
\toprule
\textsc{Dataset} & \textsc{Size} & \textsc{Level} & \textsc{Tasks} \\
\midrule
GSM8K    &  1319 & Early Algebra & Multi-step reasoning using basic arithmetic operations \\
\midrule
MATH &  5000 & Early Algebra, Intermediate Algebra, Algebra, Probability, NumTheory, Calculus, Geometry & Problems from mathematics competitions, including the AMC 10, AMC 12, AIME\\
\midrule
NumGLUE &  1042 & Early Algebra & Commonsense, Domain-specific, Arithmetic Reasoning, Quantitative Comparison, Fill-in-the-blanks Format, Reading Comprehension, Numerical Reasoning, Quantitative NLI, Arithmetic Word Problems \\
\midrule
SVAMP & 1000 & Early Algebra & Arithmetic Word Problems \\
\midrule
Mathematics & 1000 & Early Algebra, Intermediate Algebra, NumTheory, Calculus & Arithmetic Reasoning \\
\midrule
SimulEq & 514 & Linear Algebra & Single and multiple equation word problems\\
\bottomrule
\end{tabular}
}%
\end{small}
\end{center}
\vskip -0.1in
\end{table*}

\paragraph{Pipeline.} We utilize the pipeline provided by \citep{yue2023mammoth}\footnote{\href{https://github.com/TIGER-AI-Lab/MAmmoTH?tab=readme-ov-file\#large-scale-evaluation}{https://github.com/TIGER-AI-Lab/MAmmoTH?tab=readme-ov-file\#large-scale-evaluation}}, designed to first determine whether the model can be prompted to generate a code snippet. This code snippet, if successfully generated, should be executable and produce the correct answer when run. This code-based evaluation is also used for Phi models \citep{textbooks2}. In cases where the model does not directly produce a viable code solution, we employ a ``think step-by-step" prompting strategy \citep{wei2022chain}. This method prompts the model to break down its reasoning process, a technique that has been widely proven effective in fully exploiting the model’s problem-solving capacity.

\subsubsection{MIMIC-III}
Following \citep{delbrouck-etal-2023-overview,bionlp-2023-biomedical}, we include the six most common modality/anatomy pairs: CT head, CT abdomen, CT chest, MRI head, CT spine, and CT neck, and five less common pairs in the text data: MRI spine, CT sinus, MRI abdomen, MRI pelvis, and MRI neck in the evaluation. There are in total 13.7K test examples after data preprocessing and train-test splitting.

\begin{figure}
\vskip -0.05in
\begin{center}
     \begin{subfigure}[b]{0.32\columnwidth}
         \centering
         \includegraphics[width=\textwidth]{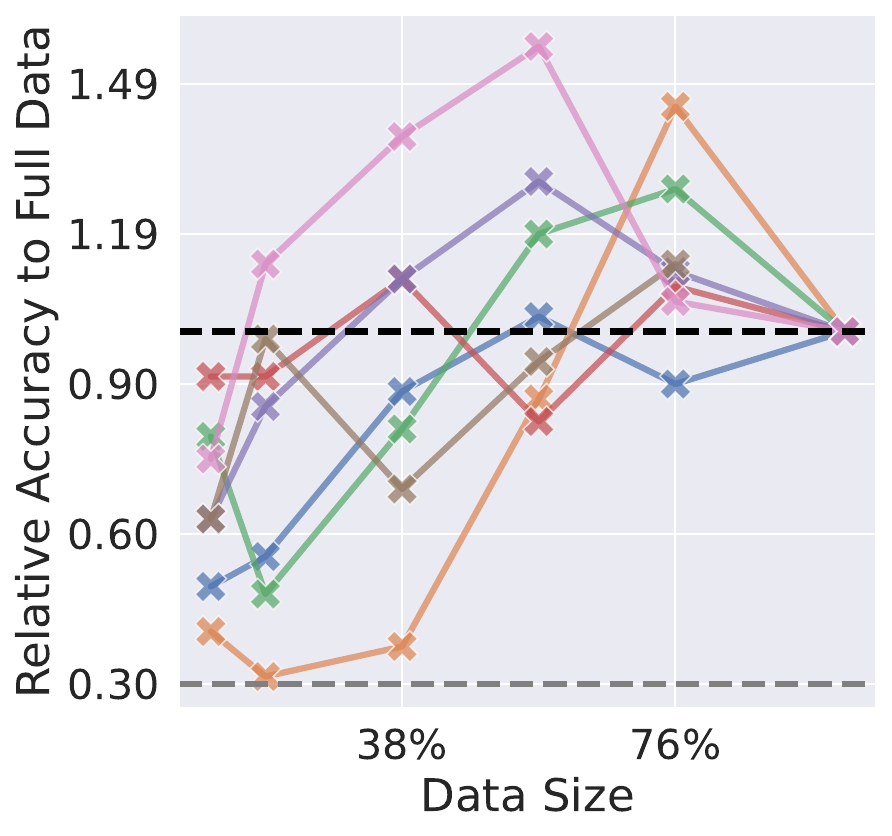}
         \caption{GSM8K}
     \end{subfigure}
     \begin{subfigure}[b]{0.32\columnwidth}
         \centering
         \includegraphics[width=\textwidth]{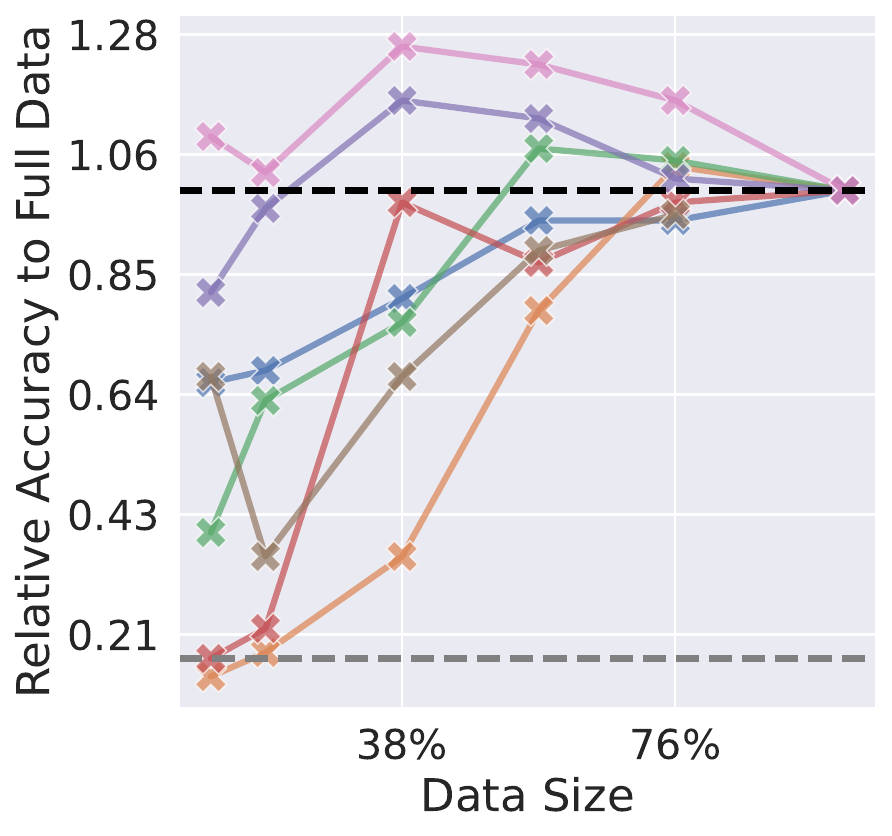}
         \caption{MATH}
     \end{subfigure}
     \begin{subfigure}[b]{0.32\columnwidth}
         \centering
         \includegraphics[width=\textwidth]{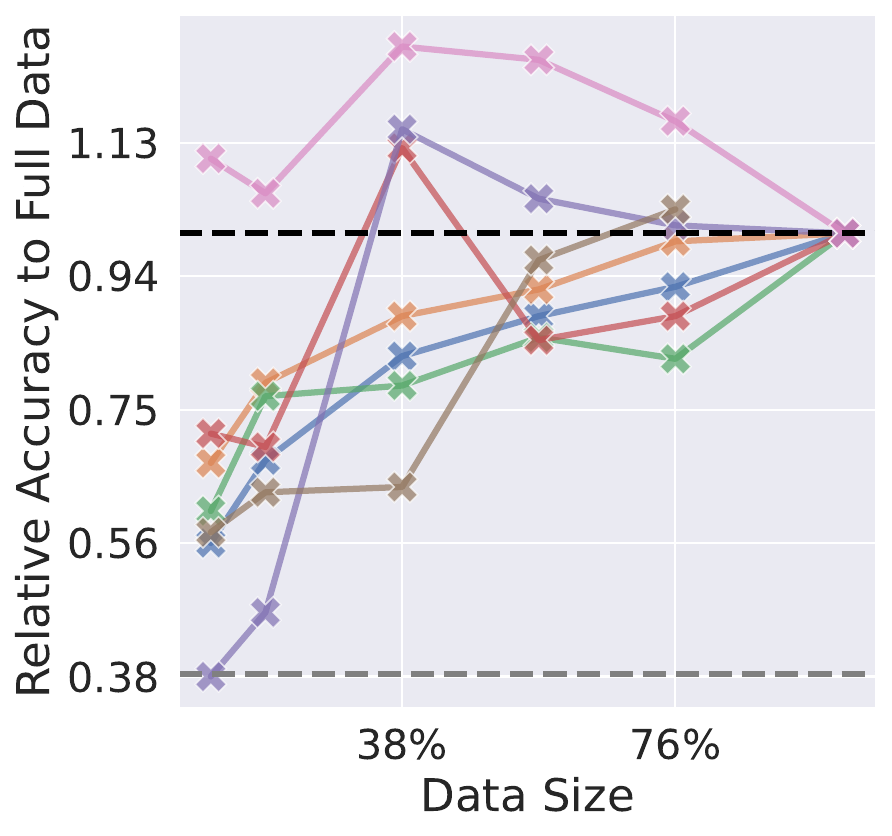}
         \caption{NumGLUE}
     \end{subfigure}
     \begin{subfigure}[b]{0.32\columnwidth}
         \centering
         \includegraphics[width=\textwidth]{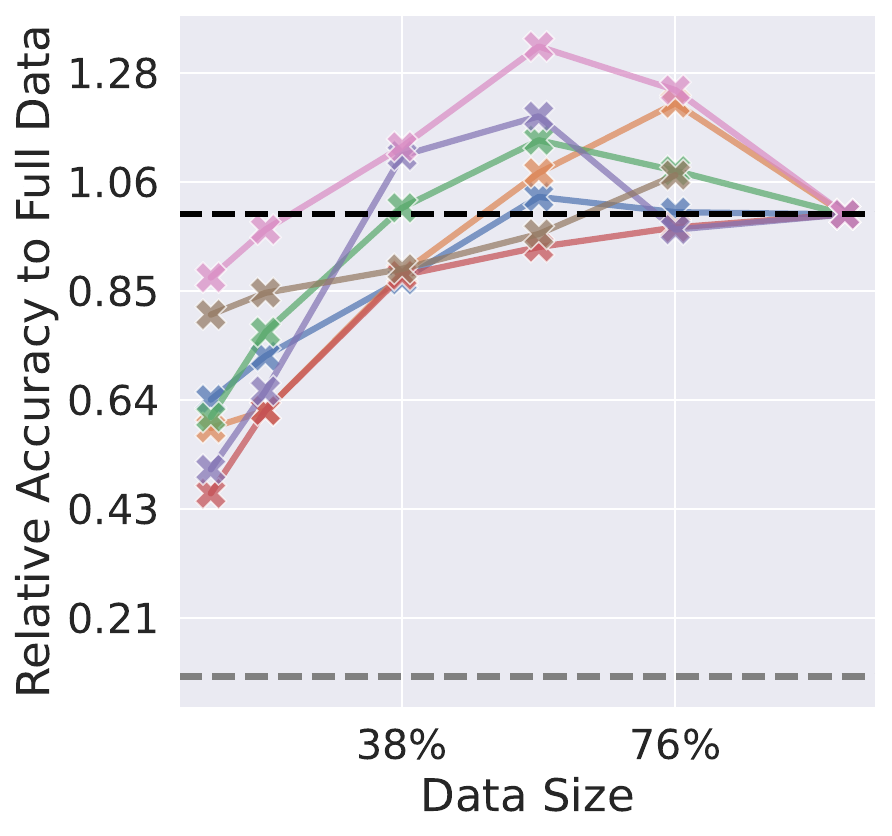}
         \caption{SVAMP}
     \end{subfigure}
     \begin{subfigure}[b]{0.32\columnwidth}
         \centering
         \includegraphics[width=\textwidth]{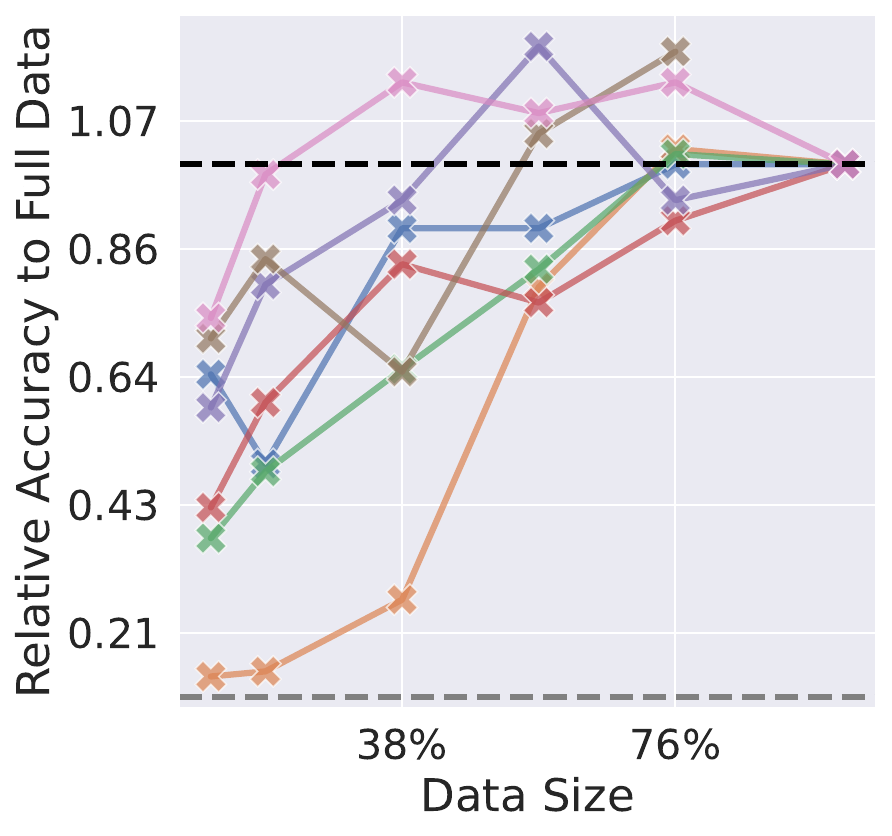}
         \caption{Mathematics}
     \end{subfigure}
     \begin{subfigure}[b]{0.32\columnwidth}
         \centering
         \includegraphics[width=\textwidth]{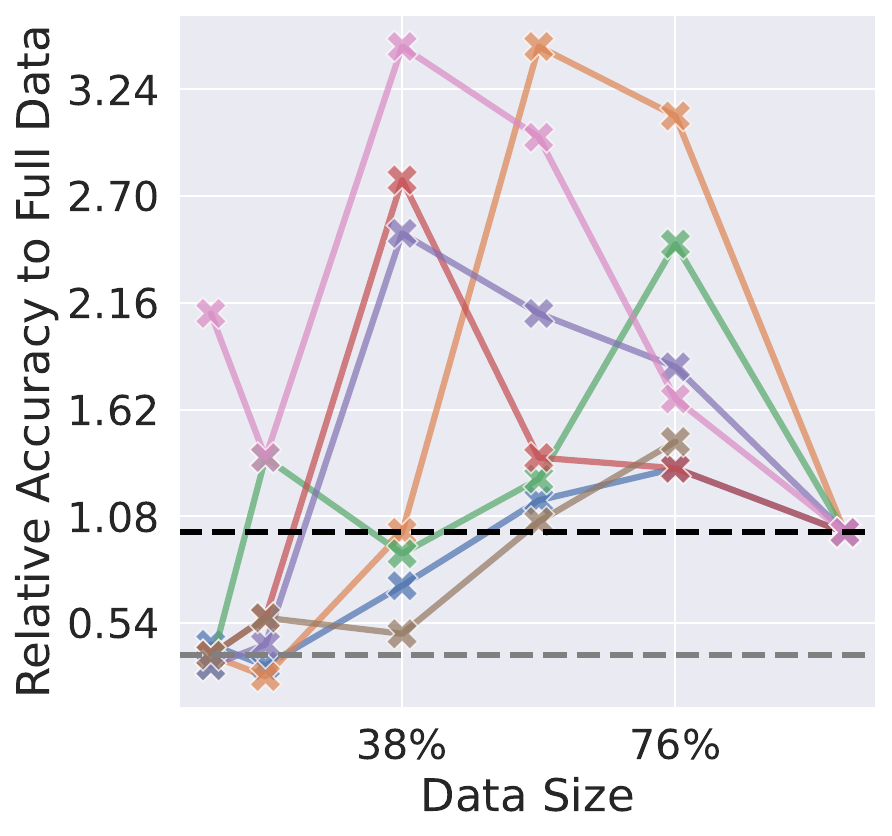}
         \caption{SimulEq}
     \end{subfigure}
     \begin{subfigure}[b]{0.32\columnwidth}
         \centering
         \includegraphics[width=\textwidth]{figures/410m/In-domain_Avg.pdf}
         \caption{In-domain Avg}
     \end{subfigure}
     \begin{subfigure}[b]{0.54\columnwidth}
         \centering
         \includegraphics[width=\textwidth]{figures/410m/Avg.pdf}
         \caption{Avg}
     \end{subfigure}
\caption{Break-down accuracies ($\uparrow$) on in-domain and out-of-domain datasets using Pythia-410M. Data size refers to the total number of unique training examples used. All experiments in this table share the same total training steps and learning rate schedule (see \cref{sec:math-train}). 
}
\label{fig:410m-full}
\end{center}
\end{figure}

\section{Examples in Different Clusters}\label{sec:examples-full}
\begin{figure}[ht!]
\vskip 0.1in
\begin{center}
\centerline{\includegraphics[width=0.8\textwidth]{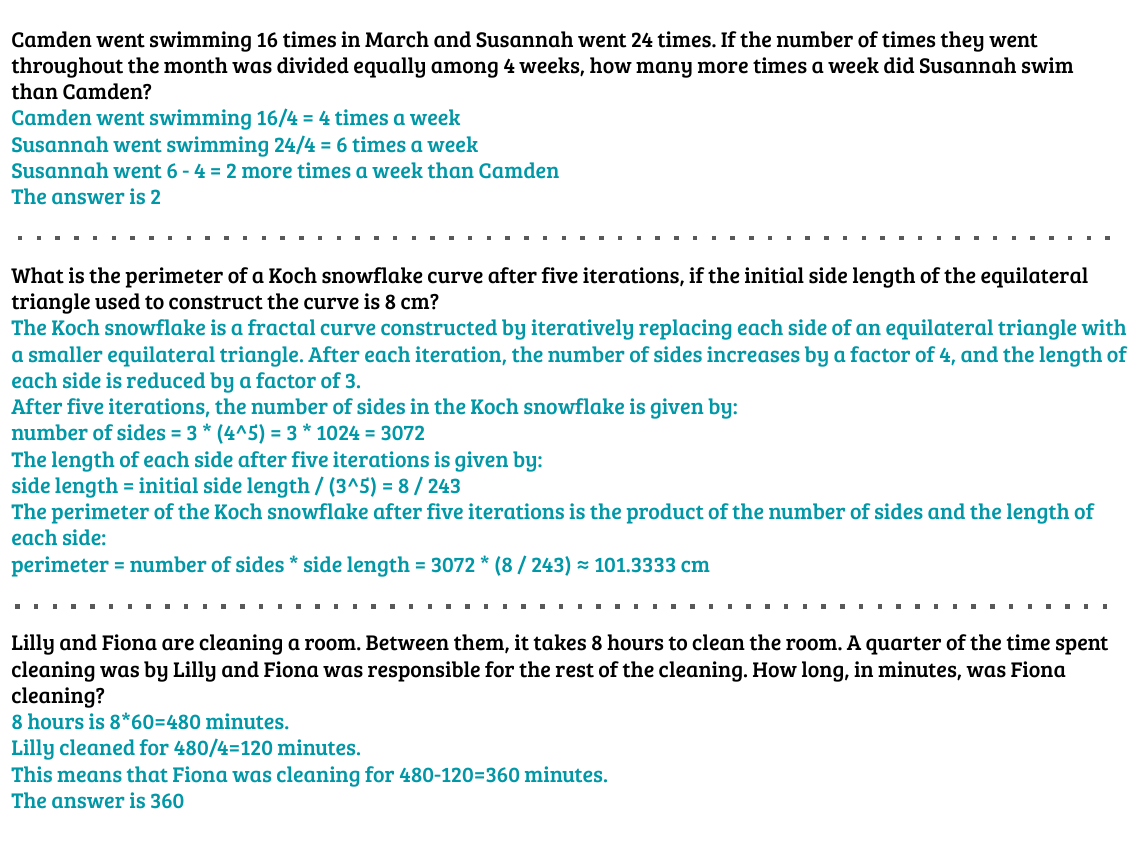}}
\vskip -0.2in
\caption{Examples in the cluster shown in \cref{fig:down}: open-formed algebra. Questions are in black and answers are in cyan.}
\label{fig:ex-down-full}
\end{center}
\vskip -0.2in
\end{figure}

\begin{figure}[ht!]
\begin{center}
\centerline{\includegraphics[width=0.8\textwidth]{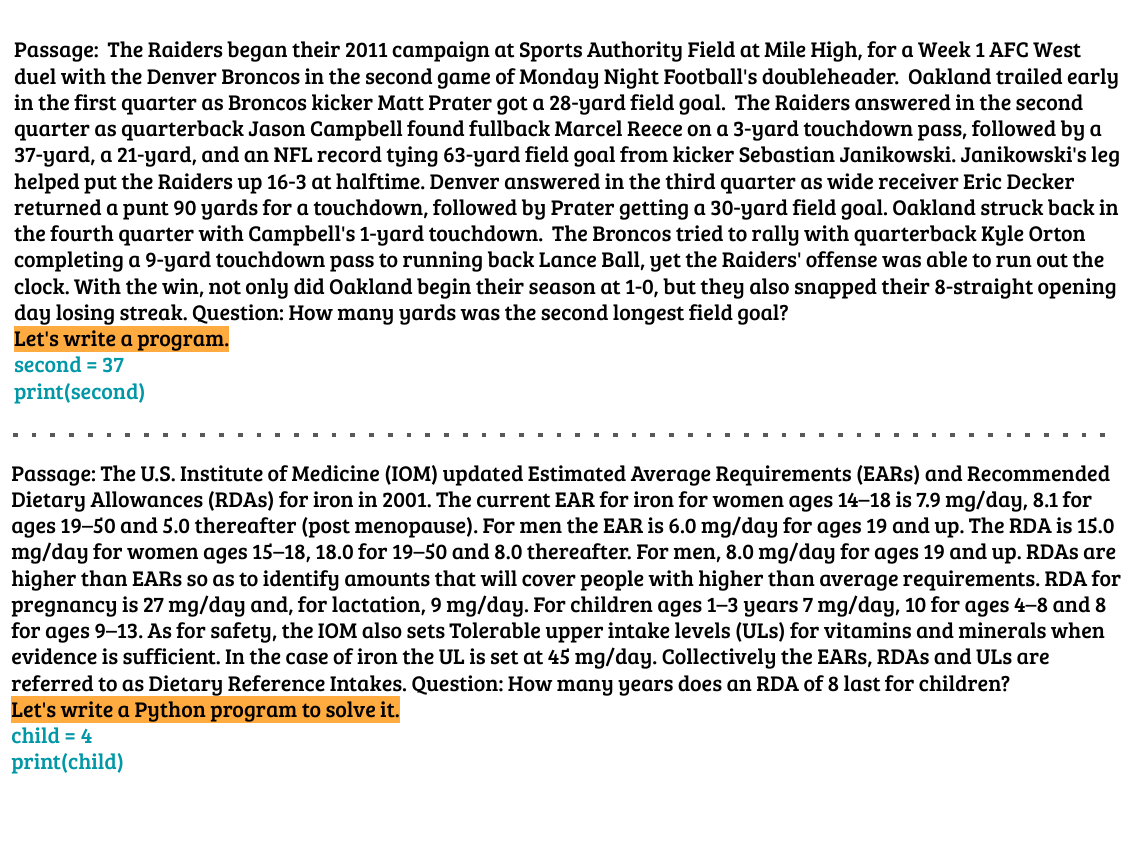}}
\vskip -0.3in
\caption{Examples in the cluster shown in \cref{fig:up}: reading comprehension + coding. Questions are in black and answers are in cyan; instructions are highlighted in orange.}
\label{fig:ex-up-full}
\end{center}
\vskip -0.2in
\end{figure}

\begin{figure}[ht!]
\begin{center}
\centerline{\includegraphics[width=0.8\textwidth]{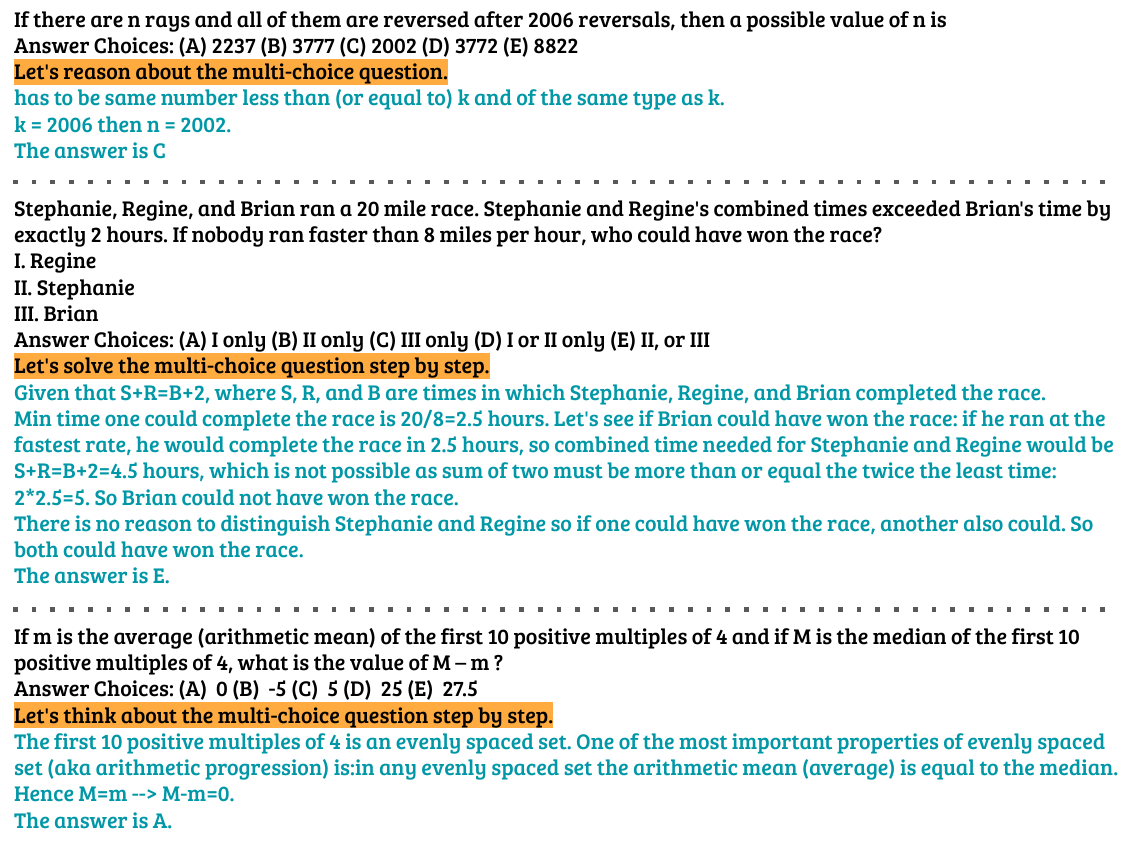}}
\vskip -0.1in
\caption{Examples in the cluster shown in \cref{fig:double}: multiple-choice + multi-step reasoning. Questions are in black and answers are in cyan; instructions are highlighted in orange.}
\label{fig:ex-double-full}
\end{center}
\vskip -0.2in
\end{figure}

We compare data points in the same and different clusters based on training trajectories, in \cref{fig:ex-down-full}, \cref{fig:ex-up-full} and \cref{fig:ex-double-full}. We can observe that examples with similar training trajectories have the same question format. Therefore, balanced sampling from all clusters can ensure different types of examples can be covered in the selected subset of training data.

\newpage
\section{Topic Distribution of Data Selected by {\alg}}
\begin{figure}[ht!]
\begin{center}
     \begin{subfigure}[b]{\columnwidth}
         \centering
         \includegraphics[width=\textwidth]{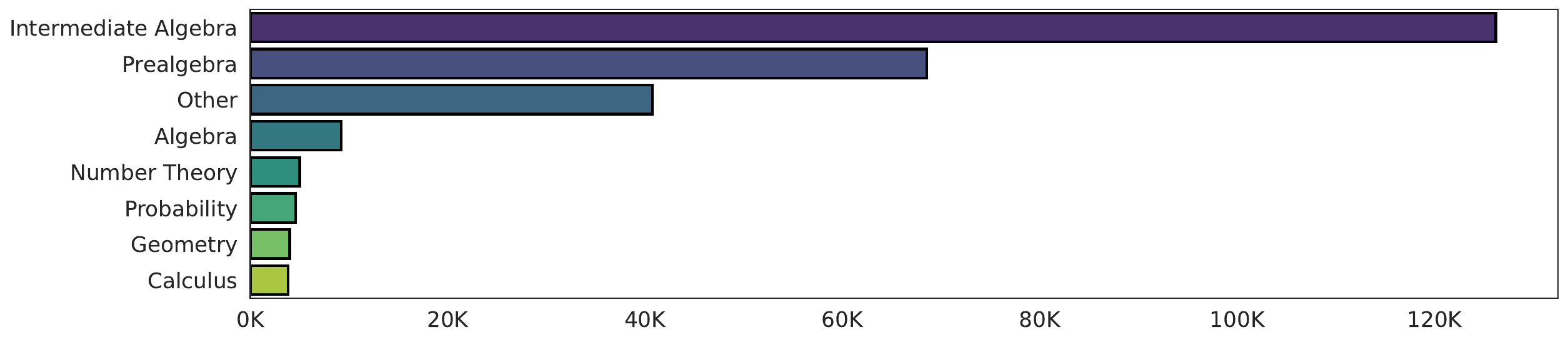}
         \caption{Topic distribution of the full MathInstruct dataset.}
         \label{fig:dist-full}
     \end{subfigure}
     \vskip 0.2in
     \begin{subfigure}[b]{0.48\columnwidth}
         \centering
         \includegraphics[width=0.6\textwidth]{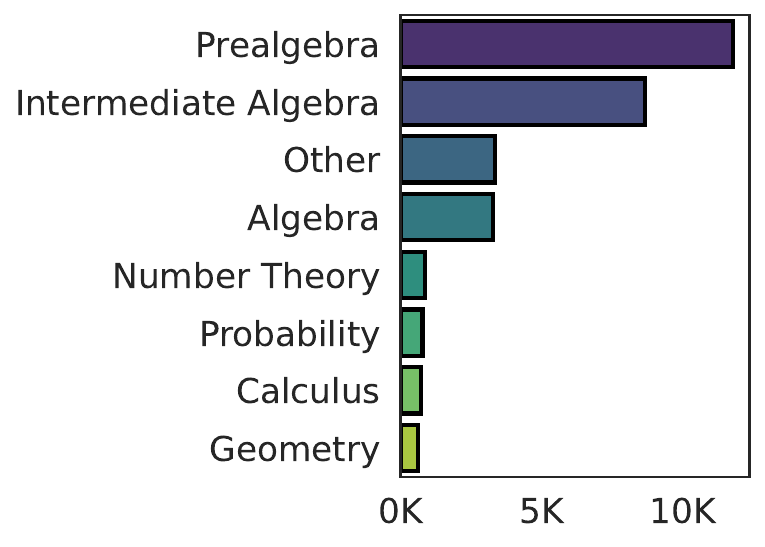}
         \caption{Topic distribution of 30K data selected by {\alg}.}
         \label{fig:dist-30k}
     \end{subfigure}
     \begin{subfigure}[b]{0.48\columnwidth}
         \centering
         \includegraphics[width=0.8\textwidth]{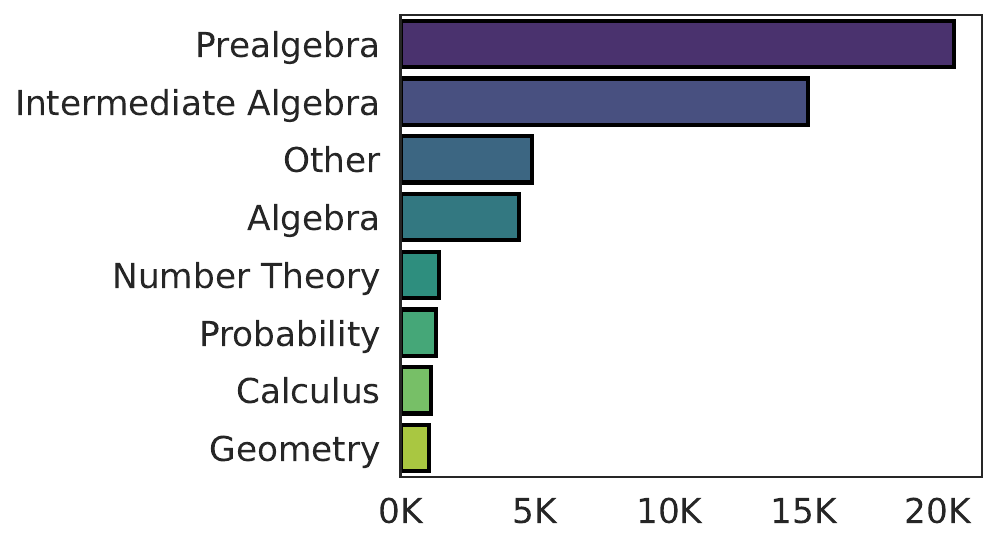}
         \caption{Topic distribution of 50K data selected by {\alg}.}
         \label{fig:dist-50k}
     \end{subfigure}
     \vskip 0.2in
    \begin{subfigure}[b]{0.6\columnwidth}
         \centering
         \includegraphics[width=\textwidth]{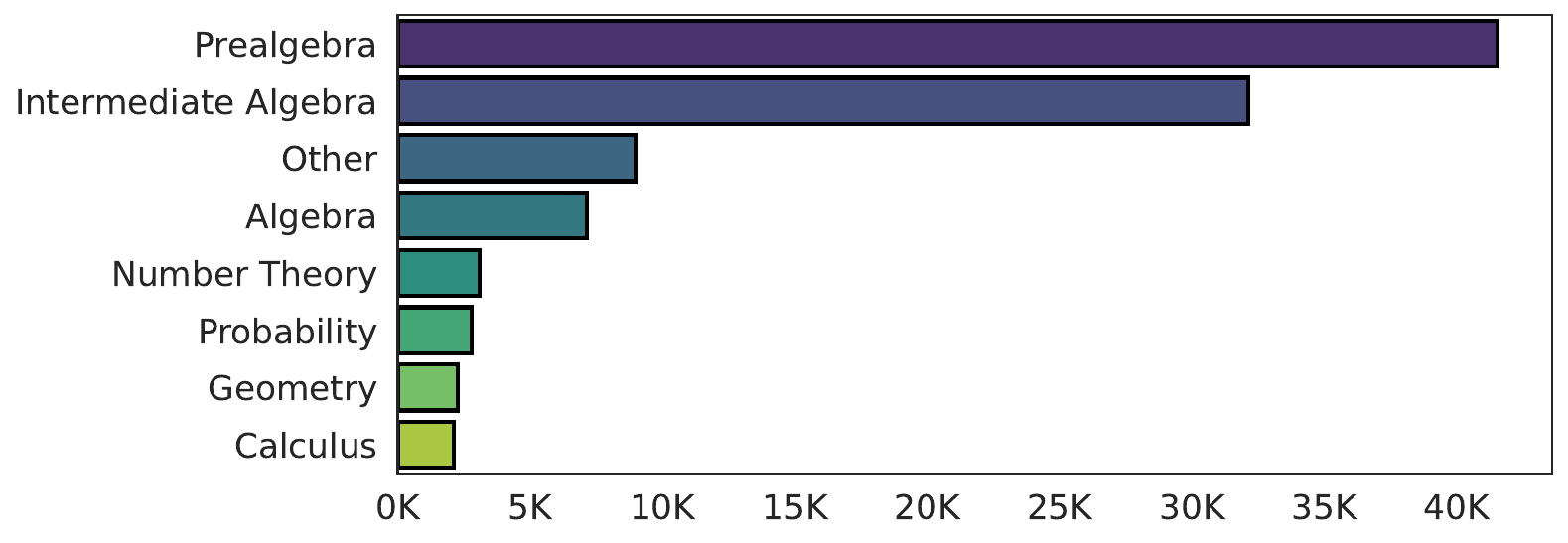}
         \caption{Topic distribution of 100K data selected by {\alg}.}
         \label{fig:dist-100k}
     \end{subfigure}
\caption{Compared to the original topic distribution, {\alg} prioritized easier topics (e.g., pre-algebra over intermediate algebra, algebra over other more advanced topics) while always ensuring complete and more balanced coverage of all topics.}
\label{fig:topic-dist}
\end{center}
\vskip -0.2in
\end{figure}

Beyond qualitative examples from different clusters, we study how {\alg} changes the data distribution to outperform using the full fine-tuning dataset as well as using random subsets of the same size that have the same distribution as the original dataset. In \cref{fig:topic-dist}, we can observe that {\alg} not only guarantees a thorough and balanced coverage across the spectrum of topics but also ensures sufficient representation of foundational topics, such as pre-algebra, which lays the groundwork for tackling more complex subjects. 

\section{Broader Impacts}\label{sec:impact}
This paper introduces a data selection method for large language models (LLMs), aiming to enhance the data efficiency in the supervised fine-tuning (SFT) of these models. 

\textbf{Positive Impacts:} Our method, by reducing the data requirements for training LLMs, can make fine-tuning LLMs more effective and accessible. This could lead to broader participation in AI research and application development across various fields, including healthcare and education. 

\textbf{Negative Impacts:} Our method does not inherently involve or encourage applications with direct negative societal impacts. The focus is on a generic improvement in the field of machine learning, particularly in the training of LLMs.

%% file: main.bbl
\begin{thebibliography}{68}
\providecommand{\natexlab}[1]{#1}
\providecommand{\url}[1]{\texttt{#1}}
\expandafter\ifx\csname urlstyle\endcsname\relax
  \providecommand{\doi}[1]{doi: #1}\else
  \providecommand{\doi}{doi: \begingroup \urlstyle{rm}\Url}\fi

\bibitem[Abbas et~al.(2023)Abbas, Tirumala, Simig, Ganguli, and Morcos]{abbas2023semdedup}
Amro Kamal~Mohamed Abbas, Kushal Tirumala, Daniel Simig, Surya Ganguli, and Ari~S Morcos.
\newblock Semdedup: Data-efficient learning at web-scale through semantic deduplication.
\newblock In \emph{ICLR 2023 Workshop on Multimodal Representation Learning: Perks and Pitfalls}, 2023.

\bibitem[Achiam et~al.(2023)Achiam, Adler, Agarwal, Ahmad, Akkaya, Aleman, Almeida, Altenschmidt, Altman, Anadkat, Avila, Babuschkin, Balaji, Balcom, Baltescu, Bao, Bavarian, Belgum, Bello, Berdine, Bernadett-Shapiro, Berner, Bogdonoff, Boiko, Boyd, Brakman, Brockman, Brooks, Brundage, Button, Cai, Campbell, Cann, Carey, Carlson, Carmichael, Chan, Chang, Chantzis, Chen, Chen, Chen, Chen, Chen, Chess, Cho, Chu, Chung, Cummings, Currier, Dai, Decareaux, Degry, Deutsch, Deville, Dhar, Dohan, Dowling, Dunning, Ecoffet, Eleti, Eloundou, Farhi, Fedus, Felix, Fishman, Forte, Fulford, Gao, Georges, Gibson, Goel, Gogineni, Goh, Gontijo-Lopes, Gordon, Grafstein, Gray, Greene, Gross, Gu, Guo, Hallacy, Han, Harris, He, Heaton, Heidecke, Hesse, Hickey, Hickey, Hoeschele, Houghton, Hsu, Hu, Hu, Huizinga, Jain, Jain, Jang, Jiang, Jiang, Jin, Jin, Jomoto, Jonn, Jun, Kaftan, Kaiser, Kamali, Kanitscheider, Keskar, Khan, Kilpatrick, Kim, Kim, Kim, Kirchner, Kiros, Knight, Kokotajlo, Kondraciuk, Kondrich, Konstantinidis, Kosic,
  Krueger, Kuo, Lampe, Lan, Lee, Leike, Leung, Levy, Li, Lim, Lin, Lin, Litwin, Lopez, Lowe, Lue, Makanju, Malfacini, Manning, Markov, Markovski, Martin, Mayer, Mayne, McGrew, McKinney, McLeavey, McMillan, McNeil, Medina, Mehta, Menick, Metz, Mishchenko, Mishkin, Monaco, Morikawa, Mossing, Mu, Murati, Murk, M'ely, Nair, Nakano, Nayak, Neelakantan, Ngo, Noh, Long, O'Keefe, Pachocki, Paino, Palermo, Pantuliano, Parascandolo, Parish, Parparita, Passos, Pavlov, Peng, Perelman, de~Avila Belbute~Peres, Petrov, de~Oliveira~Pinto, Pokorny, Pokrass, Pong, Powell, Power, Power, Proehl, Puri, Radford, Rae, Ramesh, Raymond, Real, Rimbach, Ross, Rotsted, Roussez, Ryder, Saltarelli, Sanders, Santurkar, Sastry, Schmidt, Schnurr, Schulman, Selsam, Sheppard, Sherbakov, Shieh, Shoker, Shyam, Sidor, Sigler, Simens, Sitkin, Slama, Sohl, Sokolowsky, Song, Staudacher, Such, Summers, Sutskever, Tang, Tezak, Thompson, Tillet, Tootoonchian, Tseng, Tuggle, Turley, Tworek, Uribe, Vallone, Vijayvergiya, Voss, Wainwright, Wang, Wang,
  Wang, Ward, Wei, Weinmann, Welihinda, Welinder, Weng, Weng, Wiethoff, Willner, Winter, Wolrich, Wong, Workman, Wu, Wu, Wu, Xiao, Xu, Yoo, Yu, Yuan, Zaremba, Zellers, Zhang, Zhang, Zhao, Zheng, Zhuang, Zhuk, and Zoph]{Achiam2023GPT4TR}
OpenAI~Josh Achiam, Steven Adler, Sandhini Agarwal, Lama Ahmad, Ilge Akkaya, Florencia~Leoni Aleman, Diogo Almeida, Janko Altenschmidt, Sam Altman, Shyamal Anadkat, Red Avila, Igor Babuschkin, Suchir Balaji, Valerie Balcom, Paul Baltescu, Haiming Bao, Mo~Bavarian, Jeff Belgum, Irwan Bello, Jake Berdine, Gabriel Bernadett-Shapiro, Christopher Berner, Lenny Bogdonoff, Oleg Boiko, Madelaine Boyd, Anna-Luisa Brakman, Greg Brockman, Tim Brooks, Miles Brundage, Kevin Button, Trevor Cai, Rosie Campbell, Andrew Cann, Brittany Carey, Chelsea Carlson, Rory Carmichael, Brooke Chan, Che Chang, Fotis Chantzis, Derek Chen, Sully Chen, Ruby Chen, Jason Chen, Mark Chen, Benjamin Chess, Chester Cho, Casey Chu, Hyung~Won Chung, Dave Cummings, Jeremiah Currier, Yunxing Dai, Cory Decareaux, Thomas Degry, Noah Deutsch, Damien Deville, Arka Dhar, David Dohan, Steve Dowling, Sheila Dunning, Adrien Ecoffet, Atty Eleti, Tyna Eloundou, David Farhi, Liam Fedus, Niko Felix, Sim'on~Posada Fishman, Juston Forte, Isabella Fulford, Leo Gao,
  Elie Georges, Christian Gibson, Vik Goel, Tarun Gogineni, Gabriel Goh, Raphael Gontijo-Lopes, Jonathan Gordon, Morgan Grafstein, Scott Gray, Ryan Greene, Joshua Gross, Shixiang~Shane Gu, Yufei Guo, Chris Hallacy, Jesse Han, Jeff Harris, Yuchen He, Mike Heaton, Johannes Heidecke, Chris Hesse, Alan Hickey, Wade Hickey, Peter Hoeschele, Brandon Houghton, Kenny Hsu, Shengli Hu, Xin Hu, Joost Huizinga, Shantanu Jain, Shawn Jain, Joanne Jang, Angela Jiang, Roger Jiang, Haozhun Jin, Denny Jin, Shino Jomoto, Billie Jonn, Heewoo Jun, Tomer Kaftan, Lukasz Kaiser, Ali Kamali, Ingmar Kanitscheider, Nitish~Shirish Keskar, Tabarak Khan, Logan Kilpatrick, Jong~Wook Kim, Christina Kim, Yongjik Kim, Hendrik Kirchner, Jamie~Ryan Kiros, Matthew Knight, Daniel Kokotajlo, Lukasz Kondraciuk, Andrew Kondrich, Aris Konstantinidis, Kyle Kosic, Gretchen Krueger, Vishal Kuo, Michael Lampe, Ikai Lan, Teddy Lee, Jan Leike, Jade Leung, Daniel Levy, Chak~Ming Li, Rachel Lim, Molly Lin, Stephanie Lin, Mateusz Litwin, Theresa Lopez, Ryan
  Lowe, Patricia Lue, Anna~Adeola Makanju, Kim Malfacini, Sam Manning, Todor Markov, Yaniv Markovski, Bianca Martin, Katie Mayer, Andrew Mayne, Bob McGrew, Scott~Mayer McKinney, Christine McLeavey, Paul McMillan, Jake McNeil, David Medina, Aalok Mehta, Jacob Menick, Luke Metz, Andrey Mishchenko, Pamela Mishkin, Vinnie Monaco, Evan Morikawa, Daniel~P. Mossing, Tong Mu, Mira Murati, Oleg Murk, David M'ely, Ashvin Nair, Reiichiro Nakano, Rajeev Nayak, Arvind Neelakantan, Richard Ngo, Hyeonwoo Noh, Ouyang Long, Cullen O'Keefe, Jakub~W. Pachocki, Alex Paino, Joe Palermo, Ashley Pantuliano, Giambattista Parascandolo, Joel Parish, Emy Parparita, Alexandre Passos, Mikhail Pavlov, Andrew Peng, Adam Perelman, Filipe de~Avila Belbute~Peres, Michael Petrov, Henrique~Pond{\'e} de~Oliveira~Pinto, Michael Pokorny, Michelle Pokrass, Vitchyr~H. Pong, Tolly Powell, Alethea Power, Boris Power, Elizabeth Proehl, Raul Puri, Alec Radford, Jack Rae, Aditya Ramesh, Cameron Raymond, Francis Real, Kendra Rimbach, Carl Ross, Bob
  Rotsted, Henri Roussez, Nick Ryder, Mario~D. Saltarelli, Ted Sanders, Shibani Santurkar, Girish Sastry, Heather Schmidt, David Schnurr, John Schulman, Daniel Selsam, Kyla Sheppard, Toki Sherbakov, Jessica Shieh, Sarah Shoker, Pranav Shyam, Szymon Sidor, Eric Sigler, Maddie Simens, Jordan Sitkin, Katarina Slama, Ian Sohl, Benjamin~D. Sokolowsky, Yang Song, Natalie Staudacher, Felipe~Petroski Such, Natalie Summers, Ilya Sutskever, Jie Tang, Nikolas~A. Tezak, Madeleine Thompson, Phil Tillet, Amin Tootoonchian, Elizabeth Tseng, Preston Tuggle, Nick Turley, Jerry Tworek, Juan Felipe~Cer'on Uribe, Andrea Vallone, Arun Vijayvergiya, Chelsea Voss, Carroll Wainwright, Justin~Jay Wang, Alvin Wang, Ben Wang, Jonathan Ward, Jason Wei, CJ~Weinmann, Akila Welihinda, Peter Welinder, Jiayi Weng, Lilian Weng, Matt Wiethoff, Dave Willner, Clemens Winter, Samuel Wolrich, Hannah Wong, Lauren Workman, Sherwin Wu, Jeff Wu, Michael Wu, Kai Xiao, Tao Xu, Sarah Yoo, Kevin Yu, Qiming Yuan, Wojciech Zaremba, Rowan Zellers, Chong
  Zhang, Marvin Zhang, Shengjia Zhao, Tianhao Zheng, Juntang Zhuang, William Zhuk, and Barret Zoph.
\newblock Gpt-4 technical report.
\newblock 2023.
\newblock URL \url{https://api.semanticscholar.org/CorpusID:257532815}.

\bibitem[Azerbayev et~al.(2023)Azerbayev, Schoelkopf, Paster, Santos, McAleer, Jiang, Deng, Biderman, and Welleck]{azerbayev2023llemma}
Zhangir Azerbayev, Hailey Schoelkopf, Keiran Paster, Marco~Dos Santos, Stephen McAleer, Albert~Q Jiang, Jia Deng, Stella Biderman, and Sean Welleck.
\newblock Llemma: An open language model for mathematics.
\newblock \emph{arXiv preprint arXiv:2310.10631}, 2023.

\bibitem[Bhatt et~al.(2024)Bhatt, Chen, Das, Zhang, Truong, Mussmann, Zhu, Bilmes, Du, Jamieson, et~al.]{bhatt2024experimental}
Gantavya Bhatt, Yifang Chen, Arnav~M Das, Jifan Zhang, Sang~T Truong, Stephen Mussmann, Yinglun Zhu, Jeffrey Bilmes, Simon~S Du, Kevin Jamieson, et~al.
\newblock An experimental design framework for label-efficient supervised finetuning of large language models.
\newblock \emph{arXiv preprint arXiv:2401.06692}, 2024.

\bibitem[Biderman et~al.(2023{\natexlab{a}})Biderman, Prashanth, Sutawika, Schoelkopf, Anthony, Purohit, and Raf]{biderman2023emergent}
Stella Biderman, USVSN~Sai Prashanth, Lintang Sutawika, Hailey Schoelkopf, Quentin Anthony, Shivanshu Purohit, and Edward Raf.
\newblock Emergent and predictable memorization in large language models.
\newblock \emph{arXiv preprint arXiv:2304.11158}, 2023{\natexlab{a}}.

\bibitem[Biderman et~al.(2023{\natexlab{b}})Biderman, Schoelkopf, Anthony, Bradley, O’Brien, Hallahan, Khan, Purohit, Prashanth, Raff, et~al.]{biderman2023pythia}
Stella Biderman, Hailey Schoelkopf, Quentin~Gregory Anthony, Herbie Bradley, Kyle O’Brien, Eric Hallahan, Mohammad~Aflah Khan, Shivanshu Purohit, USVSN~Sai Prashanth, Edward Raff, et~al.
\newblock Pythia: A suite for analyzing large language models across training and scaling.
\newblock In \emph{International Conference on Machine Learning}, pages 2397--2430. PMLR, 2023{\natexlab{b}}.

\bibitem[Brown et~al.(2020)Brown, Mann, Ryder, Subbiah, Kaplan, Dhariwal, Neelakantan, Shyam, Sastry, Askell, Agarwal, Herbert-Voss, Krueger, Henighan, Child, Ramesh, Ziegler, Wu, Winter, Hesse, Chen, Sigler, Litwin, Gray, Chess, Clark, Berner, McCandlish, Radford, Sutskever, and Amodei]{brown2020gpt}
Tom Brown, Benjamin Mann, Nick Ryder, Melanie Subbiah, Jared~D Kaplan, Prafulla Dhariwal, Arvind Neelakantan, Pranav Shyam, Girish Sastry, Amanda Askell, Sandhini Agarwal, Ariel Herbert-Voss, Gretchen Krueger, Tom Henighan, Rewon Child, Aditya Ramesh, Daniel Ziegler, Jeffrey Wu, Clemens Winter, Chris Hesse, Mark Chen, Eric Sigler, Mateusz Litwin, Scott Gray, Benjamin Chess, Jack Clark, Christopher Berner, Sam McCandlish, Alec Radford, Ilya Sutskever, and Dario Amodei.
\newblock Language models are few-shot learners.
\newblock In H.~Larochelle, M.~Ranzato, R.~Hadsell, M.F. Balcan, and H.~Lin, editors, \emph{Advances in Neural Information Processing Systems}, volume~33, pages 1877--1901. Curran Associates, Inc., 2020.
\newblock URL \url{https://proceedings.neurips.cc/paper_files/paper/2020/file/1457c0d6bfcb4967418bfb8ac142f64a-Paper.pdf}.

\bibitem[Chen et~al.(2024)Chen, Li, Yan, Wang, Gunaratna, Yadav, Tang, Srinivasan, Zhou, Huang, et~al.]{chen2023alpagasus}
Lichang Chen, Shiyang Li, Jun Yan, Hai Wang, Kalpa Gunaratna, Vikas Yadav, Zheng Tang, Vijay Srinivasan, Tianyi Zhou, Heng Huang, et~al.
\newblock Alpagasus: Training a better alpaca with fewer data.
\newblock In \emph{The Twelfth International Conference on Learning Representations}, 2024.
\newblock URL \url{https://openreview.net/forum?id=FdVXgSJhvz}.

\bibitem[Cheng et~al.(2023)Cheng, Huang, and Wei]{cheng2023adapting}
Daixuan Cheng, Shaohan Huang, and Furu Wei.
\newblock Adapting large language models via reading comprehension.
\newblock \emph{arXiv preprint arXiv:2309.09530}, 2023.

\bibitem[Chowdhery et~al.(2023)Chowdhery, Narang, Devlin, Bosma, Mishra, Roberts, Barham, Chung, Sutton, Gehrmann, et~al.]{chowdhery2023palm}
Aakanksha Chowdhery, Sharan Narang, Jacob Devlin, Maarten Bosma, Gaurav Mishra, Adam Roberts, Paul Barham, Hyung~Won Chung, Charles Sutton, Sebastian Gehrmann, et~al.
\newblock Palm: Scaling language modeling with pathways.
\newblock \emph{Journal of Machine Learning Research}, 24\penalty0 (240):\penalty0 1--113, 2023.

\bibitem[Cobbe et~al.(2021)Cobbe, Kosaraju, Bavarian, Chen, Jun, Kaiser, Plappert, Tworek, Hilton, Nakano, Hesse, and Schulman]{cobbe2021gsm8k}
Karl Cobbe, Vineet Kosaraju, Mohammad Bavarian, Mark Chen, Heewoo Jun, Lukasz Kaiser, Matthias Plappert, Jerry Tworek, Jacob Hilton, Reiichiro Nakano, Christopher Hesse, and John Schulman.
\newblock Training verifiers to solve math word problems.
\newblock \emph{arXiv preprint arXiv:2110.14168}, 2021.

\bibitem[Coleman et~al.(2020)Coleman, Yeh, Mussmann, Mirzasoleiman, Bailis, Liang, Leskovec, and Zaharia]{Coleman2020Selection}
Cody Coleman, Christopher Yeh, Stephen Mussmann, Baharan Mirzasoleiman, Peter Bailis, Percy Liang, Jure Leskovec, and Matei Zaharia.
\newblock Selection via proxy: Efficient data selection for deep learning.
\newblock In \emph{International Conference on Learning Representations}, 2020.
\newblock URL \url{https://openreview.net/forum?id=HJg2b0VYDr}.

\bibitem[Davies et~al.(2021)Davies, Veli{\v{c}}kovi{\'c}, Buesing, Blackwell, Zheng, Toma{\v{s}}ev, Tanburn, Battaglia, Blundell, Juh{\'a}sz, et~al.]{davies2021advancing}
Alex Davies, Petar Veli{\v{c}}kovi{\'c}, Lars Buesing, Sam Blackwell, Daniel Zheng, Nenad Toma{\v{s}}ev, Richard Tanburn, Peter Battaglia, Charles Blundell, Andr{\'a}s Juh{\'a}sz, et~al.
\newblock Advancing mathematics by guiding human intuition with ai.
\newblock \emph{Nature}, 600\penalty0 (7887):\penalty0 70--74, 2021.

\bibitem[Delbrouck et~al.(2023)Delbrouck, Varma, Chambon, and Langlotz]{delbrouck-etal-2023-overview}
Jean-Benoit Delbrouck, Maya Varma, Pierre Chambon, and Curtis Langlotz.
\newblock Overview of the {R}ad{S}um23 shared task on multi-modal and multi-anatomical radiology report summarization.
\newblock In Dina Demner-fushman, Sophia Ananiadou, and Kevin Cohen, editors, \emph{The 22nd Workshop on Biomedical Natural Language Processing and BioNLP Shared Tasks}, pages 478--482, Toronto, Canada, July 2023. Association for Computational Linguistics.
\newblock \doi{10.18653/v1/2023.bionlp-1.45}.
\newblock URL \url{https://aclanthology.org/2023.bionlp-1.45}.

\bibitem[Demner-fushman et~al.(2023)Demner-fushman, Ananiadou, and Cohen]{bionlp-2023-biomedical}
Dina Demner-fushman, Sophia Ananiadou, and Kevin Cohen, editors.
\newblock \emph{The 22nd Workshop on Biomedical Natural Language Processing and BioNLP Shared Tasks}, Toronto, Canada, July 2023. Association for Computational Linguistics.
\newblock URL \url{https://aclanthology.org/2023.bionlp-1.0}.

\bibitem[Deng et~al.(2023)Deng, Yang, Mirzasoleiman, and Gu]{deng2023robust}
Yihe Deng, Yu~Yang, Baharan Mirzasoleiman, and Quanquan Gu.
\newblock Robust learning with progressive data expansion against spurious correlation.
\newblock In \emph{Thirty-seventh Conference on Neural Information Processing Systems}, 2023.
\newblock URL \url{https://openreview.net/forum?id=9QEVJ9qm46}.

\bibitem[Douze et~al.(2024)Douze, Guzhva, Deng, Johnson, Szilvasy, Mazaré, Lomeli, Hosseini, and Jégou]{douze2024faiss}
Matthijs Douze, Alexandr Guzhva, Chengqi Deng, Jeff Johnson, Gergely Szilvasy, Pierre-Emmanuel Mazaré, Maria Lomeli, Lucas Hosseini, and Hervé Jégou.
\newblock The faiss library.
\newblock 2024.

\bibitem[Eldan and Li(2023)]{eldan2023tinystories}
Ronen Eldan and Yuanzhi Li.
\newblock Tinystories: How small can language models be and still speak coherent english?
\newblock \emph{arXiv preprint arXiv:2305.07759}, 2023.

\bibitem[Hendrycks et~al.(2021)Hendrycks, Burns, Kadavath, Arora, Basart, Tang, Song, and Steinhardt]{hendrycks2021measuring}
Dan Hendrycks, Collin Burns, Saurav Kadavath, Akul Arora, Steven Basart, Eric Tang, Dawn Song, and Jacob Steinhardt.
\newblock Measuring mathematical problem solving with the {MATH} dataset.
\newblock In \emph{Thirty-fifth Conference on Neural Information Processing Systems Datasets and Benchmarks Track (Round 2)}, 2021.
\newblock URL \url{https://openreview.net/forum?id=7Bywt2mQsCe}.

\bibitem[Jang et~al.(2023)Jang, Kim, Ye, Kim, Logeswaran, Lee, Lee, and Seo]{pmlr-v202-jang23a}
Joel Jang, Seungone Kim, Seonghyeon Ye, Doyoung Kim, Lajanugen Logeswaran, Moontae Lee, Kyungjae Lee, and Minjoon Seo.
\newblock Exploring the benefits of training expert language models over instruction tuning.
\newblock In Andreas Krause, Emma Brunskill, Kyunghyun Cho, Barbara Engelhardt, Sivan Sabato, and Jonathan Scarlett, editors, \emph{Proceedings of the 40th International Conference on Machine Learning}, volume 202 of \emph{Proceedings of Machine Learning Research}, pages 14702--14729. PMLR, 23--29 Jul 2023.
\newblock URL \url{https://proceedings.mlr.press/v202/jang23a.html}.

\bibitem[Johnson et~al.(2016)Johnson, Pollard, Shen, Lehman, Feng, Ghassemi, Moody, Szolovits, Anthony~Celi, and Mark]{johnson2016mimic}
Alistair~EW Johnson, Tom~J Pollard, Lu~Shen, Li-wei~H Lehman, Mengling Feng, Mohammad Ghassemi, Benjamin Moody, Peter Szolovits, Leo Anthony~Celi, and Roger~G Mark.
\newblock Mimic-iii, a freely accessible critical care database.
\newblock \emph{Scientific data}, 3\penalty0 (1):\penalty0 1--9, 2016.

\bibitem[Joshi and Mirzasoleiman(2023)]{joshi2023data}
Siddharth Joshi and Baharan Mirzasoleiman.
\newblock Data-efficient contrastive self-supervised learning: Most beneficial examples for supervised learning contribute the least.
\newblock In \emph{International conference on machine learning}, pages 15356--15370. PMLR, 2023.

\bibitem[Killamsetty et~al.(2021{\natexlab{a}})Killamsetty, Durga, Ramakrishnan, De, and Iyer]{killamsetty2021grad}
Krishnateja Killamsetty, Sivasubramanian Durga, Ganesh Ramakrishnan, Abir De, and Rishabh Iyer.
\newblock Grad-match: Gradient matching based data subset selection for efficient deep model training.
\newblock In \emph{International Conference on Machine Learning}, pages 5464--5474. PMLR, 2021{\natexlab{a}}.

\bibitem[Killamsetty et~al.(2021{\natexlab{b}})Killamsetty, Sivasubramanian, Ramakrishnan, and Iyer]{killamsetty2021glister}
Krishnateja Killamsetty, Durga Sivasubramanian, Ganesh Ramakrishnan, and Rishabh Iyer.
\newblock Glister: Generalization based data subset selection for efficient and robust learning.
\newblock In \emph{Proceedings of the AAAI Conference on Artificial Intelligence}, volume~35, pages 8110--8118, 2021{\natexlab{b}}.

\bibitem[Koncel-Kedziorski et~al.(2016)Koncel-Kedziorski, Roy, Amini, Kushman, and Hajishirzi]{koncel-kedziorski-etal-2016-mawps}
Rik Koncel-Kedziorski, Subhro Roy, Aida Amini, Nate Kushman, and Hannaneh Hajishirzi.
\newblock {MAWPS}: A math word problem repository.
\newblock In Kevin Knight, Ani Nenkova, and Owen Rambow, editors, \emph{Proceedings of the 2016 Conference of the North {A}merican Chapter of the Association for Computational Linguistics: Human Language Technologies}, pages 1152--1157, San Diego, California, June 2016. Association for Computational Linguistics.
\newblock \doi{10.18653/v1/N16-1136}.
\newblock URL \url{https://aclanthology.org/N16-1136}.

\bibitem[Lewkowycz et~al.(2022)Lewkowycz, Andreassen, Dohan, Dyer, Michalewski, Ramasesh, Slone, Anil, Schlag, Gutman-Solo, et~al.]{lewkowycz2022solving}
Aitor Lewkowycz, Anders Andreassen, David Dohan, Ethan Dyer, Henryk Michalewski, Vinay Ramasesh, Ambrose Slone, Cem Anil, Imanol Schlag, Theo Gutman-Solo, et~al.
\newblock Solving quantitative reasoning problems with language models.
\newblock \emph{Advances in Neural Information Processing Systems}, 35:\penalty0 3843--3857, 2022.

\bibitem[Li et~al.(2023{\natexlab{a}})Li, Bubeck, Eldan, Del~Giorno, Gunasekar, and Lee]{li2023textbooks}
Yuanzhi Li, S{\'e}bastien Bubeck, Ronen Eldan, Allie Del~Giorno, Suriya Gunasekar, and Yin~Tat Lee.
\newblock Textbooks are all you need ii: phi-1.5 technical report.
\newblock \emph{arXiv preprint arXiv:2309.05463}, 2023{\natexlab{a}}.

\bibitem[Li et~al.(2023{\natexlab{b}})Li, Bubeck, Eldan, Del~Giorno, Gunasekar, and Lee]{textbooks2}
Yuanzhi Li, S{\'e}bastien Bubeck, Ronen Eldan, Allie Del~Giorno, Suriya Gunasekar, and Yin~Tat Lee.
\newblock Textbooks are all you need ii: \textbf{phi-1.5} technical report.
\newblock \emph{arXiv preprint arXiv:2309.05463}, 2023{\natexlab{b}}.

\bibitem[Lin(2004)]{lin-2004-rouge}
Chin-Yew Lin.
\newblock {ROUGE}: A package for automatic evaluation of summaries.
\newblock In \emph{Text Summarization Branches Out}, pages 74--81, Barcelona, Spain, July 2004. Association for Computational Linguistics.
\newblock URL \url{https://aclanthology.org/W04-1013}.

\bibitem[Liu et~al.(2023)Liu, Bubeck, Eldan, Kulkarni, Li, Nguyen, Ward, and Zhang]{liu2023tinygsm}
Bingbin Liu, Sebastien Bubeck, Ronen Eldan, Janardhan Kulkarni, Yuanzhi Li, Anh Nguyen, Rachel Ward, and Yi~Zhang.
\newblock Tinygsm: achieving> 80\% on gsm8k with small language models.
\newblock \emph{arXiv preprint arXiv:2312.09241}, 2023.

\bibitem[Luo et~al.(2023{\natexlab{a}})Luo, Sun, Xu, Zhao, Lou, Tao, Geng, Lin, Chen, and Zhang]{luo2023wizardmath}
Haipeng Luo, Qingfeng Sun, Can Xu, Pu~Zhao, Jianguang Lou, Chongyang Tao, Xiubo Geng, Qingwei Lin, Shifeng Chen, and Dongmei Zhang.
\newblock Wizardmath: Empowering mathematical reasoning for large language models via reinforced evol-instruct.
\newblock \emph{arXiv preprint arXiv:2308.09583}, 2023{\natexlab{a}}.

\bibitem[Luo et~al.(2023{\natexlab{b}})Luo, Xu, Zhao, Sun, Geng, Hu, Tao, Ma, Lin, and Jiang]{luo2023wizardcoder}
Ziyang Luo, Can Xu, Pu~Zhao, Qingfeng Sun, Xiubo Geng, Wenxiang Hu, Chongyang Tao, Jing Ma, Qingwei Lin, and Daxin Jiang.
\newblock Wizardcoder: Empowering code large language models with evol-instruct.
\newblock \emph{arXiv preprint arXiv:2306.08568}, 2023{\natexlab{b}}.

\bibitem[Mahmoud et~al.(2024)Mahmoud, Elhoushi, Abbas, Yang, Ardalani, Leather, and Morcos]{mahmoud2023sieve}
Anas Mahmoud, Mostafa Elhoushi, Amro Abbas, Yu~Yang, Newsha Ardalani, Hugh Leather, and Ari Morcos.
\newblock Sieve: Multimodal dataset pruning using image-captioning models.
\newblock In \emph{Conference on Computer Vision and Pattern Recognition}, 2024.
\newblock URL \url{https://openreview.net/forum?id=DBxBPGRWjw}.

\bibitem[Marion et~al.(2023)Marion, {\"U}st{\"u}n, Pozzobon, Wang, Fadaee, and Hooker]{marion2023less}
Max Marion, Ahmet {\"U}st{\"u}n, Luiza Pozzobon, Alex Wang, Marzieh Fadaee, and Sara Hooker.
\newblock When less is more: Investigating data pruning for pretraining llms at scale.
\newblock \emph{arXiv preprint arXiv:2309.04564}, 2023.

\bibitem[Mirzasoleiman et~al.(2020)Mirzasoleiman, Bilmes, and Leskovec]{pmlr-v119-mirzasoleiman20a}
Baharan Mirzasoleiman, Jeff Bilmes, and Jure Leskovec.
\newblock Coresets for data-efficient training of machine learning models.
\newblock In Hal~Daumé III and Aarti Singh, editors, \emph{Proceedings of the 37th International Conference on Machine Learning}, volume 119 of \emph{Proceedings of Machine Learning Research}, pages 6950--6960. PMLR, 13--18 Jul 2020.
\newblock URL \url{https://proceedings.mlr.press/v119/mirzasoleiman20a.html}.

\bibitem[Mishra et~al.(2022)Mishra, Mitra, Varshney, Sachdeva, Clark, Baral, and Kalyan]{mishra-etal-2022-numglue}
Swaroop Mishra, Arindam Mitra, Neeraj Varshney, Bhavdeep Sachdeva, Peter Clark, Chitta Baral, and Ashwin Kalyan.
\newblock {N}um{GLUE}: A suite of fundamental yet challenging mathematical reasoning tasks.
\newblock In Smaranda Muresan, Preslav Nakov, and Aline Villavicencio, editors, \emph{Proceedings of the 60th Annual Meeting of the Association for Computational Linguistics (Volume 1: Long Papers)}, pages 3505--3523, Dublin, Ireland, May 2022. Association for Computational Linguistics.
\newblock \doi{10.18653/v1/2022.acl-long.246}.
\newblock URL \url{https://aclanthology.org/2022.acl-long.246}.

\bibitem[Papineni et~al.(2002)Papineni, Roukos, Ward, and Zhu]{papineni2002bleu}
Kishore Papineni, Salim Roukos, Todd Ward, and Wei-Jing Zhu.
\newblock Bleu: a method for automatic evaluation of machine translation.
\newblock In \emph{Proceedings of the 40th annual meeting of the Association for Computational Linguistics}, pages 311--318, 2002.

\bibitem[Patel et~al.(2021)Patel, Bhattamishra, and Goyal]{patel-etal-2021-nlp}
Arkil Patel, Satwik Bhattamishra, and Navin Goyal.
\newblock Are {NLP} models really able to solve simple math word problems?
\newblock In \emph{Proceedings of the 2021 Conference of the North American Chapter of the Association for Computational Linguistics: Human Language Technologies}, pages 2080--2094, Online, June 2021. Association for Computational Linguistics.
\newblock \doi{10.18653/v1/2021.naacl-main.168}.
\newblock URL \url{https://aclanthology.org/2021.naacl-main.168}.

\bibitem[Paul et~al.(2021)Paul, Ganguli, and Dziugaite]{paul2021deep}
Mansheej Paul, Surya Ganguli, and Gintare~Karolina Dziugaite.
\newblock Deep learning on a data diet: Finding important examples early in training.
\newblock \emph{Advances in Neural Information Processing Systems}, 34:\penalty0 20596--20607, 2021.

\bibitem[Pooladzandi et~al.(2022)Pooladzandi, Davini, and Mirzasoleiman]{pmlr-v162-pooladzandi22a}
Omead Pooladzandi, David Davini, and Baharan Mirzasoleiman.
\newblock Adaptive second order coresets for data-efficient machine learning.
\newblock In Kamalika Chaudhuri, Stefanie Jegelka, Le~Song, Csaba Szepesvari, Gang Niu, and Sivan Sabato, editors, \emph{Proceedings of the 39th International Conference on Machine Learning}, volume 162 of \emph{Proceedings of Machine Learning Research}, pages 17848--17869. PMLR, 17--23 Jul 2022.
\newblock URL \url{https://proceedings.mlr.press/v162/pooladzandi22a.html}.

\bibitem[Prakriya et~al.(2023)Prakriya, Yang, Mirzasoleiman, Hsieh, and Cong]{nessa}
Neha Prakriya, Yu~Yang, Baharan Mirzasoleiman, Cho-Jui Hsieh, and Jason Cong.
\newblock Nessa: Near-storage data selection for accelerated machine learning training.
\newblock In \emph{Proceedings of the 15th ACM Workshop on Hot Topics in Storage and File Systems}, HotStorage '23, page 8–15, New York, NY, USA, 2023. Association for Computing Machinery.
\newblock ISBN 9798400702242.
\newblock \doi{10.1145/3599691.3603404}.
\newblock URL \url{https://doi.org/10.1145/3599691.3603404}.

\bibitem[Roziere et~al.(2023)Roziere, Gehring, Gloeckle, Sootla, Gat, Tan, Adi, Liu, Remez, Rapin, et~al.]{roziere2023code}
Baptiste Roziere, Jonas Gehring, Fabian Gloeckle, Sten Sootla, Itai Gat, Xiaoqing~Ellen Tan, Yossi Adi, Jingyu Liu, Tal Remez, J{\'e}r{\'e}my Rapin, et~al.
\newblock Code llama: Open foundation models for code.
\newblock \emph{arXiv preprint arXiv:2308.12950}, 2023.

\bibitem[Singhal et~al.(2023{\natexlab{a}})Singhal, Azizi, Tu, Mahdavi, Wei, Chung, Scales, Tanwani, Cole-Lewis, Pfohl, et~al.]{singhal2023medpalm}
Karan Singhal, Shekoofeh Azizi, Tao Tu, S~Sara Mahdavi, Jason Wei, Hyung~Won Chung, Nathan Scales, Ajay Tanwani, Heather Cole-Lewis, Stephen Pfohl, et~al.
\newblock Large language models encode clinical knowledge.
\newblock \emph{Nature}, 620\penalty0 (7972):\penalty0 172--180, 2023{\natexlab{a}}.

\bibitem[Singhal et~al.(2023{\natexlab{b}})Singhal, Tu, Gottweis, Sayres, Wulczyn, Hou, Clark, Pfohl, Cole-Lewis, Neal, et~al.]{singhal2023medpalm2}
Karan Singhal, Tao Tu, Juraj Gottweis, Rory Sayres, Ellery Wulczyn, Le~Hou, Kevin Clark, Stephen Pfohl, Heather Cole-Lewis, Darlene Neal, et~al.
\newblock Towards expert-level medical question answering with large language models.
\newblock \emph{arXiv preprint arXiv:2305.09617}, 2023{\natexlab{b}}.

\bibitem[Sorscher et~al.(2022)Sorscher, Geirhos, Shekhar, Ganguli, and Morcos]{sorscher2022beyond}
Ben Sorscher, Robert Geirhos, Shashank Shekhar, Surya Ganguli, and Ari Morcos.
\newblock Beyond neural scaling laws: beating power law scaling via data pruning.
\newblock \emph{Advances in Neural Information Processing Systems}, 35:\penalty0 19523--19536, 2022.

\bibitem[Swayamdipta et~al.(2020)Swayamdipta, Schwartz, Lourie, Wang, Hajishirzi, Smith, and Choi]{swayamdipta-etal-2020-dataset}
Swabha Swayamdipta, Roy Schwartz, Nicholas Lourie, Yizhong Wang, Hannaneh Hajishirzi, Noah~A. Smith, and Yejin Choi.
\newblock Dataset cartography: Mapping and diagnosing datasets with training dynamics.
\newblock In Bonnie Webber, Trevor Cohn, Yulan He, and Yang Liu, editors, \emph{Proceedings of the 2020 Conference on Empirical Methods in Natural Language Processing (EMNLP)}, pages 9275--9293, Online, November 2020. Association for Computational Linguistics.
\newblock \doi{10.18653/v1/2020.emnlp-main.746}.
\newblock URL \url{https://aclanthology.org/2020.emnlp-main.746}.

\bibitem[Taylor et~al.(2022)Taylor, Kardas, Cucurull, Scialom, Hartshorn, Saravia, Poulton, Kerkez, and Stojnic]{taylor2022galactica}
Ross Taylor, Marcin Kardas, Guillem Cucurull, Thomas Scialom, Anthony Hartshorn, Elvis Saravia, Andrew Poulton, Viktor Kerkez, and Robert Stojnic.
\newblock Galactica: A large language model for science.
\newblock \emph{arXiv preprint arXiv:2211.09085}, 2022.

\bibitem[Tirumala et~al.(2023)Tirumala, Simig, Aghajanyan, and Morcos]{tirumala2023d}
Kushal Tirumala, Daniel Simig, Armen Aghajanyan, and Ari~S. Morcos.
\newblock D4: Improving {LLM} pretraining via document de-duplication and diversification.
\newblock In \emph{Thirty-seventh Conference on Neural Information Processing Systems Datasets and Benchmarks Track}, 2023.
\newblock URL \url{https://openreview.net/forum?id=CG0L2PFrb1}.

\bibitem[Toneva et~al.(2019)Toneva, Sordoni, des Combes, Trischler, Bengio, and Gordon]{toneva2018an}
Mariya Toneva, Alessandro Sordoni, Remi~Tachet des Combes, Adam Trischler, Yoshua Bengio, and Geoffrey~J. Gordon.
\newblock An empirical study of example forgetting during deep neural network learning.
\newblock In \emph{International Conference on Learning Representations}, 2019.
\newblock URL \url{https://openreview.net/forum?id=BJlxm30cKm}.

\bibitem[Touvron et~al.(2023)Touvron, Martin, Stone, Albert, Almahairi, Babaei, Bashlykov, Batra, Bhargava, Bhosale, et~al.]{touvron2023llama}
Hugo Touvron, Louis Martin, Kevin Stone, Peter Albert, Amjad Almahairi, Yasmine Babaei, Nikolay Bashlykov, Soumya Batra, Prajjwal Bhargava, Shruti Bhosale, et~al.
\newblock Llama 2: Open foundation and fine-tuned chat models.
\newblock \emph{arXiv preprint arXiv:2307.09288}, 2023.

\bibitem[Tu et~al.(2023)Tu, Azizi, Driess, Schaekermann, Amin, Chang, Carroll, Lau, Tanno, Ktena, et~al.]{tu2023towards}
Tao Tu, Shekoofeh Azizi, Danny Driess, Mike Schaekermann, Mohamed Amin, Pi-Chuan Chang, Andrew Carroll, Chuck Lau, Ryutaro Tanno, Ira Ktena, et~al.
\newblock Towards generalist biomedical ai.
\newblock \emph{arXiv preprint arXiv:2307.14334}, 2023.

\bibitem[Van~Veen et~al.(2023)Van~Veen, Van~Uden, Blankemeier, Delbrouck, Aali, Bluethgen, Pareek, Polacin, Collins, Ahuja, et~al.]{van2023clinical}
Dave Van~Veen, Cara Van~Uden, Louis Blankemeier, Jean-Benoit Delbrouck, Asad Aali, Christian Bluethgen, Anuj Pareek, Malgorzata Polacin, William Collins, Neera Ahuja, et~al.
\newblock Clinical text summarization: adapting large language models can outperform human experts.
\newblock \emph{arXiv preprint arXiv:2309.07430}, 2023.

\bibitem[Varshney et~al.(2022)Varshney, Mishra, and Baral]{varshney-mishra-and-chitta-baral-2022-model}
Neeraj Varshney, Swaroop Mishra, and Chitta Baral.
\newblock Let the model decide its curriculum for multitask learning.
\newblock In Colin Cherry, Angela Fan, George Foster, Gholamreza~(Reza) Haffari, Shahram Khadivi, Nanyun~(Violet) Peng, Xiang Ren, Ehsan Shareghi, and Swabha Swayamdipta, editors, \emph{Proceedings of the Third Workshop on Deep Learning for Low-Resource Natural Language Processing}, pages 117--125, Hybrid, July 2022. Association for Computational Linguistics.
\newblock \doi{10.18653/v1/2022.deeplo-1.13}.
\newblock URL \url{https://aclanthology.org/2022.deeplo-1.13}.

\bibitem[Wei et~al.(2022)Wei, Wang, Schuurmans, Bosma, brian ichter, Xia, Chi, Le, and Zhou]{wei2022chain}
Jason Wei, Xuezhi Wang, Dale Schuurmans, Maarten Bosma, brian ichter, Fei Xia, Ed~H. Chi, Quoc~V Le, and Denny Zhou.
\newblock Chain of thought prompting elicits reasoning in large language models.
\newblock In Alice~H. Oh, Alekh Agarwal, Danielle Belgrave, and Kyunghyun Cho, editors, \emph{Advances in Neural Information Processing Systems}, 2022.
\newblock URL \url{https://openreview.net/forum?id=_VjQlMeSB_J}.

\bibitem[Wolf et~al.(2020)Wolf, Debut, Sanh, Chaumond, Delangue, Moi, Cistac, Rault, Louf, Funtowicz, Davison, Shleifer, von Platen, Ma, Jernite, Plu, Xu, Scao, Gugger, Drame, Lhoest, and Rush]{wolf-etal-2020-transformers}
Thomas Wolf, Lysandre Debut, Victor Sanh, Julien Chaumond, Clement Delangue, Anthony Moi, Pierric Cistac, Tim Rault, Rémi Louf, Morgan Funtowicz, Joe Davison, Sam Shleifer, Patrick von Platen, Clara Ma, Yacine Jernite, Julien Plu, Canwen Xu, Teven~Le Scao, Sylvain Gugger, Mariama Drame, Quentin Lhoest, and Alexander~M. Rush.
\newblock Transformers: State-of-the-art natural language processing.
\newblock In \emph{Proceedings of the 2020 Conference on Empirical Methods in Natural Language Processing: System Demonstrations}, pages 38--45, Online, October 2020. Association for Computational Linguistics.
\newblock URL \url{https://www.aclweb.org/anthology/2020.emnlp-demos.6}.

\bibitem[Wu et~al.(2023{\natexlab{a}})Wu, Lu, Xu, Lin, Su, and Zhou]{wu2023self}
Shengguang Wu, Keming Lu, Benfeng Xu, Junyang Lin, Qi~Su, and Chang Zhou.
\newblock Self-evolved diverse data sampling for efficient instruction tuning.
\newblock \emph{arXiv preprint arXiv:2311.08182}, 2023{\natexlab{a}}.

\bibitem[Wu et~al.(2023{\natexlab{b}})Wu, Irsoy, Lu, Dabravolski, Dredze, Gehrmann, Kambadur, Rosenberg, and Mann]{wu2023bloomberggpt}
Shijie Wu, Ozan Irsoy, Steven Lu, Vadim Dabravolski, Mark Dredze, Sebastian Gehrmann, Prabhanjan Kambadur, David Rosenberg, and Gideon Mann.
\newblock Bloomberggpt: A large language model for finance.
\newblock \emph{arXiv preprint arXiv:2303.17564}, 2023{\natexlab{b}}.

\bibitem[Xia et~al.(2023)Xia, Artetxe, Zhou, Lin, Pasunuru, Chen, Zettlemoyer, and Stoyanov]{xia-etal-2023-training}
Mengzhou Xia, Mikel Artetxe, Chunting Zhou, Xi~Victoria Lin, Ramakanth Pasunuru, Danqi Chen, Luke Zettlemoyer, and Veselin Stoyanov.
\newblock Training trajectories of language models across scales.
\newblock In Anna Rogers, Jordan Boyd-Graber, and Naoaki Okazaki, editors, \emph{Proceedings of the 61st Annual Meeting of the Association for Computational Linguistics (Volume 1: Long Papers)}, pages 13711--13738, Toronto, Canada, July 2023. Association for Computational Linguistics.
\newblock \doi{10.18653/v1/2023.acl-long.767}.
\newblock URL \url{https://aclanthology.org/2023.acl-long.767}.

\bibitem[Yang et~al.(2022)Yang, Liu, and Mirzasoleiman]{pmlr-v162-yang22j}
Yu~Yang, Tian~Yu Liu, and Baharan Mirzasoleiman.
\newblock Not all poisons are created equal: Robust training against data poisoning.
\newblock In Kamalika Chaudhuri, Stefanie Jegelka, Le~Song, Csaba Szepesvari, Gang Niu, and Sivan Sabato, editors, \emph{Proceedings of the 39th International Conference on Machine Learning}, volume 162 of \emph{Proceedings of Machine Learning Research}, pages 25154--25165. PMLR, 17--23 Jul 2022.
\newblock URL \url{https://proceedings.mlr.press/v162/yang22j.html}.

\bibitem[Yang et~al.(2023{\natexlab{a}})Yang, Kang, and Mirzasoleiman]{pmlr-v202-yang23g}
Yu~Yang, Hao Kang, and Baharan Mirzasoleiman.
\newblock Towards sustainable learning: Coresets for data-efficient deep learning.
\newblock In Andreas Krause, Emma Brunskill, Kyunghyun Cho, Barbara Engelhardt, Sivan Sabato, and Jonathan Scarlett, editors, \emph{Proceedings of the 40th International Conference on Machine Learning}, volume 202 of \emph{Proceedings of Machine Learning Research}, pages 39314--39330. PMLR, 23--29 Jul 2023{\natexlab{a}}.

\bibitem[Yang et~al.(2023{\natexlab{b}})Yang, Singh, Elhoushi, Mahmoud, Tirumala, Gloeckle, Rozi{\`e}re, Wu, Morcos, and Ardalani]{yang2023decoding}
Yu~Yang, Aaditya~K Singh, Mostafa Elhoushi, Anas Mahmoud, Kushal Tirumala, Fabian Gloeckle, Baptiste Rozi{\`e}re, Carole-Jean Wu, Ari~S Morcos, and Newsha Ardalani.
\newblock Decoding data quality via synthetic corruptions: Embedding-guided pruning of code data.
\newblock \emph{arXiv preprint arXiv:2312.02418}, 2023{\natexlab{b}}.

\bibitem[Yang et~al.(2024)Yang, Gan, Karolina~Dziugaite, and Mirzasoleiman]{yang2023identifying}
Yu~Yang, Eric Gan, Gintare Karolina~Dziugaite, and Baharan Mirzasoleiman.
\newblock Identifying spurious biases early in training through the lens of simplicity bias.
\newblock In Sanjoy Dasgupta, Stephan Mandt, and Yingzhen Li, editors, \emph{Proceedings of The 27th International Conference on Artificial Intelligence and Statistics}, volume 238 of \emph{Proceedings of Machine Learning Research}, pages 2953--2961. PMLR, 02--04 May 2024.
\newblock URL \url{https://proceedings.mlr.press/v238/yang24c.html}.

\bibitem[Yu et~al.(2024)Yu, Jiang, Shi, YU, Liu, Zhang, Kwok, Li, Weller, and Liu]{yu2023metamath}
Longhui Yu, Weisen Jiang, Han Shi, Jincheng YU, Zhengying Liu, Yu~Zhang, James Kwok, Zhenguo Li, Adrian Weller, and Weiyang Liu.
\newblock Metamath: Bootstrap your own mathematical questions for large language models.
\newblock In \emph{The Twelfth International Conference on Learning Representations}, 2024.
\newblock URL \url{https://openreview.net/forum?id=N8N0hgNDRt}.

\bibitem[Yue et~al.(2024)Yue, Qu, Zhang, Fu, Huang, Sun, Su, and Chen]{yue2023mammoth}
Xiang Yue, Xingwei Qu, Ge~Zhang, Yao Fu, Wenhao Huang, Huan Sun, Yu~Su, and Wenhu Chen.
\newblock {MA}mmo{TH}: Building math generalist models through hybrid instruction tuning.
\newblock In \emph{The Twelfth International Conference on Learning Representations}, 2024.
\newblock URL \url{https://openreview.net/forum?id=yLClGs770I}.

\bibitem[Zhang et~al.(2020)Zhang, Kishore, Wu, Weinberger, and Artzi]{Zhang2020BERTScore}
Tianyi Zhang, Varsha Kishore, Felix Wu, Kilian~Q. Weinberger, and Yoav Artzi.
\newblock Bertscore: Evaluating text generation with bert.
\newblock In \emph{International Conference on Learning Representations}, 2020.
\newblock URL \url{https://openreview.net/forum?id=SkeHuCVFDr}.

\bibitem[Zhao et~al.(2023)Zhao, Gu, Varma, Luo, Huang, Xu, Wright, Shojanazeri, Ott, Shleifer, et~al.]{zhao2023pytorch}
Yanli Zhao, Andrew Gu, Rohan Varma, Liang Luo, Chien-Chin Huang, Min Xu, Less Wright, Hamid Shojanazeri, Myle Ott, Sam Shleifer, et~al.
\newblock Pytorch fsdp: experiences on scaling fully sharded data parallel.
\newblock \emph{arXiv preprint arXiv:2304.11277}, 2023.

\bibitem[Zhou et~al.(2023{\natexlab{a}})Zhou, Liu, Xu, Iyer, Sun, Mao, Ma, Efrat, Yu, YU, Zhang, Ghosh, Lewis, Zettlemoyer, and Levy]{zhou2023lima}
Chunting Zhou, Pengfei Liu, Puxin Xu, Srini Iyer, Jiao Sun, Yuning Mao, Xuezhe Ma, Avia Efrat, Ping Yu, LILI YU, Susan Zhang, Gargi Ghosh, Mike Lewis, Luke Zettlemoyer, and Omer Levy.
\newblock {LIMA}: Less is more for alignment.
\newblock In \emph{Thirty-seventh Conference on Neural Information Processing Systems}, 2023{\natexlab{a}}.
\newblock URL \url{https://openreview.net/forum?id=KBMOKmX2he}.

\bibitem[Zhou et~al.(2023{\natexlab{b}})Zhou, Liu, Ma, Yuan, Liu, You, and Yang]{zhou2023lobass}
Haotian Zhou, Tingkai Liu, Qianli Ma, Jianbo Yuan, Pengfei Liu, Yang You, and Hongxia Yang.
\newblock Lobass: Gauging learnability in supervised fine-tuning data.
\newblock \emph{arXiv preprint arXiv:2310.13008}, 2023{\natexlab{b}}.

\end{thebibliography}
